\documentclass[journal]{IEEEtran}
\usepackage{graphicx}
\usepackage{soul,xcolor}
\usepackage{cite}
\usepackage{amsmath,amssymb,amsfonts}
\usepackage{algorithmic}
\usepackage{mathtools}
\usepackage{graphicx}
\usepackage{multirow}
\usepackage{booktabs}
\usepackage{colortbl}
\usepackage{url}
\usepackage[strings]{underscore}
\usepackage{upgreek}
\usepackage{cleveref}
\usepackage{soul}
\newcommand\+{\raisebox{0.25ex}{+}}

\usepackage{paralist}
\usepackage{colortbl}

\DeclarePairedDelimiter{\abs}{\lvert}{\rvert}

\DeclarePairedDelimiter\floor{\lfloor}{\rfloor}

\newcommand{\etal}{~\textit{et~al.}}
\newcommand{\micron}{\ensuremath{\upmu}m}

\def\NVb#1{#1} %

\def\NV#1{#1} %
\def\NVd#1{} %
\def\NVold#1{#1} %
\def\NVdold#1{} %

\begin{document}
\title{SD-RetinaNet: Topologically Constrained Semi-Supervised
Retinal Lesion and Layer Segmentation in OCT}

\author{
****** ****** \and
****** ****** \and
****** ****** \and
****** ******
}
\author{Botond Fazekas, 
Guilherme Aresta, 
Philipp Seeböck,
Julia Mai, 
Ursula Schmidt-Erfurth, 
Hrvoje Bogunovi\'c
\thanks{The financial support by the the Christian Doppler Research Association, Austrian Federal Ministry of Economy, Energy and Tourism, the National Foundation for Research, Technology and Development, and Heidelberg Engineering is gratefully acknowledged.}\thanks{B. Fazekas, G. Aresta and H. Bogunovi\'c are with the Christian Doppler Laboratory for Artificial Intelligence in Retina, and with the Institute of Artificial Intelligence, Center for Medical Data Science, Medical University of Vienna, 1090 Vienna, Austria. }\thanks{P. Seeböck is with the Computational Imaging Research Lab, Department of Biomedical Imaging and Image-Guided Therapy, Medical University of Vienna, 1090 Vienna, Austria.}
\thanks{J. Mai and U. Schmidt-Erfurth are with the Department of Ophthalmology and Optometry, Medical University of Vienna, 1090 Vienna, Austria.}
}

\maketitle              %
\begin{abstract}
Optical coherence tomography (OCT) is widely used for diagnosing and monitoring retinal diseases, such as age-related macular degeneration (AMD). The segmentation of biomarkers such as layers and lesions is essential for patient diagnosis and follow-up. Recently, semi-supervised learning has shown promise in improving retinal segmentation performance. However, existing methods often produce anatomically implausible segmentations\NVold{, fail to effectively model layer-lesion interactions,} and lack guarantees on topological correctness.

To address these limitations, we propose a novel semi-supervised model that \NVdold{enforces anatomically correct segmentation} \NVold{introduces a fully differentiable biomarker topology engine to enforce anatomically correct segmentation} of lesions and layers\NVold{.} \NVdold{while leveraging unlabeled and partial datasets with anatomical priors encoded as loss terms.} \NVold{This enables joint learning with bidirectional influence between layers and lesions, leveraging unlabeled and diverse partially labeled datasets.} Our model learns a disentangled representation, separating spatial and style factors.  This approach enables more realistic layer segmentations and improves lesion segmentation, while strictly enforcing lesion location in their anatomically plausible positions \NVold{ relative to the segmented layers}.

We evaluate the proposed model on \NVdold{a dataset of OCT scans with AMD}\NVold{public and internal datasets of OCT scans} and show that it outperforms the current state-of-the-art in both lesion and layer segmentation\NVold{, while demonstrating the ability to generalize layer segmentation to pathological cases using partially annotated training data}. Our results demonstrate the potential of using anatomical constraints in semi-supervised learning for accurate\NVold{, robust,} and trustworthy retinal biomarker segmentation.

\end{abstract}


\section{Introduction}
Optical Coherence Tomography (OCT) is the gold standard imaging modality for the monitoring of retinal pathologies and the management of patients with eye diseases. OCT provides clinicians with a detailed cross-sectional depiction of the retina, facilitating the assessment of diverse imaging biomarkers, notably the thickness of the retinal layers and volumes of retinal fluid compartment.
These are crucial indicators for identifying and monitoring retinal diseases and guiding the treatment of pathologies such as diabetic macular edema (DME) and age-related macular degeneration (AMD), both prominent causes of global vision impairment~\cite{2004_Bressler}. In particular, the manifestations of AMD in the eye include, among others, intraretinal cystoid fluid (IRF), subretinal fluid (SRF), subretinal hyperreflective material (SHRM), and pigment epithelial detachment (PED)~\cite{2017_SchmidtErfurth, 2021_Liefers}. 

Large amount of data accrued during routine clinical practice, in conjunction with the demanding clinical workflow, makes manual quantification of retinal biomarkers unfeasible in clinical practice. To combat this, automated retinal biomarker segmentation techniques have been developed, along with the commercialization and widespread adoption of OCT technology~\cite{2010_Abramoff}.
At the same time, the large variability of pathological manifestations, their effect on the signal, as well as the nature of OCT imaging, characterized by factors such as speckle noise, intensity fluctuations, and shadowing induced by biological tissues, presents a challenging task for the segmentation of retinal layers. 

Recent advancements in retinal biomarker segmentation algorithms frequently rely on supervised deep learning methodologies, necessitating extensive training datasets annotated with meticulous precision\cite{2018_DeFauw}. However, the acquisition of high-fidelity manual annotations for pathological scenarios entails a multifaceted, labor-intensive, and resource-demanding process.
\NV{Therefore, automated solutions must leverage prior anatomical knowledge. Specifically, while retinal layers maintain a consistent relative position and lesions appear in well-defined locations, the presence of lesions critically distorts layer morphology. This creates a complex \emph{bidirectional influence}: layer anatomy constrains plausible lesion locations, while lesions, in turn, deform the layers. Accurately modeling this intricate relationship, especially while ensuring topological consistency, is a major challenge not fully addressed by methods treating these segmentations independently or sequentially\cite{2017_Montuoro, 2018_DeFauw, 2019_Hassan, 2021_He, 2021_Wang, 2023_Melo}}.
\NVd{Therefore, automated solutions should prioritize label efficiency, mitigating the need for extensive annotations by capitalizing on the often abundant volumes of unlabeled data acquired from clinical routine, as well as on the prior knowledge of the anatomical behavior of these structures. Specifically, within the domain of OCT retinal segmentation, there exists a structured and consistently topological organization of retinal layers and pathomorphological features.
That is, \emph{retinal layers maintain a consistent relative position, while retinal lesions exhibit well-defined locations within these layers}. However, the presence of lesions can also distort the shape of the layers, indicating a strong mutual interaction between the layers and the lesions.}

\NV{This challenge is compounded by the nature of available data. Due to labor costs and specialized expertise, simultaneously annotating both layers and lesions across large datasets is rare. Consequently, many public and clinical datasets are \emph{partially annotated}, with some containing only layer annotations, and others only lesion annotations. A prominent, unsolved challenge is, therefore, how to effectively train a single, robust model that leverages these diverse, partially annotated sources while respecting the complex anatomical relationships described above.}

\NVd{Additionally, due to labor costs and need for specialized expertise, simultaneously annotating multiple lesions and layers at the voxel level within extensive datasets presents significant challenges. 
Consequently, numerous datasets have only annotations for a subset of layers and/or lesions.
Thus, a prominent challenge in multi-biomarker segmentation arises from the \emph{issue of partial labeling}, i.e. how to effectively train a single model that leverages these diverse, partially annotated datasets while respecting the complex anatomical relationships. }

\NV{In this study, we propose a novel segmentation framework for retinal OCT designed specifically to address these dual challenges. Our approach's core innovation is a differentiable biomarker topology engine, which, for the first time, enables joint learning with true bidirectional influence between layers and lesions under strict anatomical constraints. To apply this framework in a realistic data environment, we build upon a disentangled representation \cite{2019_Chartsias,2022_Fazekas_CONF}. This supporting architecture is essential for our semi-supervised strategy, allowing the model to effectively integrate and learn from diverse sources of unlabeled and, crucially, partially annotated data. The proposed approach was extensively evaluated on multiple public and internal datasets, demonstrating state-of-the-art performance and showing its unique ability to generalize layer segmentation to pathological cases by synergistically learning from disparate data sources.}

\NVd{In this study, we propose a novel segmentation method for retinal OCT designed specifically to address these challenges. Our approach builds upon a disentangled representation framework to enable learning from unlabeled and partially annotated data. Crucially, we introduce a novel retinal biomarker topology engine and associated anatomical priors, which enforce topological consistency and facilitate bidirectional joint learning, allowing layer and lesion segmentations to mutually refine each other. The proposed approach was extensively evaluated on multiple public and private datasets, containing images acquired with different OCT devices and retinal diseases, showing state-of-the-art performance on this challenging segmentation task.}

\subsection{Related works}
\subsubsection{Retinal biomarker segmentation}
Early retinal layer segmentation algorithms typically involved extracting hand-crafted features from B-scans (cross-sectional images of the retina) and using graph-based techniques to infer the positions of retinal surfaces. For instance, the Iowa Reference Algorithms \cite{2006_Li, 2014_Zhang} transformed OCT data into a graph structure, using graph-search algorithms to identify surface positions while maintaining topological accuracy, layer thickness constraints, and smoothness. Subsequent enhancements to these graph-based techniques were made in \cite{2010_Chiu, 2013_Dufour, 2014_Srinivasan, 2024_Chen}.

Similarly, traditional lesion segmentation models also relied on hand-crafted image features. In~\cite{2010_Quellec}, they proposed an anomaly detection model based on retinal texture for lesion segmentation. Later, in~\cite{2012_Chen}, they proposed a 3D voxel classification model with a graph-cut approach. The interdependence of retinal layer and fluid segmentation was already realized in \cite{2015_Chiu} and a joint segmentation method for these tasks was proposed, based on a dynamic programming algorithm. 

However, these retinal biomarker segmentation methods rely heavily on hand-crafted image features, leading to a potentially degraded performance in the presence of noise, imaging artifacts or severe pathologies.  As a hybrid approach, a machine learning-based method was proposed in~\cite{2017_Montuoro}, which combined unsupervised feature representation and heterogeneous spatial context with a graph-theoretic surface segmentation, which allowed a joint segmentation of layers and lesions. 

With the rise of deep learning, several convolutional neural network (CNN)-based retinal biomarker segmentation methods have been introduced. These were predominantly built on U-Net \cite{2015_Ronneberger_CONF} and its variants. ReLayNet~\cite{2017_Roy} predicts the pixel-wise locations of nine retinal layers and potential fluid-filled pockets. This was later extended to 3D and various types of fluids in \cite{2018_DeFauw}. In the 2017 MICCAI Retinal OCT Fluid Challenge (RETOUCH) \cite{2019_Bogunovic}, all participants employed CNN-based methods, primarily based on the U-Net architecture. Subsequently, the self-configuring nnU-Net~\cite{2021_Isensee} established a new state-of-the-art performance on this data\footnote{\url{https://retouch.grand-challenge.org/Results/}}. In \cite{2021_Wang} a dual-branch multitask model was introduced, with the first branch outputting boundary labels while the second branch outputs region segmentation labels. The two outputs are fused together, providing constraints from region segmentation to layer surfaces\NVold{, but not explicitly modeling the influence of layers on lesions or the constraints imposed by lesion locations.}

Convolutional operators inherently possess local properties and inductive biases, and thus have a limited field of view. Consequently, transformers are increasingly adopted as an alternative to CNN-based methods for medical image segmentation, due to their global self-attention mechanism \cite{2021_Dosovitskiy}. Vision Transformers \cite{2021_Dosovitskiy} have demonstrated state-of-the-art performance on various benchmarks, though they require a large volume of training data and substantial computational resources. SwinUNETR \cite{2022_Hatamizadeh} addressed these limitations in medical image segmentation by introducing a hierarchical vision transformer, utilizing a hybrid CNN-transformer architecture, achieving both higher accuracy and improved computational efficiency. U-Mamba \cite{2024_Jun}, which inherits the self-configuring capability of nnU-Net, employs Mamba blocks in conjunction with CNNs instead of transformers, thereby efficiently enhancing the global attention mechanism of the models. 

Building on the latest developments in hybrid CNN-transformer based methodologies, TCCT \cite{2024_Tan} introduces a retinal layer segmentation model with a lightweight transformer module with cross-convolutional integration to increase the receptive field of the CNN backbone. The model yields pixel-level segmentation of an OCT scan with an optional subsequent edge-detection-like step converting the segmentation into layer boundaries, thereby partially transforming the task into a regression problem. However, this boundary conversion cannot be used effectively when fluids are present \cite{2024_Tan}\NVold{, highlighting the difficulty of layer-only methods in pathological regions}.

The aforementioned methods still allow anatomically implausible segmentations, such as multiple locations for a layer boundary position in an A-scan (a column of a B-scan). He\etal \cite{2021_He} proposed to improve the coherence of retinal layer segmentation by predicting the positions of the shallowest surface and then iteratively rectifying all deeper surfaces to ensure that they have greater or equal depth. These surfaces are obtained from a softmax mapping that encodes the most probable location of the layer boundary. In a multi-task setting, the system also predicts an independent pixel-wise prediction of both retinal layers and lesions. However, it fails to take advantage of the tight association between the different retinal lesions with their respective, plausible positions in the layers, and thus allowing for anatomically incoherent pathomorphological predictions. \NVold{Furthermore, such approaches lack mechanisms for true bidirectional influence, limiting the ability of lesion information to refine layer segmentation in complex pathological cases.} 

\NV{The above highlights the need for a unified framework where layers and lesions are not treated as independent tasks but are jointly optimized within a single, topologically-aware system.}

\subsubsection{Semi-supervised training for retinal biomarker segmentation}
\NVd{Addressing the challenge of limited or partially annotated datasets requires methods beyond standard supervised learning. Semi-supervised approaches aim to leverage the abundance of unlabeled or partially labeled data.} \NV{A primary motivation for our work is to address the challenge of leveraging diverse, partially annotated datasets. Semi-supervised learning provides a pathway to tackle this issue, moving beyond standard supervised methods that require complete annotations for every image.} A prevalent approach involves using a segmentation network to generate initial segmentations (pseudolabels). This is followed by a secondary network that evaluates the accuracy and realism of these segmentations against labeled data. Through the application of adversarial loss, the segmentation network is incentivized to generate outputs that are more realistic and aligned with the labeled data \cite{2019_Liu,2018_Sedai_CONF, 2019_Sedai_CONF}.

\NVold{Further semi-supervised methods have been proposed, often focusing on consistency regularization or contrastive learning. For instance, BE-SemiNet \cite{2023_Lu} focuses specifically on semi-supervised retinal OCT layer segmentation. It enhances boundary learning using an auxiliary distance map regression task alongside consistency regularization techniques. While effective for improving layer boundary delineation with scarce labels, BE-SemiNet is limited to layer segmentation only and does not support the joint segmentation of, or interaction between, layers and lesions. Furthermore, its reliance on distance maps may pose challenges in cases where layers are missing or severely distorted by pathology.}

\NVold{Consistency regularization methods, often building upon the Mean Teacher concept \cite{2017_Tarvainen}, enforce similar predictions for perturbed versions of the same unlabeled input. Variants like Uncertainty Guided Mean Teacher (UGMT) \cite{2022_Wang} refine this by weighting the consistency loss based on model uncertainty, while others enforce mutual consistency between different network predictions \cite{2022_Wu}. While effective at leveraging unlabeled data, these methods typically do not incorporate explicit anatomical topology constraints specific to retinal structures.}

\NVold{Contrastive learning approaches learn discriminative representations by contrasting features from similar and dissimilar regions. Methods like Pixel Contrast \cite{2023_Shi} and Cross-Patch Dense Contrastive Learning \cite{2022_Wua}, as well as state-of-the-art frameworks such as ACTION\+\+ \cite{2023_You} which uses adaptive contrast to handle class imbalance, have proven effective. However, these methods generally focus on feature discrimination rather than enforcing the global anatomical structure and spatial relationships required for accurate and plausible retinal OCT segmentation, nor are they typically designed for datasets with disparate partial annotation types (e.g., layers only vs. lesions only).}

In a different approach, SD-LayerNet\cite{2022_Fazekas_CONF}, which is a retinal layer topology-aware extension of SD-Net\cite{2019_Chartsias}, uses a disentangled representation approach, separating the input 2D OCT slice (B-scan) into spatial binarized spatial anatomical factors corresponding to layers and noise, and non-spatial style factors. Subsequently, the style factors are applied as an affine transformation on the anatomical factors to reconstruct the original image. The binarization of the anatomical factors, and the application of style factors as an affine transformation, ensures that anatomical factors cannot contain style information and vice versa. Through the retinal layer specific constraints enforced on the spatial anatomical factors, the network is forced to generate these structural building blocks corresponding to the retinal layers for a correct reconstruction. This allows the network to learn the structure of the retina even from unlabeled scans. Although SD-LayerNet lacks the ability to segment lesions, it could be extended similarly to [27] by an extra segmentation head predicting pixel-wise classifications for lesions. However, this would have the same limitations as [27], i.e. \NVdold{lack of topological coherence of the lesions and overlooking the spatial synergies with the layers.} \NVd{lack of the bidirectional synergy and the topological guarantees for lesions relative to the layers.} \NV{lack of a mechanism to enforce topological constraints on the lesions or to model the crucial \emph{bidirectional synergy} between layers and lesions, which is the central contribution of our new framework}

\subsection{Contributions}

In this study we propose an \NVold{end-to-end trained} retinal OCT segmentation model, which addresses \NVold{the challenges of anatomical inconsistency, accurately modeling bidirectional layer-lesion influence, and effectively utilizing partially labeled data}:

\begin{itemize}
    \item \NVdold{We propose a novel differentiable \emph{biomarker topology engine} that: Predicts \emph{anatomically plausible biomarker segmentations}, while simultaneously guarantees the \emph{topological consistency of retinal lesions}} \NVold{We introduce a framework for joint layer and lesion segmentation featuring a novel differentiable \emph{biomarker topology engine}. This engine enables bidirectional influence between layers and lesions, predicts \emph{anatomically plausible segmentations}, and guarantees the \emph{topological consistency of retinal lesions} relative to the layers.}
    
    \item We introduce several \emph{anatomical priors encoded as loss terms} (particularly $\mathcal{L}_{\text{lp}}$) that, \NVold{within this joint learning framework}, allow the model to \emph{effectively leverage the potential of partially labeled datasets} \NVold{(e.g., datasets containing only layer or only lesion annotations)}.
    
    \item We extensively evaluate our method via ablation and baseline studies on \emph{real-world AMD cases} from multiple \emph{public datasets} and on an \emph{independent external test set}.
\end{itemize}

\NVold{Importantly}\NVdold{While at the same time}, since our work uses a boundary regression approach \cite{2021_He}, it is not susceptible to the common issue of predicting multiple \NVold{vertical} positions for a boundary \NVdold{per} \NVold{in an} A-scan. This regression-based approach also allows for topological guarantees about the correct order of the layers. \NVdold{The utilized disentangled representation, introduced in SD-LayerNet~[7], allows the model to learn from unlabeled and partially labeled datasets, as even with the lack of annotations there is a training signal that encourages a correct reconstruction of the original image from its anatomical building blocks.} \NVold{The underlying disentangled representation framework \cite{2019_Chartsias,2022_Fazekas_CONF}, essential for our semi-supervised strategy, allows the model to learn structural features from unlabeled images via anatomical prior information and reconstruction. Crucially, this framework, combined with our anatomically informed joint learning process, also enables the effective integration and use of diverse partially labeled datasets.}

\section{Methodology}
\NV{Our proposed method, SD-RetinaNet, processes OCT images in a two-stage manner within an end-to-end framework, as illustrated in Fig.~\ref{fig:graphical-abstract}. First, an Anatomy Module generates initial, unconstrained predictions for layer boundaries and lesions. Second, our core contribution, a differentiable Biomarker Topology Engine, refines these initial predictions to enforce strict anatomical and topological consistency. This entire process is supported by a disentangled representation and reconstruction pathway, which enables the framework to be trained semi-supervisedly using labeled, partially labeled, and unlabeled data.}

\NVd{Our model (Fig. 1) exploits both annotated and non-annotated data to learn an initial intermediate regression of the layer boundaries and segmentation of retinal lesions, which is refined in a later step. The input image is disentangled into \emph{spatial factors} describing the anatomical structures in the image, and \emph{style factors}, encoding the image intensity of the spatial factors. The \emph{biomarker topology engine} corrects the inferred layer boundary positions to ensure the correct ordering of the layers, while taking the possible atrophies of retinal layers into account. The corrected boundary positions are then converted into 2D layer segmentations, which are used to restrict the possible locations of the retinal lesions. Driven by the observation that adjacent layers and lesions have different intensity values, during training time, the 2D layer surfaces, the lesion segmentations, and the inferred style factors are used to reconstruct the original image, to assist the semi-supervised learning. Thus, the network is implicitly encouraged to correctly delineate the retinal biomarkers to optimize reconstruction quality. During inference, the style encoder, the style factors and the reconstruction are not needed anymore, as the final result of the method is the set of the layer boundaries with the topological guarantees ensured, and the spatial factors, ie. the layer and corrected lesion masks.}

\subsection{\NVd{Spatial factor generation}\NV{Anatomy Module: Initial prediction}}
\begin{figure*}[tb]
\centering
\includegraphics[width=0.99\textwidth,page=1]{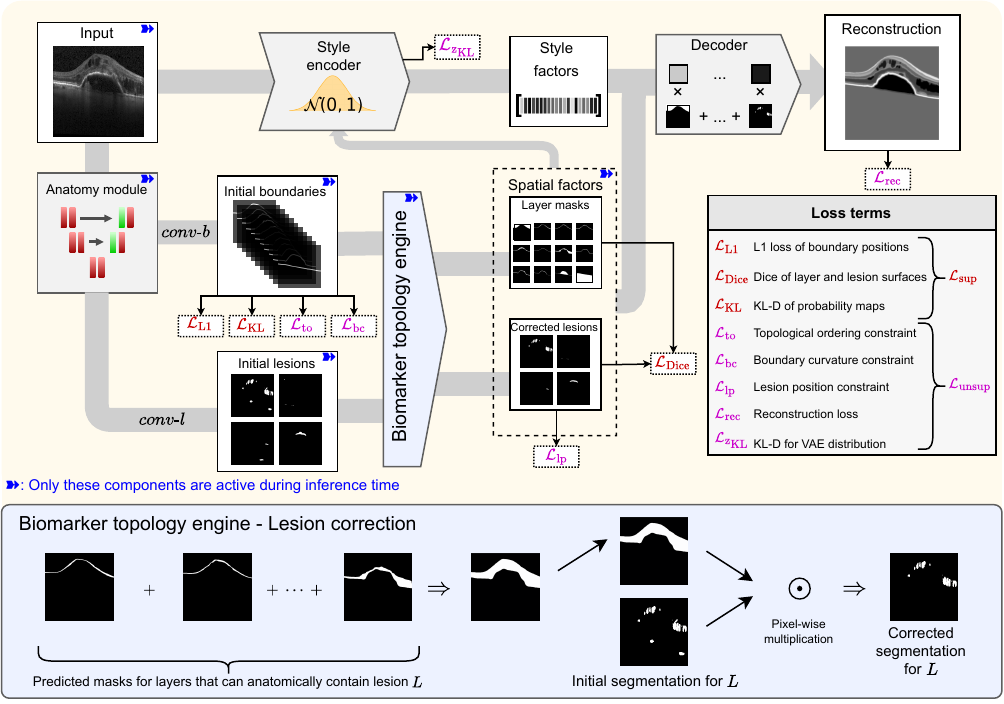}
\caption{A U-Net (\NVold{anatomy module}) creates a latent space representation from the input image. The convolutional heads \emph{conv-b} and \emph{conv-l} generate the initial layer boundaries as a probability map, and the initial lesion segmentation, respectively. The \textit{biomarker topology engine} ensures the coherent position of the layers and lesions and generates a spatial representation of the layers.
During training time, both layer and lesion segmentations (\NVold{2D} spatial factors) and the input image are used to create the style factors \NVold{(1D vectors)}, which encode the intensity values of the spatial factors\NVold{, and thus defines the general appearance of the input image}. \NVdold{The decoder then produces a reconstructed image using both factors.} \NVold{The decoder then produces a reconstructed image by creating a weighted sum of the 2D spatial factors, where the weights for each spatial factor are derived from the 1D style factors.} During inference, the style factors and the reconstructions are not generated.
} 
\label{fig:graphical-abstract}
\end{figure*}

The initial predictions are generated by a U-Net [18] with an EfficientNet-b4 encoder \cite{2019_Tan_CONF}, and a four-stage decoder with 512, 256, 128, and 64 channels (Fig.~\ref{fig:graphical-abstract} - \NVold{Anatomy module}). Two separate convolutional heads are attached to the decoder's output:

\NVd{The spatial factors are binarized maps that encode the position of the structures of interest (layers and lesions). These are generated via a U-Net [18] with an EfficientNet-b4 encoder [38], and a four-stage decoder with 512, 256, 128, and 64 channels (Fig.~1 - Anatomy module). Two convolutional output branches are connected to the output of the anatomy module: \emph{conv-b}, which performs layer boundary regression, and \emph{conv-l}, which is responsible for lesion segmentation.}%

\subsubsection{Layer boundary regression}
\NV{ The \emph{conv-b} head performs initial layer boundary regression as described in \cite{2021_He, 2022_Fazekas_CONF}, predicting a probability mass function $P(Y\mid i,b)$ for each column (A-scan) $i$ and surface boundary $b$, \NV{where Y is a discrete random variable representing the vertical position (i.e., row index) within the A-scan}. These probability mass function encodes the likelihood of each potential vertical position being the true location of boundary $b$. The final boundary position $p_{i}^{b}$ is then calculated as the expected value of this probability distribution: $p_{i}^{b}=\sum_{r}r\cdot P(r|i,b)$, where the summation is over all possible row indices $r$.}
\NVd{The initial layer boundary regression is performed as described in \cite{2021_He, 2022_Fazekas_CONF}. This involves using the convolutional layer \emph{conv-b} to predict probability mass functions $P(Y\mid i,b)$ for each column (A-scan) $i$ and surface boundary $b$,  These probability mass functions encode the likelihood Y representing the vertical position of the boundary $b$ at the $i$th A-scan, using column-wise softmax. Thus, the boundary position $p^b_i$ of the surface $b$ at the $i$th A-scan is calculated as the expected value of the probability distribution: $p^b_i = \sum_{r}{r\cdot P(r\mid i,b)}$.}

\subsubsection{Lesion segmentation}
The output of \emph{conv-l} are the lesion segmentations, obtained via Softmax. \NVold{These predictions are binarized using a differentiable rounding function before being further processed in the biomarker topology engine. This function performs standard rounding to the nearest integer during the forward pass, effectively converting the continuous output to a binary segmentation (0 or 1). During the backward pass, it acts as an identity function, allowing gradients to pass through unchanged. This approach enables gradient-based optimization while producing discrete segmentation outputs.}

Each output channel $L^k$ produced by \emph{conv-l} is associated with a lesion type $k$, while an additional channel $L^n$ is dedicated to detecting the background.

\subsection{Biomarker topology engine}
\NV{The initial predictions from the Anatomy Module can be anatomically inconsistent. The Biomarker Topology Engine is a novel, fully differentiable module designed to correct these predictions by enforcing anatomical priors. Its operation consists of three sequential steps:}
\NVd{The spatial factors obtained from \emph{conv-b} and \emph{conv-l} can be suboptimal, as they do not account for known anatomical constraints.}
\subsubsection{\NV{Iterative Layer Correction}}
To enforce a strict layer topological order, we use an iterative process to correct the boundary positions. Unlike previous approaches \cite{2021_He,2022_Fazekas_CONF}, we correct the positions bottom-up, since atrophy in the lower layers can cause the upper layers to shift downwards \cite{2021_Cleland}. Specifically, we use $\hat{p}^b_i = p^{b + 1}_i - \abs*{p^{b + 1}_i - p^{b}_i}_+$, where $\abs*{\cdot}_+$ represents the rectifier function. Here, $\hat{p}^b_i$ contains the rectified positions for an OCT A-scan, and $\hat{P}^b$ represents the collection of $\hat{p}^b_i$, containing the retinal layer positions for a B-scan.

\subsubsection{\NV{Surface-to-Mask Conversion}}
The rectified boundary positions of a B-scan $\hat{P}^b$ are then converted to 2D surfaces $C^b$ using a novel approach. In particular, $C^b = \sigma(A - \hat{P}^b)$, where $\sigma$ is the sigmoid function and $A \in \mathbb{R}^{H \times W}$, where $A_{i,j} = i$ for $j=1,2,\ldots,W$ is a row counter. $C^b$ is a  2D representation of the boundary $b$, with values close to 0 above $P^b$ and close to 1 below it. Finally, non-overlapping layer maps $M^s$ are obtained iteratively through $M^s = C^b - C^{b+1}$. During the forward pass only, $M^s$ are rounded to form binarized spatial factors.

Unlike \cite{2022_Fazekas_CONF}, which predicted 2D surfaces from the cumulative sum of the probability distribution, our novel approach operates directly on the coordinate space, facilitating possible processing of the predicted boundaries prior to the lesion segmentation.

\subsubsection{\NV{Topological Lesion Correction}}
Finally, the predicted lesion segmentations  $L^k$ are topologically corrected via $\hat{L}^k = L^k \odot \sum{M^l}_{l \in \mathcal{S}_k}$,  where the predefined $\mathcal{S}$ 
contains the indices of the layers that may contain the lesion $k$, and $\odot$ denotes the Hadamard product. 
This differentiable step ensures that \NVold{in the final output and spatial factors} no lesions are predicted outside their prespecified 
layer boundaries.

\subsection{\NVd{Image reconstruction}\NV{Disentangled Representation for Semi-Supervised Learning}}
\NVd{Note to reviewers: This sections were reordered, image reconstruction was moved boefore the loss functions and optimizations}
\NVold{The overall image reconstruction procedure, including the use of style and spatial factors, is adapted from \cite{2019_Chartsias} and \cite{2022_Fazekas_CONF}. The purpose of this component is to encourage the network to learn segmentations (corresponding to the 2D spatial factors) that, when combined with appropriate intensity values, can accurately reconstruct the original input image. This indirectly promotes the segmentation of regions with uniform intensities, which typically correspond to the same anatomical structures (layers or lesions).  This approach leverages the inherent property that neighboring retinal structures have different intensity values, and that individual layers and lesions generally exhibit uniform intensity within their boundaries.}

\subsubsection{Style factor generation}
The style factors ($\Omega$) are 1D vectors that encode the intensity values of the spatial factors. The generation of style factors follows the modality factor generation methodology of \cite{2019_Chartsias} \NVold{ and \cite{2022_Fazekas_CONF}}. Specifically, a Variational Autoencoder (VAE) is employed to generate the style factors, which are subsequently utilized in the reconstruction stage by the image decoder. \NVd{During training, the loss function $\mathcal{L}_{z{\mathrm{KL}}}$ is employed to minimize the Kullback-Leibler (KL) divergence of the estimated Gaussian distribution from the standard Gaussian distribution.}

\subsubsection{Image decoder}

The image decoder procedure is adapted from \cite{2019_Chartsias, 2022_Fazekas_CONF}, where a combination of style, binarized layer surfaces and lesions segmentations is used to generate an image of the same size as the input. The fusion process is carried out through a model that is conditioned using four FiLM layers \cite{2018_Perez} to scale and offset the intensities of the spatial factors without transmitting any spatial information via the style factors. \NVd{We focus on the reconstruction between the predicted \NVold{Internal Limiting Membrane} (ILM)  and \NVold{Bruch's membrane} (BM) surfaces, which are at the top and bottom of the retina, to avoid the impact of noise or uncertainty outside the retina. For this purpose, we use the reconstruction loss, $\mathcal{L}_{\text{rec}}$, which measures the mean absolute error (MAE) between the intensity values of the original and reconstructed images within this specific region.}

\subsection{\NV{Training objectives and loss functions}}
\NV{The model is trained end-to-end using a multi-component loss function that combines supervised signals with self-supervised anatomical priors and reconstruction objectives.}

\subsubsection{\NVd{Self-supervised training with anatomical priors} \NV{Anatomical prior losses (self-supervised)}}
\NV{These losses are calculated on the initial predictions from the Anatomy Module to guide the network towards plausible outputs, even on unlabeled data.}
\NVd{To assist training, especially when manual annotations are scarce or unavailable, we propose the following set of constraints:}

\begin{figure}[tb]
\centering
\includegraphics[width=0.45\textwidth,page=3]{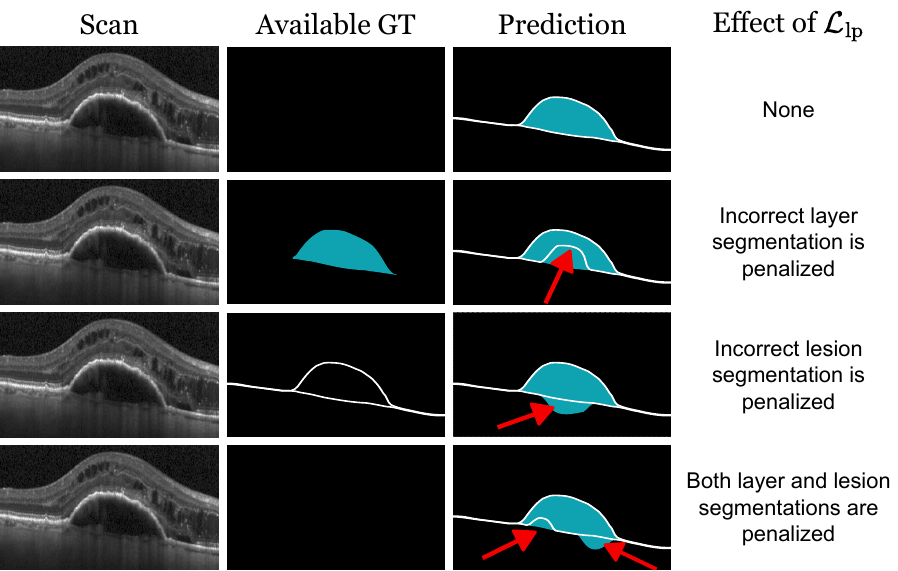}
\caption{Illustration of the lesion position constraint in various settings. A PED (blue) is defined as being located between the BM and RPE (white lines). When this condition is met in the prediction (first row), the lesion position constraint ($\mathcal{L}_{\text{lp}}$) does not influence the loss. In the presence of lesion annotation but no layer annotation (second row), if there is a discrepancy between the predicted layer and lesion segmentations, $\mathcal{L}_{\text{lp}}$ imposes a penalty on the incorrect layer segmentation. Conversely, with available layer annotation (third row), a contradiction between the predicted layer and lesion segmentations results in $\mathcal{L}_{\text{lp}}$ penalizing the incorrect lesion segmentation. In cases where no annotations are available (fourth row) but inconsistencies exist between the layer and lesion segmentations, both segmentations are penalized by $\mathcal{L}_{\text{lp}}$.}
\label{fig:lp_constraint}
\end{figure}

\begin{itemize}
\item \textit{Topological ordering constraint \NV{($\mathcal{L}_{\text{to}}$)}}
To ensure the strict ordering of retinal layers, we aim to reduce the occurrence and magnitude of topological order violations across A-scans, similarly to \cite{2022_Fazekas_CONF}: $\mathcal{L}_{\text{to}} = \frac{1}{W}\sum_s\sum_i \abs*{p^{b-1}_i - p^{b}_i}_+$.

\item \textit{Boundary curvature constraint \NV{($\mathcal{L}_{\text{bc}}$)}}
\NVold{We calculate the second-order derivative (SoD) specifically on the predicted boundary positions, using finite differences. This approach differs from \cite{2022_Fazekas_CONF}, which constrained the boundary slope to a predefined range.}
\NVdold{We use the second-order derivative (SoD) of the boundaries approximated using finite differences, unlike \mbox{\cite{2022_Fazekas_CONF}}, where the boundary slope was constrained to a specific range.}
The SoD is invariant to B-scan inclination and approximates the boundary curvature, which is a more accurate descriptor of a layer boundary than the slope. To determine the maximum expected curvature $\kappa_s$ per layer for A-scans $\delta$ pixels apart (where $\delta$ is a hyperparameter), we use the annotated training data\NVdold{ (Suppl. Table~1)}. \NVold{To remove outliers, for each layer we take the 0.999 percentile of the calculated maximum expected curvatures for each layer in the training.} With this in mind, each inferred surface curvature greater than $\kappa_s$ is penalized: $\mathcal{L}_{\text{bc}} =  \frac{1}{W}\sum_s\sum_i \abs*{\frac{|-p^{s - \floor*{\frac{\delta}{2}}}_i + 2p^{s}_i - p^{s+ \floor*{\frac{\delta}{2}}}_{i }|}{\delta} - \kappa_s}_+$.

\item \textit{Lesion position constraint  \NV{($\mathcal{L}_{\text{lp}}$)}}
To promote the correct location of lesion segmentations and to learn from unannotated samples as well, we penalize predictions $k$ lying outside of the layers defined in $\mathcal{S}_k$\NVdold{ (Suppl. Table~2)}:
$\mathcal{L}_{\text{lp}} = \frac{1}{W\cdot{}H}\sum_{k}{-\log\left(1-  L^k \odot (1 - \sum{M^l}_{l \in \mathcal{S}_k})\right)}$\NVold{, where $L^k$ is the initial segmentation prediction for lesion $k$}. The effect of this constraint is illustrated in Figure~\ref{fig:lp_constraint}.
\NVold{This constraint facilitates learning in scenarios even without complete annotations. Even if there are no annotations present, it encourages the network to predict topologically coherent segmentations. If only lesion annotations are available for a given image, and the network predicts layer boundaries that would place the lesion in an anatomically impossible location, the layer predictions are penalized via $\mathcal{L}_{\text{lp}}$, even in the absence of explicit layer ground truth. This allows the network to learn the distortions caused by lesions to the layers from the lesion segmentation. Conversely, if only layer annotations are present, incorrect lesion predictions falling outside the allowed layer ranges are penalized, even without lesion ground truth. This introduces a bidirectional influence between the layer and lesion segmentations.}

\end{itemize}

\subsubsection{\NV{Supervised losses}\NVd{Supervised training with annotations}}
\NV{These losses are calculated on the final, corrected outputs from the Biomarker Topology Engine to ensure accuracy against ground truth annotations.}
\NVd{Three loss terms are used for the supervised training:}
\begin{inparaenum}
\item $\mathcal{L}_{\mathrm{KL}}$, to describe the target distribution of  $P(Y\mid i,b)$,
\item $\mathcal{L}_{\text{L1}}$, to penalize wrongly regressed layer boundary locations with L1 loss and
\item $\mathcal{L}_{\text{dice}}$, to penalize wrongly segmented lesions.
\end{inparaenum}

In particular, to handle potential inaccuracies in pixel-level annotations, we use a normal distribution to model $P(Y\mid i,b)$ in $\mathcal{L}_{\mathrm{KL}}$. In this model, the mean of the distribution for each A-scan $i$ is the reference standard position \NVdold{$\mu^b{i}$} \NVold{$\mu^s_{i}$} of the surface, and the standard deviation $\sigma$ is a hyperparameter. During training, we use the mean Kullback-Leibler (KL) divergence as a loss term to encourage the network to learn the target distribution. We calculate the Dice loss between each lesion $\hat{L}^k$ and the reference $L^k_{ref}$, resulting in the supervised loss term:

\begin{gather}
   \mathcal{L}_{\text{sup}} = \lambda_1\mathcal{L}_{\mathrm{KL}} + \lambda_2\mathcal{L}_{\text{L1}}  + \lambda_3\mathcal{L}_{\text{dice}} = \nonumber\\ \lambda_1\sum_{s,r,i}{T\left(r \mid i, s\right)\ln \frac{T\left(r\mid i, s\right)}{P\left(r\mid i,s\right)}}%
   +\frac{\lambda_2}{S\cdot W}\sum_{s,i}\left(y^s_i - \mu^s_i\right)^2\nonumber\\ 
   +\frac{\lambda_3}{|\mathbf{k}|}\sum_{k}{\left(1 - \text{Dice}\left(\hat{L}^k, L^k_{ref}\right)\right)}
\end{gather}%
\NVold{\noindent where $T\left(Y \mid i,s\right)$ represents the target probability map, constructed using Gaussian distributions with means $\mu^s_i$ and standard deviation $\sigma$, one for each layer boundary on A-scan $i$ at surface $s$, $r \in Y$ and $|\mathbf{k}|$ is the number of different lesion types used in the training.}

\subsubsection{\NV{Reconstruction and disentanglement losses}}
\NV{We focus on the reconstruction between the predicted \NVold{Internal Limiting Membrane} (ILM)  and \NVold{Bruch's membrane} (BM) surfaces, which are at the top and bottom of the retina, to avoid the impact of noise or uncertainty outside the retina. For this purpose, we use the reconstruction loss, $\mathcal{L}_{\text{rec}}$, which measures the mean absolute error (MAE) between the intensity values of the original and reconstructed images within this specific region.}

\NV{During training, the loss function $\mathcal{L}_{z{\mathrm{KL}}}$ is employed to minimize the Kullback-Leibler (KL) divergence of the estimated Gaussian distribution by the style-encoder VAE from the standard Gaussian distribution.
}

\subsubsection{\NV{Optional triplet loss for }Forced Disentanglement}
\begin{figure}[tb]
\centering
\includegraphics[width=0.45\textwidth,page=1]{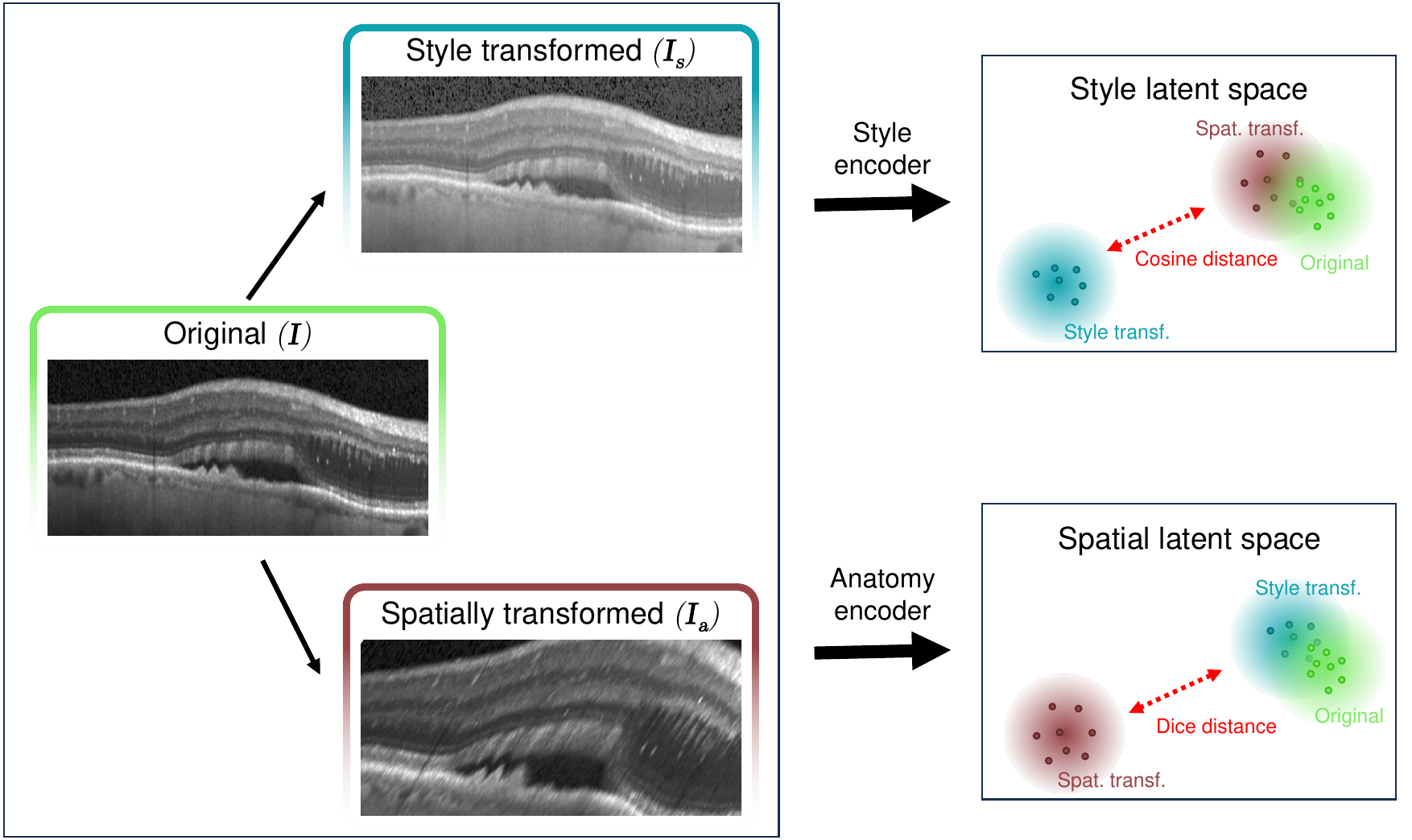}
\caption{Illustration of the forced disentanglement process. For each B-scan $I$ in the batch, a randomly spatially transformed B-scan $I_a$ and a style-transformed B-scan $I_s$ are generated. \emph{Style latent space:} A triplet loss function ensures that the cosine distance margin between the style factors of the original B-scan and the spatially transformed B-scan remains smaller than the margin between the original and the style-transformed B-scan. \emph{Spatial latent space:} The layer and lesion segmentations \NVold{(i.e. spatial factors)} should be consistent between the style-transformed and the original B-scan, and must differ for the spatially transformed B-scan.}
\label{fig:forced_disentanglement}
\end{figure}

To enhance the disentanglement of the spatial and style factors, we propose a variant of our method with two additional loss terms. This approach trains the network in a contrastive manner, ensuring that variations in the overall appearance of the scans are captured exclusively by the style factors, while spatial changes are reflected exclusively in the spatial factors (Figure~\ref{fig:forced_disentanglement}).

Specifically, for each input B-scan $I$ we generate a B-scan $I_a$ which has random affine transformations applied on, as well as a B-scan $I_s$ on which random style transformations are applied, including change in the contrast, noise level, blur. We introduce a triplet loss for both the style factors $(\mathcal{L}_{\text{triplet-s}})$ and for the spatial factors $(\mathcal{L}_{\text{triplet-a}})$:
\begin{align}
  \mathcal{L}_{\text{triplet-s}} &= \abs{S_C\left(\Omega_{I}, \Omega_{I_a}\right) - S_C\left(\Omega_{I}, \Omega_{I_s}\right)}_+
\end{align}
where $S_C$ denotes the cosine similarity function, and $\Omega$ denotes the style factor vector,

\begin{align}
  \mathcal{L}_{\text{triplet-a}} &= \abs{\mathrm{MSE}\left(P_{I}, P_{I_s}\right) - \mathrm{MSE}\left(P_{I}, P_{I_a}\right)}_+ \notag\\
  &+ \abs{\mathrm{DL}\left(L_{I}, L_{I_s}\right) - \mathrm{DL}\left(L_{I}, L_{I_a}\right)}_+
\end{align}
where $\mathrm{MSE}$ denotes the mean squared error, $\mathrm{DL}$ denotes the Dice loss, $P$ denotes the layer boundaries and $L$ denotes the predicted lesions.
We sum the two triplet losses under the common term: $\mathcal{L}_{\text{triplet}} = \mathcal{L}_{\text{triplet-a}} + \mathcal{L}_{\text{triplet-s}}$

\subsubsection{Overall cost function}
The overall cost function $\mathbf{L}$ \NVold{for the end-to-end training} is a composition of the supervised $\mathcal{L}_{\text{sup}} = \lambda_1 \mathcal{L}_{\mathrm{KL}} + \lambda_2 \mathcal{L}_{\text{mse}} + \lambda_3 \mathcal{L}_{\text{dice}}$, the anatomical priors $\mathcal{L}_{\text{prior}} = \lambda_4\mathcal{L}_{\text{to}} + \lambda_5\mathcal{L}_{\text{bc}} + \lambda_6\mathcal{L}_{\text{lp}}$ loss, the style encoder loss $\mathcal{L}_{z_{\mathrm{KL}}}$, the triplet losses $\mathcal{L}_{\text{triplet}}$ and the reconstruction loss $\mathcal{L}_{\text{rec}}$:
    \begin{align*}
    \mathbf{L} &= \lambda_7\mathcal{L}_{z_{\mathrm{KL}}} + \lambda_8\mathcal{L}_{\text{rec}} + \lambda_9\mathcal{L}_{\text{sup}} + \lambda_{10}\mathcal{L}_{\text{triplet}} + \mathcal{L}_{\text{prior}}
    \end{align*}

\section{Experimental setup}

\subsection{Datasets} 
\subsubsection{Public datasets}

The \textbf{HCMS} dataset~\cite{2021_He} comprises 14 healthy controls (HCs) and 21 individuals diagnosed with multiple sclerosis (MS). Each participant's data includes 49 B-scans obtained with a Spectralis (Heidelberg Engineering, Germany) scanner, with nine surfaces manually annotated. The scans measure 496 × 1024 in dimension and possess a lateral resolution of 5.8\micron, an axial resolution of 3.9\micron, and a B-scan interval of 123.6\micron. As we do not validate/test our method in this dataset, all of the scans are used for training.

The \textbf{RETOUCH} dataset includes 112 OCT volumes centered on the macula obtained from 112 patients diagnosed with macular edema due to AMD or RVO. Each B-scan was manually annotated pixel-wise for intraretinal fluid (IRF), subretinal fluid (SRF), and pigment epithelial detachment (PED). The training dataset comprises 70 OCT volumes, while the test dataset contains 42 volumes, however we did not utilize the latter. The OCT volumes were captured using three different devices: Cirrus HD-OCT (Zeiss Meditec, Germany), Spectralis (Heidelberg Engineering, Germany), and T-1000/T-2000 (Topcon, Japan), with approximately equal number of scans from each. All volumes encompass a macular region measuring 6$\times$6 mm$^2$. Depending on the experimental setup, we either use all of the scans available for training regardless of the vendors, or the Spectralis scans only.

As an independent external test set, we used the \textbf{DukeDME} dataset \cite{2015_Chiu}. It contains 10 Spectralis volumes with a resolution of $496 \times 768 \times 61$ voxels from 10 patients with diabetic macular edema (DME). Within each volume, 11 B-scans are annotated by both a senior and a junior ophthalmologist for 8 retinal layers and macular edema. \NVdold{The annotated layers correspond to those in the HCMS dataset, except for the ELM, which is not annotated in this dataset.}  \NVold{These 8 annotated layers form a direct subset of the 9 layers annotated in the HCMS dataset, lacking only the ELM annotation, thus ensuring compatibility for evaluation when training on HCMS data.}  Similar to \cite{2015_Chiu}, \cite{2017_Rathke_CONF}, \cite{2016_Karri}, and \cite{2017_Roy}, we used the 5 patients (55 B-scans) of DukeDME with severe edema. However, as we used this dataset as an external test set, we did not perform training on it, thus the scans of the last 5 patients were not utilized.

\subsubsection{Internal dataset}
The dataset comprises 509 macula centered Spectralis (Heidelberg Engineering, Germany) OCT volumes from 462 patients, with resolutions spanning 18-97 B-scans, each with a size of $512-1024 \times 496$ pixels. The scans were taken during clinical routine check-ups in neovascular AMD patients who underwent treatment, thus representing a real-world clinical scenario. For $85$ volumes, the lesions were manually annotated by experts on every B-scan, while \NVold{11} layers were manually annotated on 7 B-scans per volume. From these, the number of volumes containing IRF, SRF and SHRM is $42$, $50$ and $42$ respectively.
\NVdold{The annotated data was randomly split patient-wise with multi-label aware data stratification into training (42 scans), validation (17) and test (26) set. The remaining unannotated part of the dataset, consisting of 387 scans, was used only for self-supervised training.} \NVold{We performed 3-fold cross-validation on the 85 annotated volumes. Folds were created using patient-wise, multi-label aware data stratification ($\sim$28 volumes/fold) to ensure similar lesion distributions and prevent subject overlap between training/validation and test sets across iterations. In each iteration, 2 folds formed the training/validation set, with the remaining fold used for testing. This combined training/validation data was further split patient-wise and stratified into 70\% for internal training and 30\% for internal validation. This internal validation set was used during each training run to monitor performance and select the optimal model epoch (i.e., the specific model weights) for final evaluation on the held-out test fold. The 387 unannotated scans were used alongside the respective internal training data in each fold for semi-supervised training.} All scans were resampled to a uniform lateral resolution of 6.0 $\upmu$m/px.
\subsection{\NV{Evaluation metrics and statistical analysis}}
\NV{To assess performance, we used distinct metrics tailored for each segmentation task. For \emph{layer segmentation}, we report the \emph{Mean Absolute Distance (MAD)} in micrometers (\micron). MAD is chosen as it directly measures the average boundary localization error in a clinically relevant scale and, unlike region-based metrics like Dice, is not biased by the natural variation in layer thickness across the retina. For \emph{lesion segmentation}, we report the \emph{Dice Similarity Coefficient (DSC)}, the standard metric for evaluating the volumetric overlap of segmentation masks, which is well-suited for these variably shaped regions.}

\NV{To account for the interdependence of B-scans from the same patient volume, we performed statistical analysis using a linear mixed-effects model \cite{2014_Bates}, with $p < 0.05$ considered significant.}

\subsection{\NV{Experimental scenarios}\NVd{Experiments}}
We designed four distinct experimental scenarios to comprehensively evaluate our method's performance, robustness, and generalization capabilities.

\subsubsection{\NV{Main performance evaluation}}
\paragraph{Baseline studies}
To evaluate the performance of our proposed approach, we compared it with several segmentation methods: ReLayNet~\cite{2017_Roy}, nnU-Net\cite{2021_Isensee}, U-Mamba~\cite{2024_Jun}, SwinUNETR~\cite{2022_Hatamizadeh}, TCCT~\cite{2024_Tan}, and the \NVdold{retinal biomarker segmentation method} \NVold{layer boundary regression method (LBRM)} introduced by \cite{2021_He}. \NVold{Additionally, we included comparisons with two state-of-the-art semi-supervised methods: ACTION\+\+~\cite{2023_You}, a general medical image segmentation model, and BE-SemiNet~\cite{2023_Lu}, designed for retinal OCT layer segmentation.}

\NV{For a fair and consistent comparison, several implementation details were necessary. We focused our evaluation on these 2D-based methods as they are better suited to the sparse, slice-wise annotation protocol common in clinical datasets and addressed in our work. While 3D methods like Liu\etal~\cite{2024_Liu} exist, their reliance on inter-slice coherence is challenged by the large, variable distances between the few annotated B-scans in our data. Our preliminary evaluation of this representative 3D method confirmed it struggled in this setting. Since TCCT, SD-LayerNet and BE-SemiNet were originally designed for layer segmentation only, we adapted their architectures by adding output channels for lesion segmentation. For all pixel-based methods (ReLayNet, nnU-Net, U-Mamba, SwinUNETR, ACTION\+\+, and TCCT), layer boundary positions were extracted in a post-processing step, defined as the first occurrence of the corresponding layer class in each A-scan. To handle erroneous predictions outside the retina, only the largest connected component for each layer was retained. Furthermore, for fairness, the backbones of ReLayNet and LBRM were replaced with a U-Net with an EfficientNet-b4 encoder to match our anatomy module.}

\NVd{Since BE-SemiNet were originally designed for layer segmentation only, we adapted them by adding output channels for lesion segmentation to enable a comparison across all tasks. As ReLayNet, nnU-Net, U-Mamba, SwinUNETR, ACTION and TCCT are pixel-based segmentation methods,}\NVd{ the boundary positions were calculated in a subsequent post-processing step. These were defined as the first occurrence of the class corresponding to the layer in each A-scan. As these methods sometimes erroneously predict layer classes above the ILM and below the BM, only the largest connected component belonging to a layer's class was retained. }

\NVd{For fairness, we compared our approach to an adapted version of ReLayNet\cite{2017_Roy} and LBRM\cite{2021_He}, where the segmentation backbones were replaced for a U-Net with an EfficientNet-b4 feature encoder to match our anatomy module.}

\paragraph{Ablation studies}
To investigate the impact of the different modules of our model, we conducted several ablation studies. To evaluate the effect of the semi-supervised lesion position constraint $\mathcal{L}_{\text{lp}}$, we trained our network without this loss term (\emph{SD-RetinaNet - $\mathcal{L}_{\text{lp}}$}). \NVold{To assess the importance of the topological constraint and the boundary curvature constraints, we trained the network without these loss terms (\emph{SD-RetinaNet - $\mathcal{L}_{\text{to}}$} and \emph{SD-RetinaNet - $\mathcal{L}_{\text{bc}}$}). } We assessed the performance of our model when trained without disentanglement and reconstruction (\emph{SD-RetinaNet - $\mathcal{L}_\text{rec}$}). Also, we observed the performance of the biomarker segmentation method of He\etal \NVb{, with adding biomarker topology engine with the 1D-2D conversion and the lesion position constraint} $\mathcal{L}_{\text{lp}}$ thus creating an additional semi-supervised baseline (He\etal + $\mathcal{L}_{\text{lp}}$). This corresponds to a configuration of our model without reconstruction and the anatomical priors $\mathcal{L}_{\text{to}}$ and $\mathcal{L}_{\text{bc}}$.
We also evaluated the effect of changing the U-Net + EfficientNet-b4 backbone to the proposed architecture of TCCT~\cite{2024_Tan} (\emph{SD-RetinaNet w. TCCT BB.}).

\subsubsection{Partial annotation generalization}
Although numerous public OCT segmentation datasets are available, only a few include annotations for both retinal layers and lesions. To assess the performance of our method with partially annotated datasets, we trained our model using the HCMS dataset\cite{2021_He}, which provides layer segmentation annotations for both healthy and multiple sclerosis (MS) scans, and the Spectralis scans of the RETOUCH dataset, which includes annotations for PED, SRF, and IRF in patients with AMD and RVO. To evaluate the performance of our model trained on partially labeled datasets on an independent external dataset, we use the first 5 volumes of the DukeDME dataset as this is common practice within the field \cite{2015_Chiu, 2021_He, 2024_Tan}. For this evaluation, we combined our model's predictions for IRF and SRF to match the "macular edema" annotation present in the DukeDME dataset.%

We were not able to test the non-semi-supervised methods under these experimental settings because none of them yielded meaningful results after training. Instead, for the layer segmentation performance, we trained these methods solely on the HCMS dataset and evaluated their performance on the first 5 volumes of the DukeDME dataset. Similarly, we trained these methods on the RETOUCH dataset and evaluated the lesion segmentation performance on the same volumes in the DukeDME dataset. This approach also provides insight into the impact of domain shift in this experimental setting.

\subsubsection{\NV{Robustness to limited labeled data}\NVd{Label efficiency evaluation}} 

\NVold{To further investigate the label efficiency and robustness of our semi-supervised framework, especially relevant in clinical settings where extensive annotations may be scarce, we conducted additional experiments using significantly reduced amounts of labeled training data. The goal was to assess how effectively our proposed methods, compared to other semi-supervised approaches, could leverage anatomical priors and the reconstruction task when direct supervision is minimal.}

\NVold{These experiments were performed using the cross-validation setup on our internal dataset. For each of the 3 folds, we created smaller training subsets containing only 10\% and 25\% of the volumes available in that fold's original training split. These subsets were generated using the same patient-wise, multi-label aware stratification method used for the main split to maintain representative distributions of lesion types and severities, even within these smaller subsets.}

\NVold{Recognizing that training outcomes can have higher variability with smaller datasets, we performed multiple independent runs for these limited data conditions. Specifically, for the 10\% data setting, we generated 10 distinct subsets and trained the models independently on each. For the 25\% data setting, we used 4 distinct subsets and corresponding training runs. This repetition helps provide a more stable estimate of performance under data scarcity.}

\NVold{We evaluated our proposed methods (SD-RetinaNet and SD-RetinaNet + $\mathcal{L}_{\text{triplet}}$) and the relevant semi-supervised baseline methods (BE-SemiNet, ACTION++, LBRM + $\mathcal{L}_{\text{lp}}$, SD-LayerNet) under these 10\% and 25\% training data conditions. Models trained on each subset were evaluated on the fixed test set corresponding to their respective cross-validation fold.}

\subsubsection{\NV{Cross-device generalization}}
\NV{To investigate the effect of our disentanglement framework on cross-device generalization, we designed a leave-one-out cross-vendor validation using the multi-device RETOUCH dataset, which contains scans from Spectralis, Topcon and Cirrus devices.}

\NV{The first two scenarios evaluated generalization from a mixed-vendor training set to a single unseen vendor:}
\begin{itemize}
    \item \NV{Models were trained on 50\% of the Spectralis volumes and 50\% of the Cirrus volumes, and evaluated on all held-out Topcon volumes.}
    \item \NV{Models were trained on 50\% of the Spectralis volumes and a corresponding number of Topcon volumes, and evaluated on all held-out Cirrus volumes.}
\end{itemize}

\NV{The third scenario was designed as a more demanding generalization task, testing the ability to learn from a single manufacturer and apply to multiple unseen ones:}
\begin{itemize}
    \item \NV{Models were trained using only the Spectralis volumes from RETOUCH and HCMS, and evaluated on a combined test set of all Topcon and Cirrus volumes.}
\end{itemize}
\NV{In all scenarios, the layer-annotated HCMS dataset was included during training to provide layer supervision. As the RETOUCH dataset only contains lesion annotations, performance for this experiment was evaluated solely on lesion segmentation using the Dice Similarity Coefficient.}

\subsubsection{\NV{Analysis of disentangled latent space}\NVd{Effect of forced disentanglement on the latent space}}
In order to assess the effect of the forced disentanglement process, we conducted experiments with the forced disentanglement term added.
To achieve this, for each B-scan $I$, we generated a B-scan $I_a$, with random rotation (between -30° and 30°), random shearing (between -0.2 and 0.2) and random scaling (from 0.9 to 1.1).
Additionally, we generated a B-scan $I_s$, with random style transformations, including change in the contrast, noise level and blurring. (\emph{SD-RetinaNet + $\mathcal{L}_{\text{triplet}}$})

\begin{table*}[tbh]
\centering

\caption{Quantitative layer segmentation results on the \NVdold{MultiBioMarker}\NVold{internal} dataset including the ablation experiments. The values correspond to the mean absolute distance (MAD) metric in terms of \micron.}
\label{tab:mb_layer_results}
\resizebox{\linewidth}{!}{%
\begin{tabular}{l|ccccccccccc!{\vrule width \lightrulewidth}c} 
\toprule
\multicolumn{1}{l}{Method}                               & ILM                     & RNFL/GCL                & GCL/IPL           & IPL/INL                 & INL/OPL                 & OPL/ONL                 & ELM                     & IS/OS                   & OB-RPE                  & IB-RPE                  & \multicolumn{1}{c}{BM} & \textbf{Total}                    \\ 
\midrule
\multicolumn{13}{c}{\textbf{Baseline methods}} \\
\midrule
\rowcolor[rgb]{0.92,0.92,0.92} ReLayNet \cite{2017_Roy}		& \shortstack{$3.58^{*\dagger}$  \tiny{(2.19)}} 	& \shortstack{$14.78^{*\dagger}$  \tiny{(16.02)}}	& \shortstack{$16.19^{*\dagger}$  \tiny{(13.50)}}	& \shortstack{$17.03^{*\dagger}$  \tiny{(14.07)}}	& \shortstack{$6.36^{*\dagger}$  \tiny{(3.40)}}	& \shortstack{$6.88^{*\dagger}$  \tiny{(3.22)}}	& \shortstack{$5.69^{*\dagger}$  \tiny{(2.61)}}	& \shortstack{$4.39^\dagger$  \tiny{(2.46)}}	& \shortstack{$5.23^{*\dagger}$  \tiny{(3.58)}}	& \shortstack{$5.20^\dagger$  \tiny{(4.22)}}	& \shortstack{$9.35^{*\dagger}$  \tiny{(13.79)}}	& \shortstack{$8.61^{*\dagger}$  \tiny{(2.70)}}		\\
nnU-Net \cite{2021_Isensee}		& \shortstack{$3.64^{*\dagger}$  \tiny{(1.89)}} 	& \shortstack{$4.07^*$  \tiny{(1.23)}}	& \shortstack{$6.27^\dagger$  \tiny{(2.27)}}	& \shortstack{$5.07$  \tiny{(3.10)}}	& \shortstack{$5.26^\dagger$  \tiny{(2.02)}}	& \shortstack{$6.34^\dagger$  \tiny{(2.58)}}	& \shortstack{$6.12^{*\dagger}$  \tiny{(3.32)}}	& \shortstack{$4.54^{*\dagger}$  \tiny{(2.51)}}	& \shortstack{$5.78^{*\dagger}$  \tiny{(4.35)}}	& \shortstack{$6.07^\dagger$  \tiny{(11.58)}}	& \shortstack{$7.79^{*\dagger}$  \tiny{(7.38)}}	& \shortstack{$5.54^{*\dagger}$  \tiny{(2.65)}}		\\
\rowcolor[rgb]{0.92,0.92,0.92} U-Mamba	\cite{2024_Jun}	& \shortstack{$3.54^{*\dagger}$  \tiny{(1.27)}} 	& \shortstack{$4.11^*$  \tiny{(1.38)}}	& \shortstack{$6.17$  \tiny{(2.20)}}	& \shortstack{$4.92^\dagger$  \tiny{(2.14)}}	& \shortstack{$5.14^\dagger$  \tiny{(1.95)}}	& \shortstack{$6.41^\dagger$  \tiny{(3.15)}}	& \shortstack{$6.01^{*\dagger}$  \tiny{(3.38)}}	& \shortstack{$4.57^{*\dagger}$  \tiny{(2.35)}}	& \shortstack{$5.98^{*\dagger}$  \tiny{(5.48)}}	& \shortstack{$6.11^\dagger$  \tiny{(12.12)}}	& \shortstack{$7.80^{*\dagger}$  \tiny{(7.74)}}	& \shortstack{$5.52^{*\dagger}$  \tiny{(2.97)}}		\\
SwinUNETR \cite{2022_Hatamizadeh}		& \shortstack{$3.46^{*\dagger}$  \tiny{(1.26)}} 	& \shortstack{$3.95$  \tiny{(1.21)}}	& \shortstack{$6.30^\dagger$  \tiny{(1.64)}}	& \shortstack{$4.87^\dagger$  \tiny{(1.56)}}	& \shortstack{$5.23^\dagger$  \tiny{(1.68)}}	& \shortstack{$7.09^{*\dagger}$  \tiny{(3.58)}}	& \shortstack{$6.18^{*\dagger}$  \tiny{(3.73)}}	& \shortstack{$5.27^{*\dagger}$  \tiny{(5.38)}}	& \shortstack{$5.30^{*\dagger}$  \tiny{(2.58)}}	& \shortstack{$5.40^\dagger$  \tiny{(5.27)}}	& \shortstack{$8.36^{*\dagger}$  \tiny{(8.00)}}	& \shortstack{$5.58^{*\dagger}$  \tiny{(2.02)}}		\\
\rowcolor[rgb]{0.92,0.92,0.92} TCCT-BP \cite{2024_Tan} & \shortstack{$3.54$  \tiny{(1.84)}} 	& \shortstack{$3.94$  \tiny{(1.20)}}	& \shortstack{$6.14$  \tiny{(2.19)}}	& \shortstack{$4.78^\dagger$  \tiny{(1.71)}}	& \shortstack{$4.99$  \tiny{(1.79)}}	& \shortstack{$6.23^\dagger$  \tiny{(2.54)}}	& \shortstack{$5.92^{*\dagger}$  \tiny{(3.12)}}	& \shortstack{$4.43^\dagger$  \tiny{(2.61)}}	& \shortstack{$6.12^{*\dagger}$  \tiny{(8.46)}}	& \shortstack{$5.33^\dagger$  \tiny{(6.79)}}	& \shortstack{$7.62^{*\dagger}$  \tiny{(7.11)}}	& \shortstack{$5.37^{*\dagger}$  \tiny{(2.47)}}		\\
\NVdold{He\etal}\NVold{LBRM}\cite{2021_He} & \shortstack{$3.28$  \tiny{(1.29)}} 	& \shortstack{$4.06$  \tiny{(1.40)}}	& \shortstack{$6.35^\dagger$  \tiny{(2.00)}}	& \shortstack{$4.88^\dagger$  \tiny{(1.74)}}	& \shortstack{$5.53^\dagger$  \tiny{(2.18)}}	& \shortstack{$6.49^{*\dagger}$  \tiny{(2.69)}}	& \shortstack{$5.53^\dagger$  \tiny{(2.28)}}	& \shortstack{$4.38^{*\dagger}$  \tiny{(1.97)}}	& \shortstack{$4.70^{*\dagger}$  \tiny{(2.12)}}	& \shortstack{$4.88^\dagger$  \tiny{(2.62)}}	& \shortstack{$5.30^{*\dagger}$  \tiny{(5.96)}}	& \shortstack{$5.04^{*\dagger}$  \tiny{(1.66)}}		\\
\midrule

\multicolumn{13}{c}{\textbf{Semi-supervised baselines}} \\
\midrule

\rowcolor[rgb]{0.92,0.92,0.92} \NVdold{He\etal}\NVold{LBRM} + $\mathcal{L}_{\text{lp}}$ & \shortstack{$3.43^{*\dagger}$  \tiny{(1.47)}} 	& \shortstack{$4.23^{*\dagger}$  \tiny{(1.69)}}	& \shortstack{$6.32^\dagger$  \tiny{(1.97)}}	& \shortstack{$4.77^\dagger$  \tiny{(1.62)}}	& \shortstack{$5.68^{*\dagger}$  \tiny{(2.59)}}	& \shortstack{$6.53^{*\dagger}$  \tiny{(2.92)}}	& \shortstack{$5.78^{*\dagger}$  \tiny{(2.50)}}	& \shortstack{$4.32^\dagger$  \tiny{(1.83)}}	& \shortstack{$4.71^{*\dagger}$  \tiny{(1.91)}}	& \shortstack{$4.69^\dagger$  \tiny{(2.16)}}	& \shortstack{$4.64^\dagger$  \tiny{(4.14)}}	& \shortstack{$5.01^{*\dagger}$  \tiny{(1.58)}}		\\
\NVold{ACTION\+\+} \cite{2023_You} & \shortstack{$3.40^{*\dagger}$  \tiny{(1.27)}} 	& \shortstack{$\mathbf{3.87}$  \tiny{(1.22)}}	& \shortstack{$6.22^\dagger$  \tiny{(2.17)}}	& \shortstack{$4.71$  \tiny{(1.39)}}	& \shortstack{$5.12^\dagger$  \tiny{(1.83)}}	& \shortstack{$6.70^{*\dagger}$  \tiny{(2.87)}}	& \shortstack{$6.44^{*\dagger}$  \tiny{(4.55)}}	& \shortstack{$5.21^{*\dagger}$  \tiny{(6.01)}}	& \shortstack{$5.05^{*\dagger}$  \tiny{(2.41)}}	& \shortstack{$5.36^\dagger$  \tiny{(5.25)}}	& \shortstack{$8.36^{*\dagger}$  \tiny{(8.31)}}	& \shortstack{$5.49^{*\dagger}$  \tiny{(2.13)}}		\\
\rowcolor[rgb]{0.92,0.92,0.92} \NVold{BE-SemiNet} \cite{2023_Lu} & \shortstack{$3.26$  \tiny{(1.34)}} 	& \shortstack{$4.30^{*\dagger}$  \tiny{(1.72)}}	& \shortstack{$6.51^\dagger$  \tiny{(2.60)}}	& \shortstack{$5.24^{*\dagger}$  \tiny{(2.21)}}	& \shortstack{$5.69^{*\dagger}$  \tiny{(2.53)}}	& \shortstack{$6.62^{*\dagger}$  \tiny{(2.83)}}	& \shortstack{$5.64^{*\dagger}$  \tiny{(2.37)}}	& \shortstack{$4.25^\dagger$  \tiny{(1.69)}}	& \shortstack{$4.49^\dagger$  \tiny{(2.08)}}	& \shortstack{$4.48^\dagger$  \tiny{(2.31)}}	& \shortstack{$6.05^{*\dagger}$  \tiny{(7.70)}}	& \shortstack{$5.14^{*\dagger}$  \tiny{(1.87)}}		\\
SD-LayerNet \cite{2022_Fazekas_CONF} & \shortstack{$3.45^{*\dagger}$  \tiny{(1.52)}} 	& \shortstack{$4.14^*$  \tiny{(1.55)}}	& \shortstack{$6.22^\dagger$  \tiny{(2.04)}}	& \shortstack{$4.94^{*\dagger}$  \tiny{(1.77)}}	& \shortstack{$5.37^\dagger$  \tiny{(2.07)}}	& \shortstack{$6.30^\dagger$  \tiny{(2.45)}}	& \shortstack{$5.49^\dagger$  \tiny{(2.15)}}	& \shortstack{$4.47^{*\dagger}$  \tiny{(2.52)}}	& \shortstack{$4.77^{*\dagger}$  \tiny{(2.19)}}	& \shortstack{$4.78^\dagger$  \tiny{(2.83)}}	& \shortstack{$4.89^\dagger$  \tiny{(5.93)}}	& \shortstack{$4.98^{*\dagger}$  \tiny{(1.58)}}		\\
\midrule

\multicolumn{13}{c}{\textbf{Ablation studies}} \\
\midrule
\rowcolor[rgb]{0.92,0.92,0.92} SD-RetinaNet w/o recon & \shortstack{$3.40^{*\dagger}$  \tiny{(1.36)}} 	& \shortstack{$4.15^*$  \tiny{(1.60)}}	& \shortstack{$6.36^\dagger$  \tiny{(2.42)}}	& \shortstack{$4.83^\dagger$  \tiny{(1.99)}}	& \shortstack{$5.54^\dagger$  \tiny{(2.43)}}	& \shortstack{$6.32^\dagger$  \tiny{(2.63)}}	& \shortstack{$5.65^{*\dagger}$  \tiny{(2.46)}}	& \shortstack{$4.25^\dagger$  \tiny{(1.78)}}	& \shortstack{$4.65^\dagger$  \tiny{(2.02)}}	& \shortstack{$4.65^\dagger$  \tiny{(2.20)}}	& \shortstack{$4.48^\dagger$  \tiny{(4.09)}}	& \shortstack{$4.93^{*\dagger}$  \tiny{(1.52)}}		\\
\NVold{SD-RetinaNet - $\mathcal{L}_{\text{to}}$} & \shortstack{$3.43^{*\dagger}$  \tiny{(1.49)}} 	& \shortstack{$4.03$  \tiny{(1.47)}}	& \shortstack{$6.23^\dagger$  \tiny{(2.03)}}	& \shortstack{$4.82^\dagger$  \tiny{(1.78)}}	& \shortstack{$5.31^\dagger$  \tiny{(2.06)}}	& \shortstack{$6.14^\dagger$  \tiny{(2.37)}}	& \shortstack{$5.35^\dagger$  \tiny{(2.10)}}	& \shortstack{$4.22^\dagger$  \tiny{(1.89)}}	& \shortstack{$4.57^\dagger$  \tiny{(2.33)}}	& \shortstack{$4.79^\dagger$  \tiny{(2.85)}}	& \shortstack{$4.11^\dagger$  \tiny{(3.24)}}	& \shortstack{$4.82^\dagger$  \tiny{(1.50)}}		\\
\rowcolor[rgb]{0.92,0.92,0.92} \NVold{SD-RetinaNet - $\mathcal{L}_{\text{bc}}$} & \shortstack{$3.26$  \tiny{(1.27)}} 	& \shortstack{$4.05$  \tiny{(1.56)}}	& \shortstack{$6.23^\dagger$  \tiny{(2.75)}}	& \shortstack{$4.77$  \tiny{(2.32)}}	& \shortstack{$5.35^\dagger$  \tiny{(2.42)}}	& \shortstack{$6.26^\dagger$  \tiny{(2.56)}}	& \shortstack{$5.28^\dagger$  \tiny{(1.85)}}	& \shortstack{$4.04^\dagger$  \tiny{(1.57)}}	& \shortstack{$4.24^\dagger$  \tiny{(1.44)}}	& \shortstack{$4.35^{*\dagger}$  \tiny{(1.77)}}	& \shortstack{$4.48^\dagger$  \tiny{(3.82)}}	& \shortstack{$4.76^\dagger$  \tiny{(1.45)}}		\\
SD-RetinaNet - $\mathcal{L}_{\text{lp}}$ & \shortstack{$3.29$  \tiny{(1.39)}} 	& \shortstack{$4.04$  \tiny{(1.47)}}	& \shortstack{$6.23^\dagger$  \tiny{(2.16)}}	& \shortstack{$4.84^\dagger$  \tiny{(1.74)}}	& \shortstack{$5.41^\dagger$  \tiny{(2.10)}}	& \shortstack{$6.41^{*\dagger}$  \tiny{(2.51)}}	& \shortstack{$5.53^\dagger$  \tiny{(2.28)}}	& \shortstack{$4.51^{*\dagger}$  \tiny{(2.63)}}	& \shortstack{$4.71^{*\dagger}$  \tiny{(2.23)}}	& \shortstack{$4.74^\dagger$  \tiny{(2.41)}}	& \shortstack{$4.43^\dagger$  \tiny{(4.56)}}	& \shortstack{$4.92^{*\dagger}$  \tiny{(1.54)}}		\\
\rowcolor[rgb]{0.92,0.92,0.92} SD-RetinaNet w. TCCT BB. & \shortstack{$3.22$  \tiny{(1.25)}} 	& \shortstack{$3.96$  \tiny{(1.50)}}	& \shortstack{$6.06$  \tiny{(2.33)}}	& \shortstack{$4.81^\dagger$  \tiny{(1.90)}}	& \shortstack{$5.40^\dagger$  \tiny{(2.40)}}	& \shortstack{$6.33^\dagger$  \tiny{(2.43)}}	& \shortstack{$5.29^\dagger$  \tiny{(1.96)}}	& \shortstack{$4.07^\dagger$  \tiny{(1.60)}}	& \shortstack{$4.39^\dagger$  \tiny{(1.56)}}	& \shortstack{$4.55^\dagger$  \tiny{(2.49)}}	& \shortstack{$4.76^\dagger$  \tiny{(4.64)}}	& \shortstack{$4.80^\dagger$  \tiny{(1.44)}}		\\
\midrule

\multicolumn{13}{c}{\textbf{Proposed methods}} \\
\midrule
\rowcolor[rgb]{0.92,0.92,0.92} \textbf{SD-RetinaNet}  + $\mathcal{L}_{\text{triplet}}$ & \shortstack{$3.24$  \tiny{(1.28)}} 	& \shortstack{$3.99$  \tiny{(1.48)}}	& \shortstack{$\mathbf{5.82}^*$  \tiny{(1.58)}}	& \shortstack{$\mathbf{4.50}$  \tiny{(1.50)}}	& \shortstack{$\mathbf{4.72}^*$  \tiny{(1.74)}}	& \shortstack{$\mathbf{5.42}^*$  \tiny{(1.98)}}	& \shortstack{$\mathbf{4.64}^*$  \tiny{(1.45)}}	& \shortstack{$\mathbf{3.51}^*$  \tiny{(1.16)}}	& \shortstack{$\mathbf{3.72}^*$  \tiny{(1.17)}}	& \shortstack{$\mathbf{3.71}^*$  \tiny{(1.50)}}	& \shortstack{$\mathbf{3.55}^*$  \tiny{(2.35)}}	& \shortstack{$\mathbf{4.26}^*$  \tiny{(1.05)}}		\\
\textbf{SD-RetinaNet} & \shortstack{$\mathbf{3.22}$  \tiny{(1.31)}} 	& \shortstack{$3.88$  \tiny{(1.31)}}	& \shortstack{$6.17^\dagger$  \tiny{(2.33)}}	& \shortstack{$4.64$  \tiny{(1.94)}}	& \shortstack{$5.28^\dagger$  \tiny{(2.17)}}	& \shortstack{$6.06^\dagger$  \tiny{(2.30)}}	& \shortstack{$5.31^\dagger$  \tiny{(1.86)}}	& \shortstack{$4.08^\dagger$  \tiny{(1.64)}}	& \shortstack{$4.35^\dagger$  \tiny{(1.72)}}	& \shortstack{$4.71^\dagger$  \tiny{(2.50)}}	& \shortstack{$4.29^\dagger$  \tiny{(3.71)}}	& \shortstack{$4.73^\dagger$  \tiny{(1.33)}}		\\
\bottomrule
\multicolumn{7}{l}{* indicates statistically significant differences to \emph{SD-RetinaNet}}   \\
\multicolumn{7}{l}{$\dagger$ indicates statistically significant differences to \emph{SD-RetinaNet $ + \mathcal{L}_\text{triplet}$}}                                    \\
\end{tabular}

}
\end{table*}
\begin{table}[tbh]
\centering
\setlength{\extrarowheight}{0pt}
\addtolength{\extrarowheight}{\aboverulesep}
\addtolength{\extrarowheight}{\belowrulesep}
\setlength{\aboverulesep}{0pt}
\setlength{\belowrulesep}{0pt}
\caption{Quantitative lesion segmentation results on the \NVdold{MultiBioMarker}\NVold{internal} dataset including the ablation experiments. The values correspond to the volume-wise dice metric.}
\label{tab:lesions}
\resizebox{\linewidth}{!}{%
\begin{tabular}{l|ccc|c} 

\toprule
\multicolumn{1}{l}{\textbf{Method}}              & \textbf{SRF}       & \textbf{IRF}           & \multicolumn{1}{c}{\textbf{SHRM}} & \textbf{Total}        \\ 
\midrule
\multicolumn{5}{c}{\textbf{Baseline methods}} \\
\midrule
\rowcolor[rgb]{0.922,0.922,0.922} ReLayNet  \cite{2017_Roy}                        & $\shortstack{$0.50^{*\dagger}$  \tiny{(0.29)}}$       & $\shortstack{$0.30^{*\dagger}$  \tiny{(0.29)}}$       & $\shortstack{$0.19^{*\dagger}$  \tiny{(0.25)}}$                   & $0.33$        \\
nnU-Net \cite{2021_Isensee} & $\shortstack{$0.74^{*\dagger}$  \tiny{(0.23)}}$         & $\shortstack{$0.56^{*\dagger}$  \tiny{(0.29)}}$         & $\shortstack{$0.66^{*\dagger}$  \tiny{(0.29)}}$                     & $0.65$          \\
\rowcolor[rgb]{0.922,0.922,0.922} U-Mamba	\cite{2024_Jun}                          & $\shortstack{$0.74^{*\dagger}$  \tiny{(0.23)}}$         & $\shortstack{$0.53^{*\dagger}$  \tiny{(0.28)}}$         & $\shortstack{$0.63^{*\dagger}$  \tiny{(0.29)}}$                     & $0.63$          \\
SwinUNETR \cite{2022_Hatamizadeh}     & $\shortstack{$0.74^{*\dagger}$  \tiny{(0.23)}}$ & $\shortstack{$0.54^{*\dagger}$  \tiny{(0.30)}}$ & $\shortstack{$0.62^{*\dagger}$  \tiny{(0.28)}}$             & $0.63$  \\
\rowcolor[rgb]{0.922,0.922,0.922} TCCT-BP \cite{2024_Tan} \NV{$\ddagger$}                              & $\shortstack{$0.75^{*\dagger}$  \tiny{(0.23)}}$           & $\shortstack{$0.57^{*\dagger}$  \tiny{(0.29)}}$           & $\shortstack{$0.66^{*\dagger}$  \tiny{(0.29)}}$                       & $0.66$            \\
\NVdold{He\etal}\NVold{LBRM} \cite{2021_He} & $\shortstack{$0.79^{*\dagger}$  \tiny{(0.22)}}$          & $\shortstack{$0.56^{*\dagger}$  \tiny{(0.29)}}$          & $\shortstack{$0.70^{*\dagger}$  \tiny{(0.19)}}$                      & $0.68$           \\
\midrule
\multicolumn{5}{c}{\textbf{Semi-supervised baselines}} \\
\midrule

\rowcolor[rgb]{0.922,0.922,0.922} \NVold{LBRM} + $\mathcal{L}_{\text{lp}}$                                       & $\shortstack{$0.79^{*\dagger}$  \tiny{(0.21)}}$    & $\shortstack{$0.58^{*\dagger}$  \tiny{(0.29)}}$    & $\shortstack{$0.67^{*\dagger}$  \tiny{(0.20)}}$                & $0.68$     \\ 
\NVold{ACTION\+\+}\cite{2023_You}                                       & $\shortstack{$0.76^{*\dagger}$  \tiny{(0.25)}}$    & $\shortstack{$0.55^{*\dagger}$  \tiny{(0.31)}}$    & $\shortstack{$0.65^{*\dagger}$  \tiny{(0.24)}}$                & $0.65$     \\ 
\rowcolor[rgb]{0.922,0.922,0.922} \NVold{BE-SemiNet}  \cite{2023_Lu} \NV{$\ddagger$}                                     & $\shortstack{$0.78^{*\dagger}$  \tiny{(0.20)}}$    & $\shortstack{$0.57^{*\dagger}$  \tiny{(0.27)}}$    & $\shortstack{$0.63^{*\dagger}$  \tiny{(0.19)}}$                & $0.66$     \\ 
SD-LayerNet \cite{2022_Fazekas_CONF} \NV{$\ddagger$}            & $\shortstack{$0.79^{*\dagger}$  \tiny{(0.22)}}$    & $\shortstack{$0.56^{*\dagger}$  \tiny{(0.29)}}$    & $\shortstack{$0.68^{*\dagger}$  \tiny{(0.19)}}$                & $0.68$     \\ 

\midrule
\multicolumn{5}{c}{\textbf{Ablation studies}} \\
\midrule
\rowcolor[rgb]{0.922,0.922,0.922} SD-RetinaNet w/o recon & $\shortstack{$0.79^{*\dagger}$  \tiny{(0.21)}}$  & $\shortstack{$0.59^{*\dagger}$  \tiny{(0.28)}}$  & $\shortstack{$0.66^{*\dagger}$  \tiny{(0.21)}}$              & $0.68$   \\
\NVold{SD-RetinaNet - $\mathcal{L}_{\text{to}}$}                                     & $\shortstack{$0.80$  \tiny{(0.21)}}$     & $\shortstack{$0.60$  \tiny{(0.27)}}$     & $\shortstack{$0.68^{*\dagger}$  \tiny{(0.19)}}$                 & $0.69$      \\ 
\rowcolor[rgb]{0.922,0.922,0.922} \NVold{SD-RetinaNet - $\mathcal{L}_{\text{bc}}$}                                     & $\shortstack{$0.81$  \tiny{(0.20)}}$     & $\shortstack{$0.58^{*\dagger}$  \tiny{(0.27)}}$     & $\shortstack{$0.70^{*\dagger}$  \tiny{(0.20)}}$                 & $0.69$      \\ 
SD-RetinaNet - $\mathcal{L}_{\text{lp}}$                                     & $\shortstack{$0.80$  \tiny{(0.21)}}$     & $\shortstack{$0.58^{*\dagger}$  \tiny{(0.29)}}$     & $\shortstack{$0.70^{*\dagger}$  \tiny{(0.19)}}$                 & $0.69$      \\ 
\rowcolor[rgb]{0.922,0.922,0.922} SD-RetinaNet w. TCCT BB.                                & $\shortstack{$0.80$  \tiny{(0.20)}}$     & $\shortstack{$0.59$  \tiny{(0.27)}}$     & $\shortstack{$0.65^{*\dagger}$  \tiny{(0.21)}}$                 & $0.68$      \\ 
\midrule
\multicolumn{5}{c}{\textbf{Proposed methods}} \\
\midrule
\textbf{SD-RetinaNet} + $\mathcal{L}_{\text{triplet}}$                       & $\shortstack{$\mathbf{0.81}$  \tiny{(0.19)}}$    & $\shortstack{$\mathbf{0.62}$  \tiny{(0.25)}}$    & $\shortstack{$\mathbf{0.73}$  \tiny{(0.18)}}$                & $\mathbf{0.72}$     \\
\rowcolor[rgb]{0.922,0.922,0.922} \textbf{SD-RetinaNet}  & $\shortstack{$0.81$  \tiny{(0.20)}}$           & $\shortstack{$0.60$  \tiny{(0.28)}}$           & $\shortstack{$0.68^{*\dagger}$  \tiny{(0.20)}}$                       & $0.70$            \\
\multicolumn{5}{l}{* indicates statistically significant differences to \emph{SD-RetinaNet}}   \\
\multicolumn{5}{l}{$\dagger$ indicates statistically significant differences to \emph{SD-RetinaNet $ + \mathcal{L}_\text{triplet}$}}  \\
\multicolumn{5}{l}{\NV{$\ddagger$ Method originally designed for layer segmentation only; Adapted with additional}}\\
\multicolumn{5}{l}{\NV{output channels to enable lesion segmentation for this comparison.}} \\
\bottomrule
\end{tabular}

}
\end{table}

\begin{figure*}[tbh]
    \centering
    \setlength{\tabcolsep}{1pt}
    
\resizebox{\linewidth}{!}{%
    \begin{tabular}{cccccc}
            \multicolumn{1}{c}{\small{B-scan}} & \multicolumn{1}{c}{\small{Expert Annotation}} & \multicolumn{1}{c}{\small{\NVdold{He\etal}\NVold{LBRM} \cite{2021_He}}} & \multicolumn{1}{c}{\small{TCCT-BP \cite{2024_Tan}}} & \multicolumn{1}{c}{\small{SD-RetinaNet + $\mathcal{L}_{\text{triplet}}$}}& \multicolumn{1}{c}{\small{SD-RetinaNet  }} \\
        \includegraphics[width=0.2\textwidth]{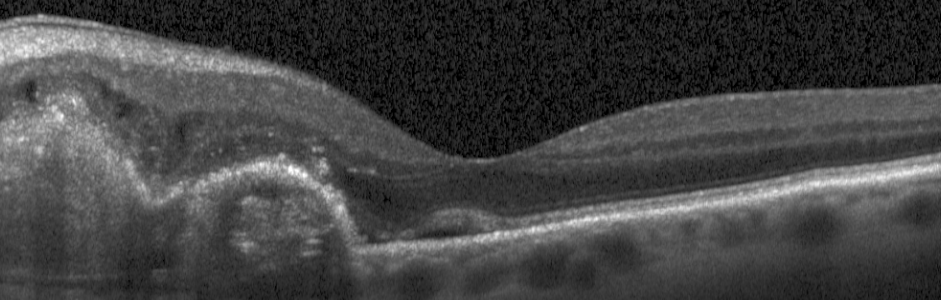} &
        \includegraphics[width=0.2\textwidth]{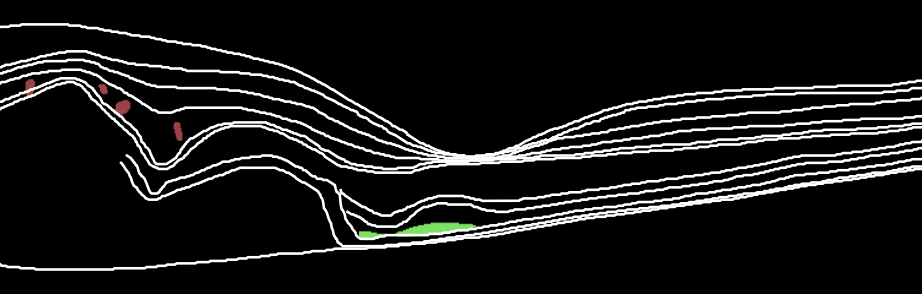} &
        \includegraphics[width=0.2\textwidth]{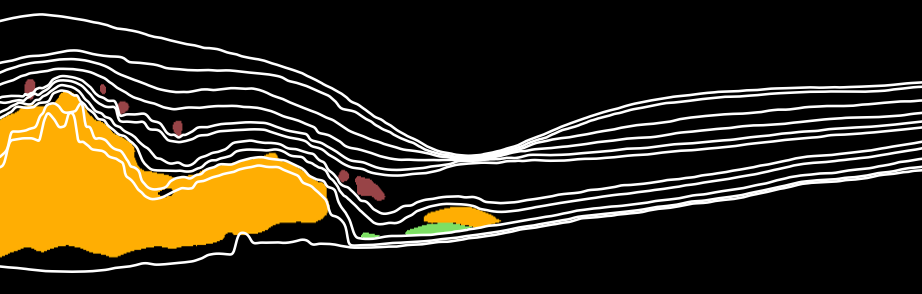} &
        \includegraphics[width=0.2\textwidth]{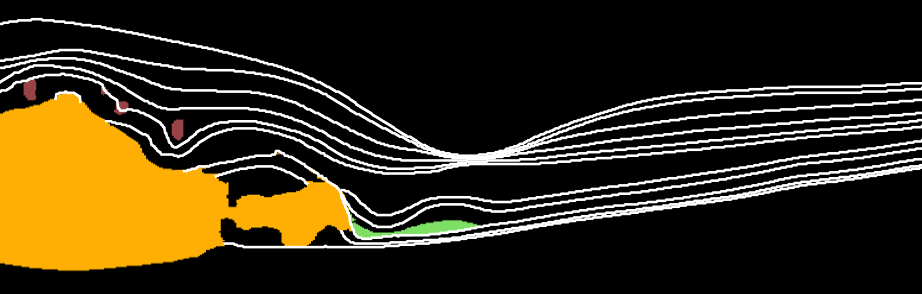} &
        \includegraphics[width=0.2\textwidth]{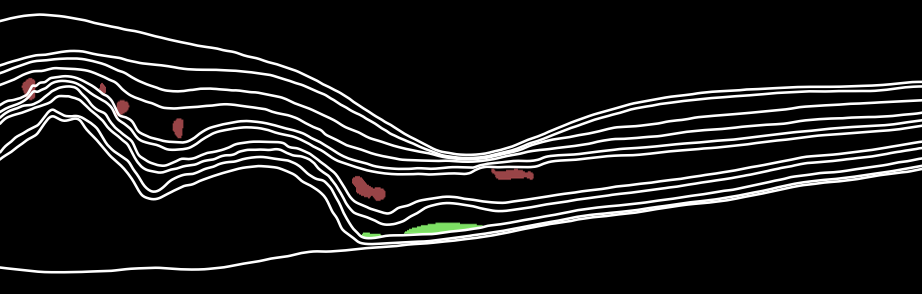} &
        \includegraphics[width=0.2\textwidth]{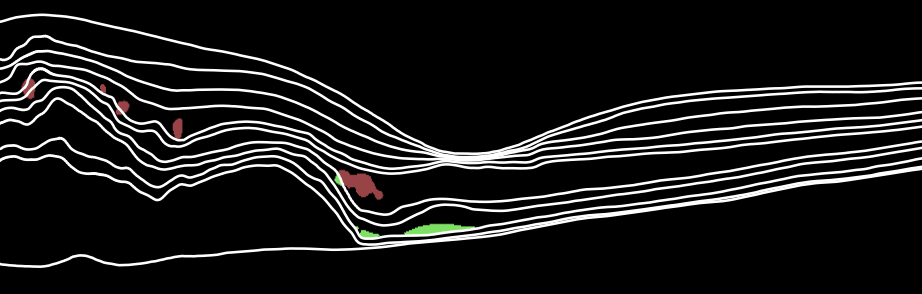} \\
        \includegraphics[width=0.2\textwidth]{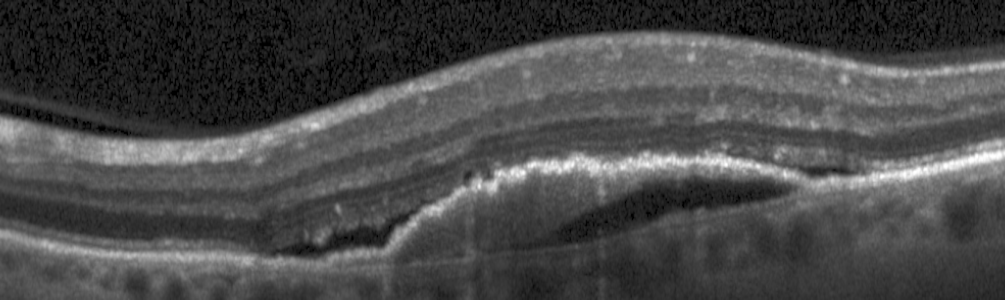} &
        \includegraphics[width=0.2\textwidth]{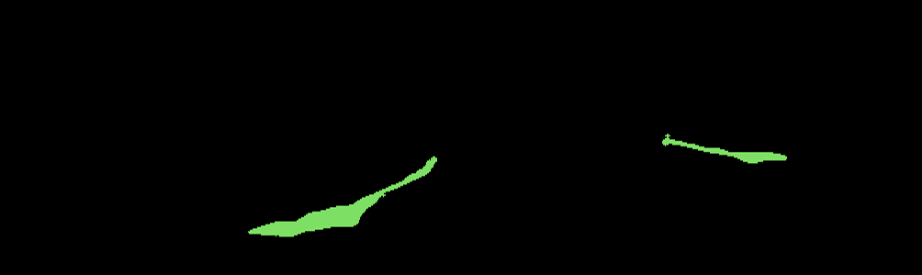} &
        \includegraphics[width=0.2\textwidth]{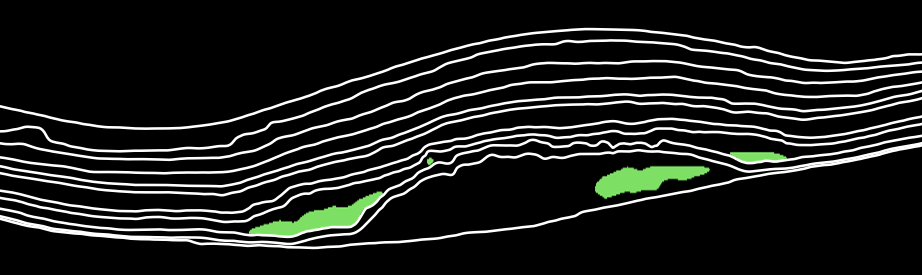} &
        \includegraphics[width=0.2\textwidth]{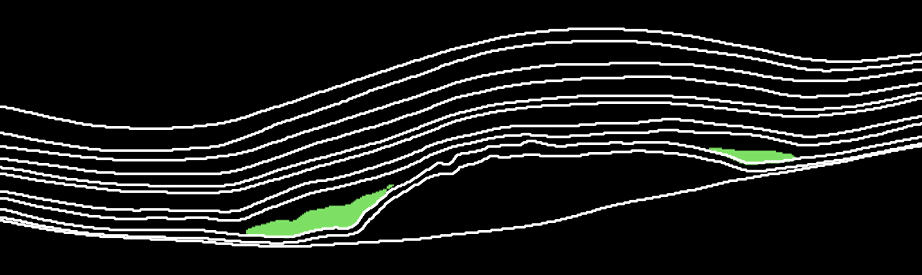} &
        \includegraphics[width=0.2\textwidth]{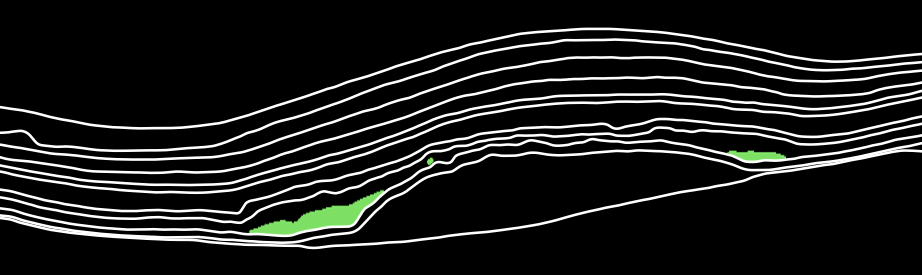} &
        \includegraphics[width=0.2\textwidth]{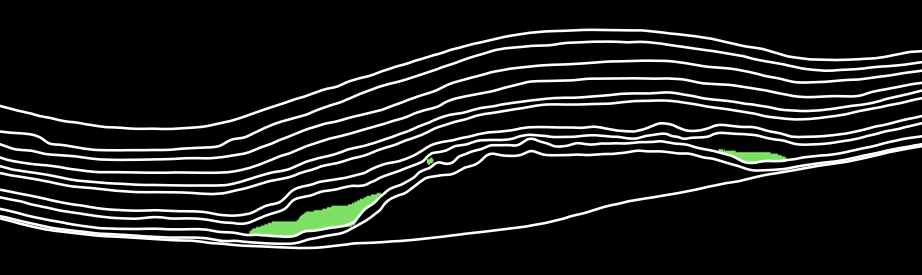} \\
        \includegraphics[width=0.2\textwidth]{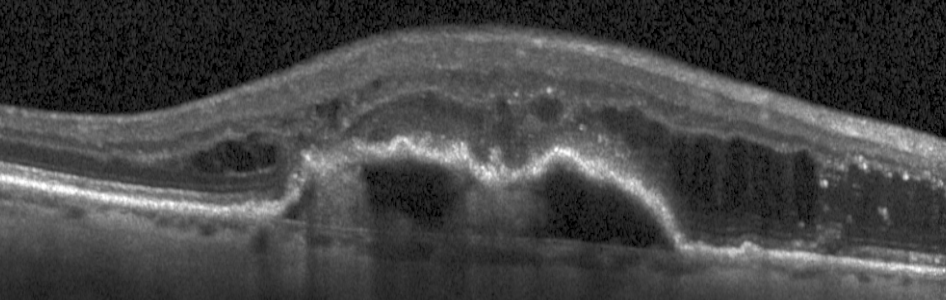} &
        \includegraphics[width=0.2\textwidth]{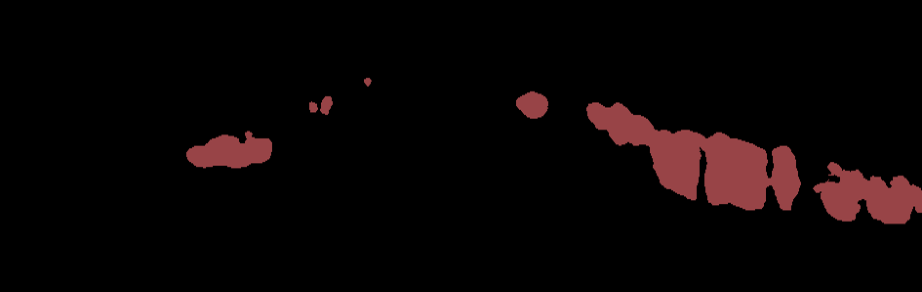} &
        \includegraphics[width=0.2\textwidth]{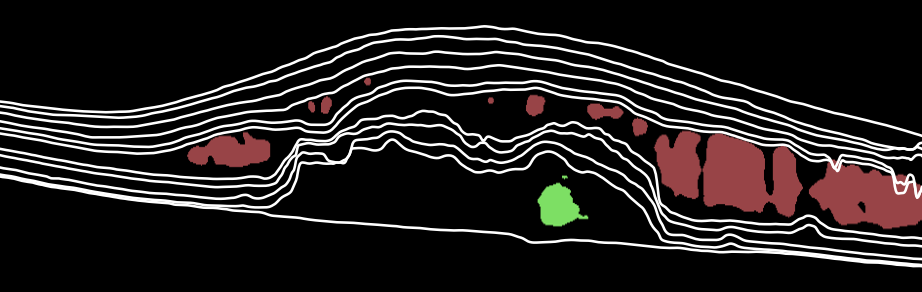} &
        \includegraphics[width=0.2\textwidth]{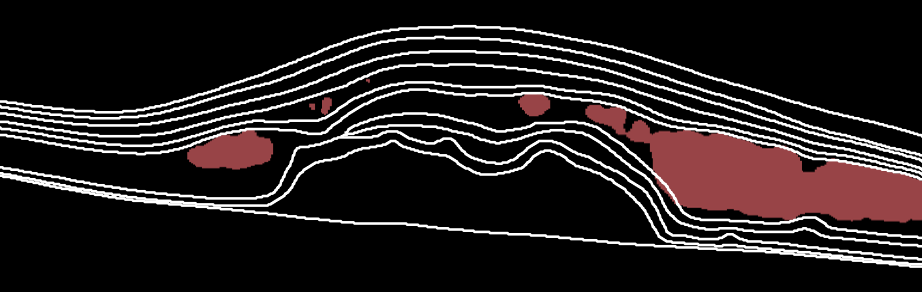} &
        \includegraphics[width=0.2\textwidth]{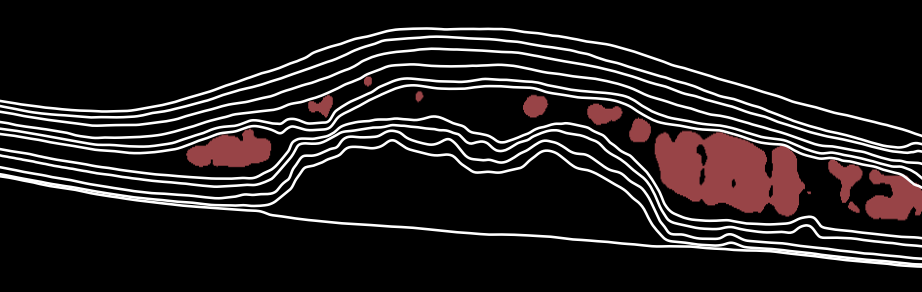} &
        \includegraphics[width=0.2\textwidth]{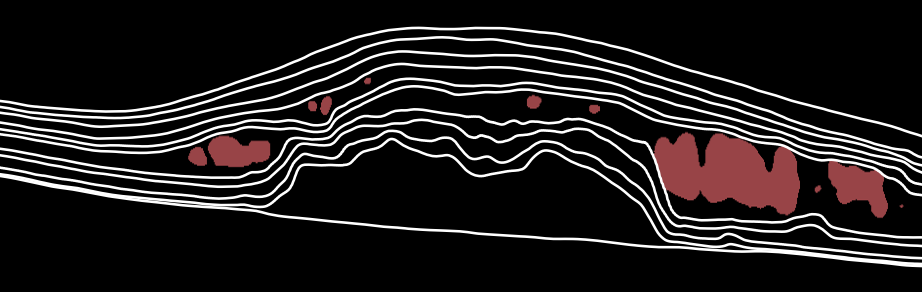} \\
        \includegraphics[width=0.2\textwidth]{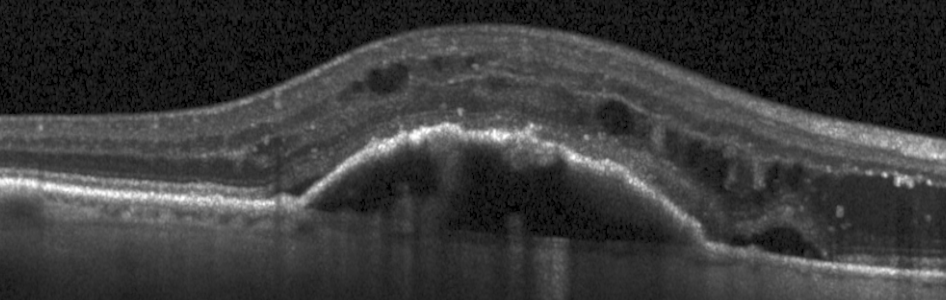} &
        \includegraphics[width=0.2\textwidth]{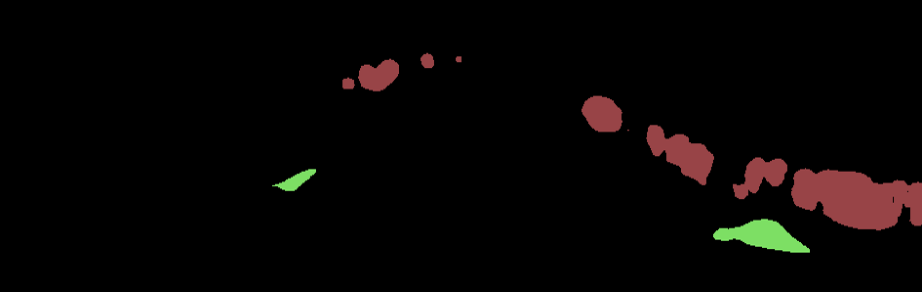} &
        \includegraphics[width=0.2\textwidth]{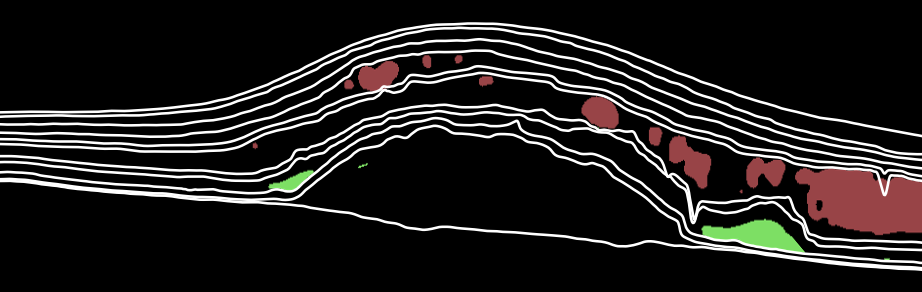} &
        \includegraphics[width=0.2\textwidth]{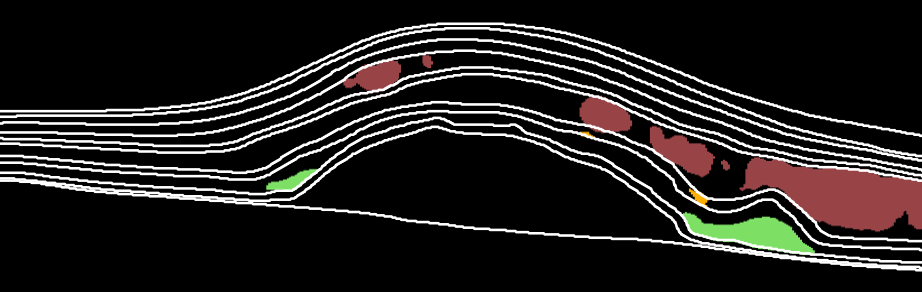} &
        \includegraphics[width=0.2\textwidth]{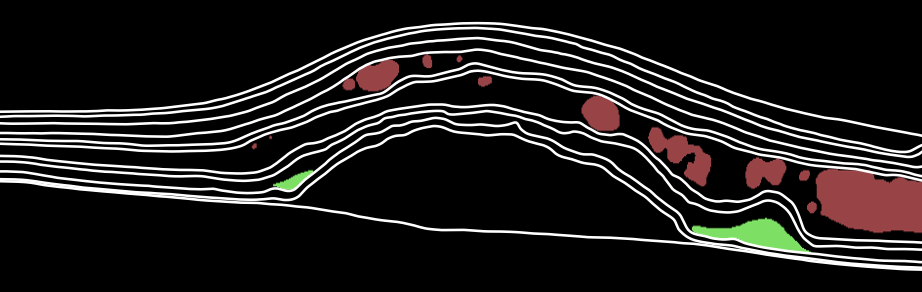} &
        \includegraphics[width=0.2\textwidth]{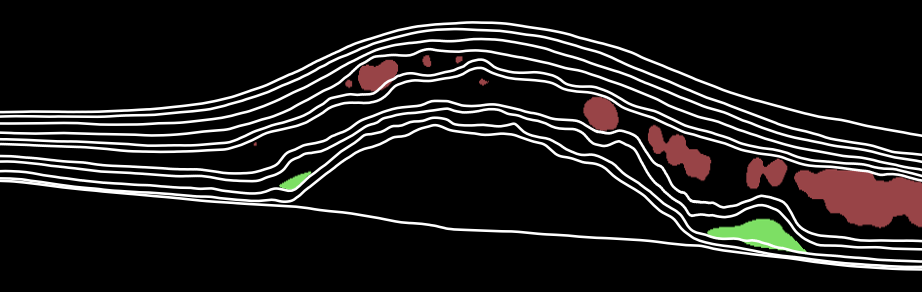} \\
        \includegraphics[width=0.2\textwidth]{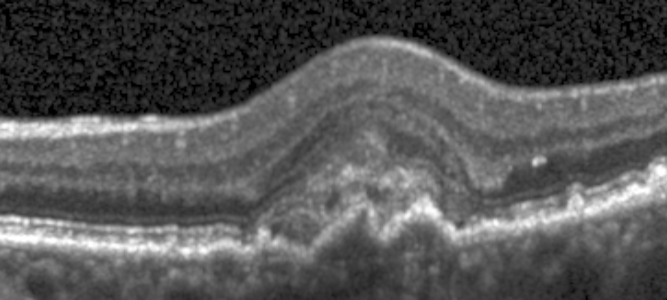} &
        \includegraphics[width=0.2\textwidth]{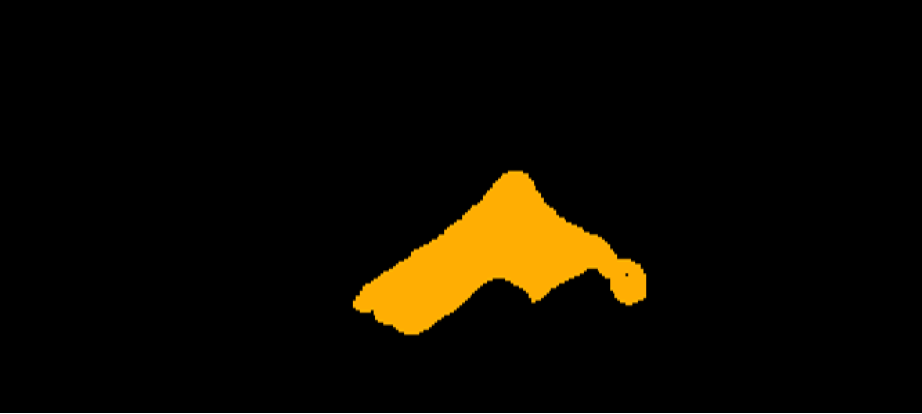} &
        \includegraphics[width=0.2\textwidth]{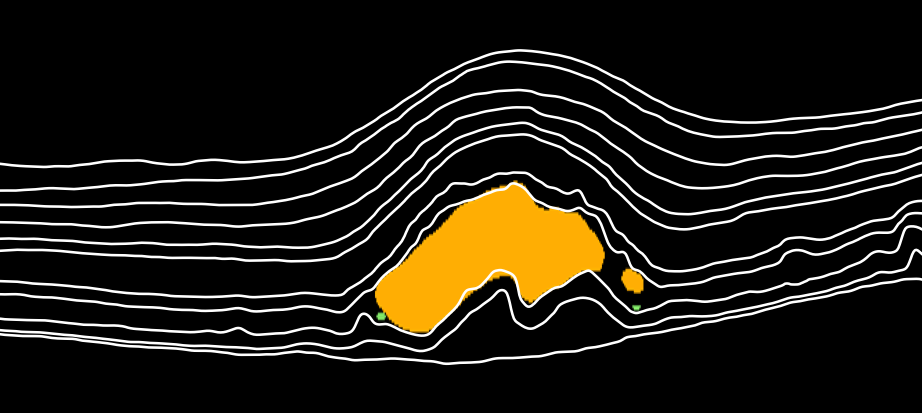} &
        \includegraphics[width=0.2\textwidth]{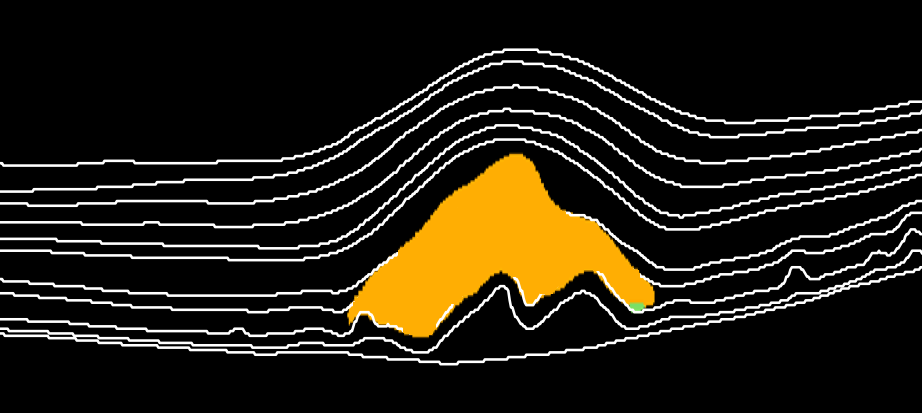} &
        \includegraphics[width=0.2\textwidth]{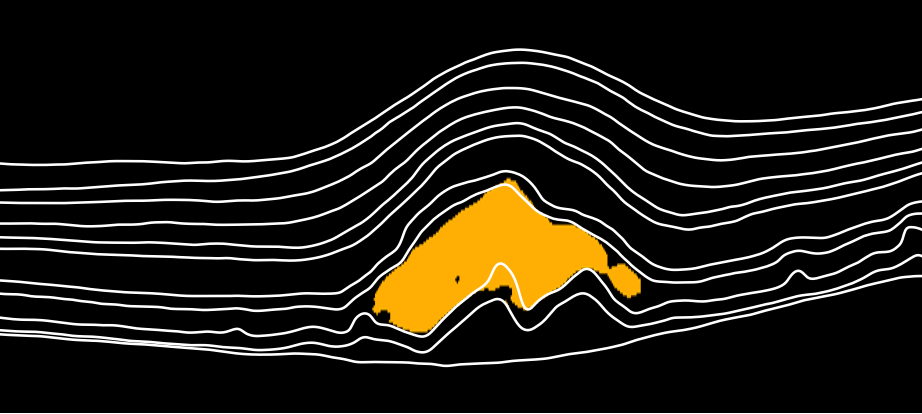} &
        \includegraphics[width=0.2\textwidth]{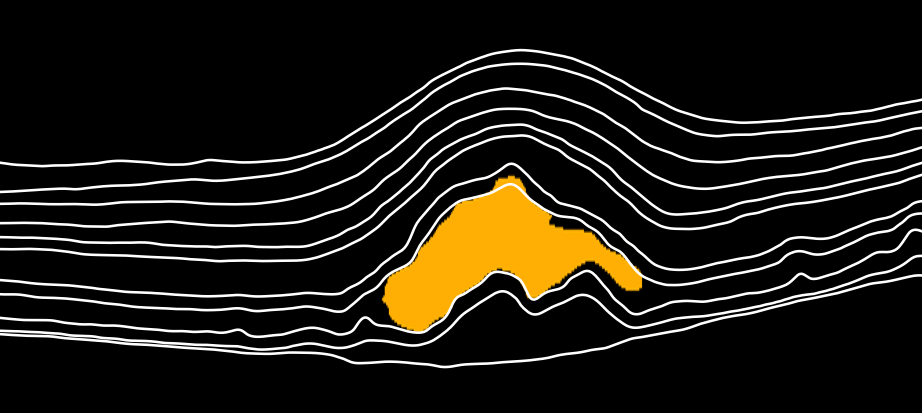} \\
        \includegraphics[width=0.2\textwidth]{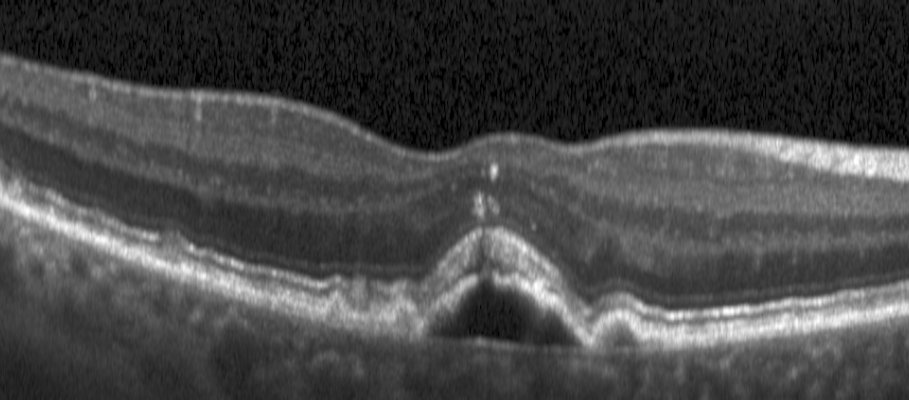} &
        \includegraphics[width=0.2\textwidth]{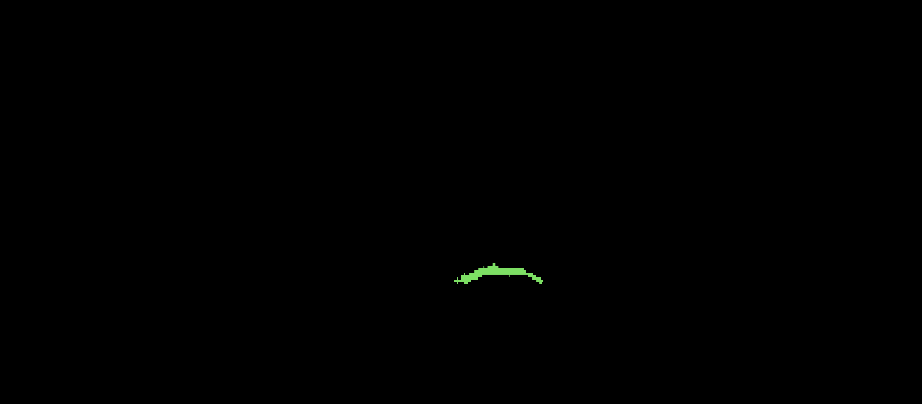} &
        \includegraphics[width=0.2\textwidth]{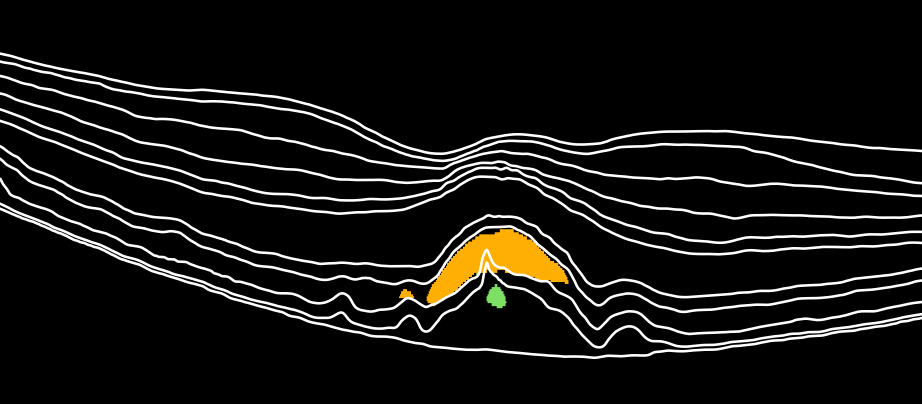} &
        \includegraphics[width=0.2\textwidth]{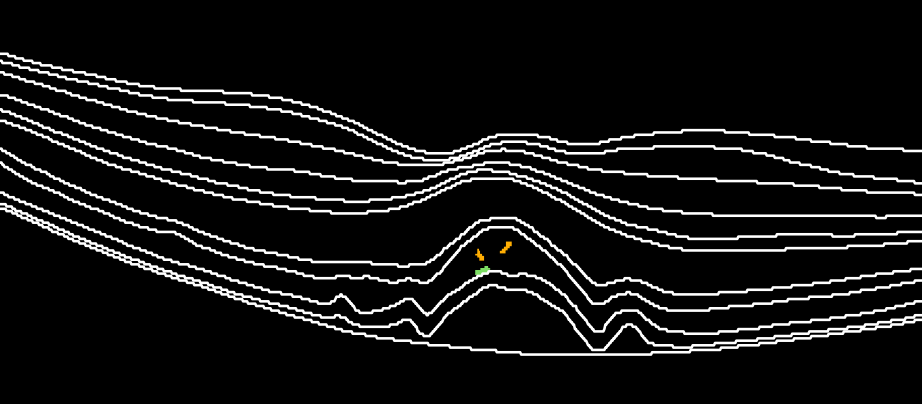} &
        \includegraphics[width=0.2\textwidth]{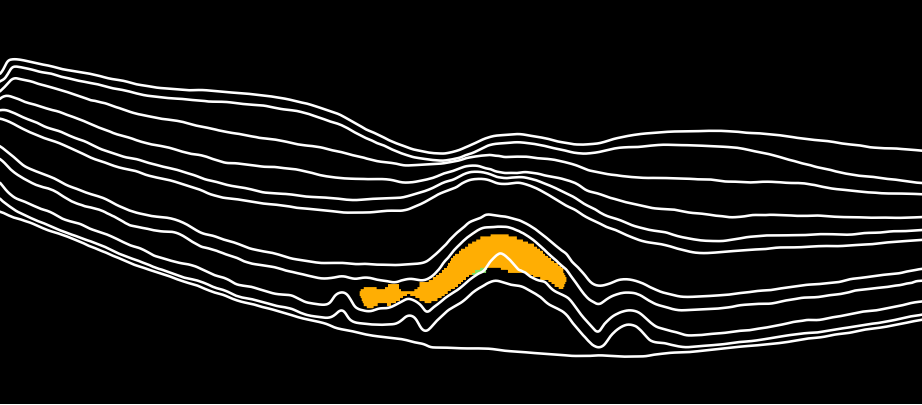} &
        \includegraphics[width=0.2\textwidth]{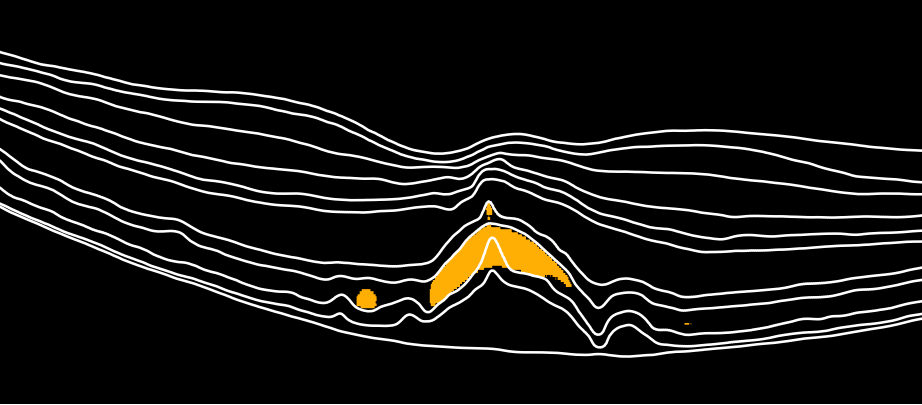} \\
    \end{tabular}
    }
    \caption{Qualitative results in the \NVdold{MultiBioMarker}\NVold{internal} test set with the original B-scan, expert annotations, predictions of the two best baseline methods \NVdold{He\etal}\NVold{LBRM}\cite{2021_He}, TCCT-BP\cite{2024_Tan} and ours. White: layer boundaries (not always available in the dataset); Green: SRF; Red: IRF; Orange: SHRM.}
    \label{fig:mb_qualitative}
\end{figure*}

\subsection{Experimental settings}
\subsubsection{Training details}
\NVdold{For the baseline studies on the MultiBioMarker dataset, we use a batch size of 40 samples (20 labeled and 20 unlabeled), with AdamW optimizer (learning rate $10^{-4}$). For the partial annotation studies, the batch size is the same, but no unlabeled samples are used.
The training was stabilized with a gradient clipping of maximum norm 2 and lasted for 300 epochs.}\NVold{All models were trained using the AdamW optimizer with a learning rate of $10^{-4}$. Training was stabilized with gradient clipping using a maximum norm of 2 and lasted for 300 epochs.} The model with the \NVold{combined} lowest validation mean average distance (MAD) \NVold{on the layers} and \NVold{lowest} Dice loss\NVold{ on the lesions} was selected. \NVold{For the forced disentanglement experiments the following affine transformations were carried out on $I_a$ with a probability (p) of 1.0: rotation (Range: -0.2 - 0.2rad), shearing (vertical and horizontal: -0.2 - 0.2), translation (x-axis: -0.3 - 0.3, y-axis: 0), scaling: (x-axis: -0.1 - 0.1, y-axis: 0). On $I_s$ the following style transformations were applied: random gaussian noise ($p = 0.1, \mu = 0.0, \sigma = 0.1$), random rician noise ($p = 0.1, \mu = 0.0, \sigma = 1.0$), random gaussian noise ($p = 0.1, \mu = 0.0, \sigma = 0.1$), random gaussian smoothing ($p = 0.1, \sigma_x = 0.25 - 1.5, \sigma_y = 0.25 - 1.5$), random gaussian sharpening ($p = 0.1, \sigma_x = 0.5 - 1.0, \sigma_y = 0.5 - 1.0$). The same random transformations were applied as data augmentation on the whole training dataset (both labeled and unlabeled samples).}

\NVold{For experiments on the internal dataset (using 3-fold cross-validation), semi-supervised training utilized a batch size of 40. Each batch consisted of 20 labeled samples drawn from the respective internal training split of the current cross-validation fold (derived from the 85 annotated volumes) and 20 unlabeled samples drawn from the separate set of 387 unannotated volumes. This 20/20 batch composition was applied consistently to our proposed method, relevant ablation studies involving semi-supervision, and the semi-supervised baseline comparisons.}

\NVold{For the partial annotation experiments, the batch size was also 40 but consisted entirely of labeled samples drawn proportionally from the HCMS and RETOUCH training sets. No unlabeled volumes were used in this specific experimental setting. For these experiments, the model checkpoint after 150 training epochs was used for evaluation, as no combined validation set was readily available.}

The training was carried out in a mixed-precision environment on an Nvidia DGX server on a single A100 (40GB) GPU, using Ubuntu 20.04, Python 3.8.12, and PyTorch 1.11.0. The training took 4 hours on this setup. The source code for our implementation can be found at \url{http://github.com/ABotond/SD-RetinaNet}.
\subsubsection{Hyperparameter tuning}
Owing to the disentangled design of the proposed method and the incorporated anatomical priors, the model contains numerous hyperparameters for the weights of the loss terms. To streamline the training process, we used the SoftAdapt method\cite{2019_Heydari}. \NVold{This eliminates the need for manual hyperparameter tuning of these weights, by utilizing} \NVdold{This algorithm utilizes }live statistics for multi-part loss functions, monitoring the performance and magnitude of each loss term and adjusting their weights based on individual performance. Consequently, this approach effectively reduces the hyperparameters $\lambda_{1-10}$ into a single hyperparameter, $\beta$. This parameter regulates whether SoftAdapt assigns more weight to the worst-performing component of the loss function ($\beta > 0$) or to the best-performing ones ($\beta < 0$).
We observed no statistically significant performance differences for $\beta$ within the range $\left[0.05, \ldots, 0.5\right]$, and thus adopted the default value $\beta = 0.1$.

For the $\delta$ parameter in $\mathcal{L}_{\text{bc}}$, all values within the range $\left[11, \ldots, 21\right]$ resulted in similar performance, leading us to select $\delta = 15$. For $\mathcal{L}_{\text{KL}}$, we set $\sigma = 0.5$ without further tuning, following the approach in \cite{2022_Fazekas_CONF}.

\NVd{Evaluation metrics}

\NVd{We used the Dice score to evaluate the lesion segmentation performance and Mean Absolute Difference (MAD) for evaluating the layer segmentation performance. MAD offers an unbiased evaluation by treating all segmentation errors uniformly, regardless of the thickness of the layer, as opposed to the Dice loss. MAD also provides a balanced assessment that accounts for both minor and major discrepancies between automated and manual delineations.}

\NVd{We performed a statistical analysis using a linear mixed-effect model \cite{2014_Bates}, to account for the interdependence of B-scans within the same volume.}

\section{Results}
\subsection{\NV{Main performance evaluation}\NVd{Baseline and ablation studies}}
\label{section:baseline_results}
Based on the presented layer segmentation results, our proposed models demonstrate statistically significant superior performance compared to all of the baseline methods (Table~\ref{tab:mb_layer_results}). Specifically, the configuration without forced disentanglement, 'SD-RetinaNet', improves performance by 6\% compared to the \NVold{overall} best-performing baseline, while the second configuration, 'SD-RetinaNet + $\mathcal{L}_{\text{triplet}}$' achieves 15\% improvement.

\NVb{The ablation study showed that both the lesion position constraint $\mathcal{L}_{\text{lp}}$, as well as the reconstruction has a statistically significant positive effect on the performance. While the addition of $\mathcal{L}_{\text{lp}}$ to the He\etal~ lead to some improvements in the layer segmentation task, the results suggest that this is more effective when combined with the other semi-supervised factors. We found that using a state-of-the-art hybrid transformer - CNN architecture (TCCT) instead of the fully convolutional EfficientNet-b4 feature encoder lead to an inferior layer segmentation performance.}

In lesion segmentation, both of our proposed models demonstrate superior performance in the Dice score compared to all baselines and ablation study methods (Table~\ref{tab:lesions}).
In most cases, these improvements were statistically significant, highlighting the importance and benefits of incorporating topological and anatomical constraints during the training process. \NVb{The ablation study in the lesion segmentation task also showed, that the removal of the $\mathcal{L}_\text{lp}$ factor has lead to statistically significantly worse performance in the SRF segmentation. Underlining our findings in the layer segmentation tasks, using the TCCT backbone instead of a fully convolutional encoder has lead to a performance decrease in the lesion segmentation as well.}

\NV{The qualitative results in Figure~\ref{fig:mb_qualitative} highlight the practical impact of our biomarker topology engine by directly contrasting our method against baselines on challenging cases. For instance, the LBRM baseline\cite{2021_He} frequently produces anatomically implausible segmentations, such as placing SHRM in an incorrect location (first row) or SRF below the RPE (second and third rows). In these same cases, our method correctly confines the lesions to their valid topological positions, demonstrating the effectiveness of the enforced constraints. While both of our proposed configurations perform well, the variant with the forced disentanglement loss ($\mathcal{L}_{\text{triplet}}$) appears to produce the most robust predictions. Finally, the last row presents a representative failure case, where a subtle hyperreflective signal leads our model to produce a false positive SHRM prediction. While the segmentation is still localized in an anatomically plausible area, this underscores the challenging nature of this real-world clinical dataset.}

\NVd{The qualitative results (Figure~\ref{fig:mb_qualitative}) reveal the expected limitations of the baseline methods. For example, they segment SHRM in an anatomically implausible location (first row), as well as SRF between the RPE and BM (He\etal; second, third and fourth row). In contrast, both of our proposed model configurations segmented lesions only in their plausible locations, however, the variant with $\mathcal{L}_{\text{triplet}}$ has seemingly more robust predictions.}

\subsection{Partial annotation generalization}

\begin{table*}[tbh]
\centering
\setlength{\extrarowheight}{0pt}
\addtolength{\extrarowheight}{\aboverulesep}
\addtolength{\extrarowheight}{\belowrulesep}
\caption{Quantitative results of the partial annotation experiments on the external DukeDME dataset, including the ablation experiments. The layer segmentation values correspond to the mean absolute distance (MAD) metric in terms of \micron, while the fluid segmentation results are presented with the Dice score.}
\label{tab:duke_layer_results}
\resizebox{\linewidth}{!}{%
\begin{tabular}{l|cccccccc|c|c} 
\toprule
\multicolumn{1}{l}{Method}                         & ILM                  & RNFL/GCL                 & IPL/INL                 & INL/OPL                 & OPL/ONL                 & IS/OS                 & OB-RPE                 & \multicolumn{1}{c}{BM} & \multicolumn{1}{l}{\textbf{Total}}  & \textbf{Fluid}        \\ 
\midrule
\midrule
\multicolumn{11}{c}{\textbf{Baseline methods trained on HCMS}} \\
\midrule
U-Mamba	\cite{2024_Jun}                     & \shortstack{$5.47$ \tiny{(1.20)}}       & \shortstack{$17.26$ \tiny{(9.33)}}       & \shortstack{$30.87$ \tiny{(15.41)}}       & \shortstack{$26.90$ \tiny{(13.37)}}       & \shortstack{$23.58$ \tiny{(9.82)}}       & \shortstack{$4.85$ \tiny{(1.33)}}       & \shortstack{$8.68$ \tiny{(3.69)}}       & \shortstack{$5.51$ \tiny{(0.87)}}          & \shortstack{$14.11$ \tiny{(5.22)}}        & \shortstack{$0.33$$^{*\dagger}$ \tiny{(0.27)}}        \\
\rowcolor[rgb]{0.922,0.922,0.922}TCCT-BP  \cite{2024_Tan} \NV{$\ddagger$}                    & \shortstack{$5.63$ \tiny{(1.22)}}       & \shortstack{$14.80$ \tiny{(6.13)}}       & \shortstack{$28.95$ \tiny{(13.92)}}       & \shortstack{$26.51$ \tiny{(13.79)}}       & \shortstack{$23.16$ \tiny{(9.45)}}       & \shortstack{$4.85$ \tiny{(1.22)}}       & \shortstack{$8.34$ \tiny{(3.50)}}       & \shortstack{$5.38$ \tiny{(0.91)}}          & \shortstack{$13.50$ \tiny{(4.68)}}& \emph{N/A}        \\
LBRM \cite{2021_He}                     & \shortstack{$5.02$ \tiny{(1.00)}}       & \shortstack{$12.01$ \tiny{(6.77)}}       & \shortstack{$28.15$ \tiny{(19.09)}}       & \shortstack{$28.52$ \tiny{(15.68)}}       & \shortstack{$22.53$ \tiny{(9.48)}}       & \shortstack{$4.85$ \tiny{(0.90)}}       & \shortstack{$10.05$ \tiny{(3.75)}}       & \shortstack{$5.54$ \tiny{(0.93)}}          & \shortstack{$13.06$ \tiny{(5.27)}} & \shortstack{$0.39$$^{*\dagger}$ \tiny{(0.21)}}        \\

\midrule
\midrule

\multicolumn{10}{c}{\textbf{Semi-supervised models trained on HCMS + RETOUCH}} \\
\midrule
\rowcolor[rgb]{0.922,0.922,0.922} SD-LayerNet \cite{2022_Fazekas_CONF} \NV{$\ddagger$}                      & \shortstack{$4.83$ \tiny{(1.64)}}       & \shortstack{$7.85$ \tiny{(1.70)}}       & \shortstack{$10.49$$^{*\dagger}$ \tiny{(3.27)}}       & \shortstack{$12.56$$^{*\dagger}$  \tiny{(5.18)}}      & \shortstack{$21.16$$^{*\dagger}$ \tiny{(12.32)}}       & \shortstack{$5.92$$^{*}$  \tiny{(1.30)}}      & \shortstack{$7.99$ \tiny{(2.62)}}       & \shortstack{$7.25$ \tiny{(2.69)}}          & \shortstack{$9.10$$^{*\dagger}$ \tiny{(2.40)}}  & \shortstack{$0.40$ \tiny{(0.23)}}       \\
Action\+\+\cite{2023_You}                     & \shortstack{$5.22$  \tiny{(1.19)}}       & \shortstack{$10.20$  \tiny{(3.09)}}       & \shortstack{$16.17$$^{*\dagger}$  \tiny{(5.43)}}       & \shortstack{$15.04$$^{*\dagger}$  \tiny{(5.60)}}       & \shortstack{$21.02$$^{*\dagger}$  \tiny{(7.24)}}       & \shortstack{$4.58$  \tiny{(0.62)}}       & \shortstack{$8.55$  \tiny{(3.92)}}       & \shortstack{$5.01$  \tiny{(0.76)}}          & \shortstack{$9.96$$^{*\dagger}$  \tiny{(2.10)}}  &    \shortstack{$0.41$  \tiny{(0.29)}}    \\
\rowcolor[rgb]{0.922,0.922,0.922} BE-Seminet\+\+\cite{2023_Lu} \NV{$\ddagger$}                    & \shortstack{$5.48$  \tiny{(1.26)}}       & \shortstack{$11.96$$^{*\dagger}$  \tiny{(4.13)}}       & \shortstack{$21.51$$^{*\dagger}$  \tiny{(9.34)}}       & \shortstack{$18.97$$^{*\dagger}$  \tiny{(8.62)}}       & \shortstack{$18.09$$^{*\dagger}$  \tiny{(6.08)}}       & \shortstack{$4.81$  \tiny{(0.88)}}       & \shortstack{$8.19$  \tiny{(3.57)}}       & \shortstack{$6.31$  \tiny{(2.10)}}          & \shortstack{$11.02$$^{*\dagger}$  \tiny{(3.08)}}        & \shortstack{$0.43$  \tiny{(0.31)}} \\
LBRM + $\mathcal{L}_\text{lp}$                       & \shortstack{$\mathbf{4.47}$ \tiny{(1.09)}}      & \shortstack{$7.43$ \tiny{(1.89)}}      & \shortstack{$8.65$ \tiny{(2.41)}}      & \shortstack{$11.29$ \tiny{(3.72)}}      & \shortstack{$21.52$$^{*\dagger}$ \tiny{(11.60)}}      & \shortstack{$\mathbf{4.39}$$^{*\dagger}$ \tiny{(0.96)}}      & \shortstack{$8.84$ \tiny{(2.88)}}      & \shortstack{$\mathbf{4.77}$$^{*\dagger}$ \tiny{(0.95)}}         & \shortstack{$8.36$$^{*\dagger}$ \tiny{(1.75)}}     & \shortstack{$0.44$ \tiny{(0.22)}}   \\ 
\midrule
\rowcolor[rgb]{0.922,0.922,0.922}  SD-RetinaNet - $\mathcal{L}_\text{rec}$   & \shortstack{$4.75$ \tiny{(1.52)}}    & \shortstack{$7.55$ \tiny{(2.81)}}    & \shortstack{$8.23$ \tiny{(2.44)}}    & \shortstack{$10.78$ \tiny{(3.14)}}    & \shortstack{$16.93$$^{*\dagger}$ \tiny{(7.82)}}    & \shortstack{$7.71$$^{*\dagger}$ \tiny{(2.79)}}    & \shortstack{$8.07$ \tiny{(2.58)}}    & \shortstack{$6.02$$^{*\dagger}$ \tiny{(1.43)}}       & \shortstack{$8.21$$^{*\dagger}$ \tiny{(1.68)}}  & \shortstack{$0.41$ \tiny{(0.25)}}    \\
 
 SD-RetinaNet - $\mathcal{L}_\text{lp}$                     & \shortstack{$4.63$ \tiny{(1.12)}}      & \shortstack{$7.58$ \tiny{(1.72)}}      & \shortstack{$8.23$ \tiny{(2.58)}}      & \shortstack{$10.66$ \tiny{(2.68)}}      & \shortstack{$20.19$$^{*\dagger}$ \tiny{(8.55)}}      & \shortstack{$6.84$$^{*\dagger}$ \tiny{(2.11)}}      & \shortstack{$7.81$ \tiny{(2.92)}}      & \shortstack{$5.34$$^{*\dagger}$   \tiny{(2.11)}}       & \shortstack{$8.31$$^{*\dagger}$ \tiny{(1.75)}}  & \shortstack{$0.43$$^{*\dagger}$ \tiny{(0.24)}}      \\ 
\midrule
\rowcolor[rgb]{0.922,0.922,0.922} \textbf{SD-RetinaNet} + $\mathcal{L}_\text{triplet}$       & \shortstack{$4.87$ \tiny{(1.43)}}      & \shortstack{$\mathbf{7.17}$ \tiny{(1.67)}}      & \shortstack{$8.27$ \tiny{(2.48)}}      & \shortstack{$\mathbf{9.16}$ \tiny{(3.55)}}      & \shortstack{$\mathbf{11.15}$ \tiny{(3.18)}}      & \shortstack{$5.47$ \tiny{(2.62)}}      & \shortstack{$\mathbf{6.98}$$^{*}$ \tiny{(2.44)}}      & \shortstack{$5.19$$^{*}$  \tiny{(0.91)}}        & \shortstack{$\mathbf{6.90}$ \tiny{(1.51)}}$^{*}$  & \shortstack{$0.48$ \tiny{(0.21)}}      \\
  \textbf{SD-RetinaNet}    & \shortstack{$4.62$ \tiny{(1.21)}}             & \shortstack{$7.85$ \tiny{(1.65)}}             & \shortstack{$\mathbf{7.87}$ \tiny{(2.02)}}             & \shortstack{$10.37$ \tiny{(3.64)}}             & \shortstack{$12.51$ \tiny{(4.70)}}             & \shortstack{$4.95$ \tiny{(1.18)}}             & \shortstack{$8.26$ \tiny{(2.73)}}             & \shortstack{$5.56$ \tiny{(4.55)}}                & \shortstack{$7.54$ \tiny{(1.74)}}  & \shortstack{$0.52$ \tiny{(0.25)}}             \\

\midrule
\multicolumn{10}{l}{\scriptsize{* indicates statistically significant differences to \emph{SD-RetinaNet}}}   \\
\multicolumn{10}{l}{\scriptsize{$\dagger$ indicates statistically significant differences to \emph{SD-RetinaNet} $ + \mathcal{L}_\text{triplet}$}} \\
\multicolumn{10}{l}{\NV{$\ddagger$ Method originally designed for layer segmentation only; Adapted with additional output channels to enable lesion segmentation for this comparison.}}\\
\bottomrule
\end{tabular}
}
\end{table*}

\begin{figure*}[tbh]
    \centering
    \setlength{\tabcolsep}{1pt}
    
\resizebox{\linewidth}{!}{%
    \begin{tabular}{ccccccc}
            \multicolumn{1}{c}{\small{B-scan}} & \multicolumn{1}{c}{\small{Expert Annotation}} & \multicolumn{1}{c}{\small{\NVdold{He\etal}\NVold{LBRM} \cite{2021_He}}} & \multicolumn{1}{c}{\small{U-Mamba	\cite{2024_Jun}}} & \multicolumn{1}{c}{\small{SD-RetinaNet + $\mathcal{L}_{\text{triplet}}$}}& \multicolumn{1}{c}{\small{SD-RetinaNet  }} \\
        \includegraphics[trim={0cm 3cm 0 0},clip,width=0.2\textwidth]{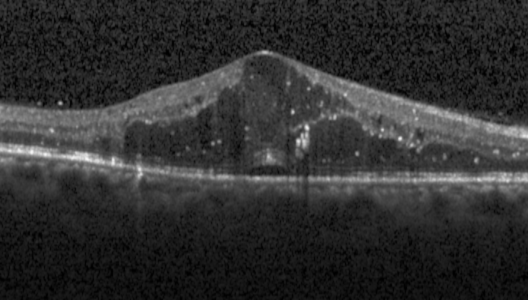} &
        \includegraphics[trim={0cm 2cm 0 0},clip,width=0.2\textwidth]{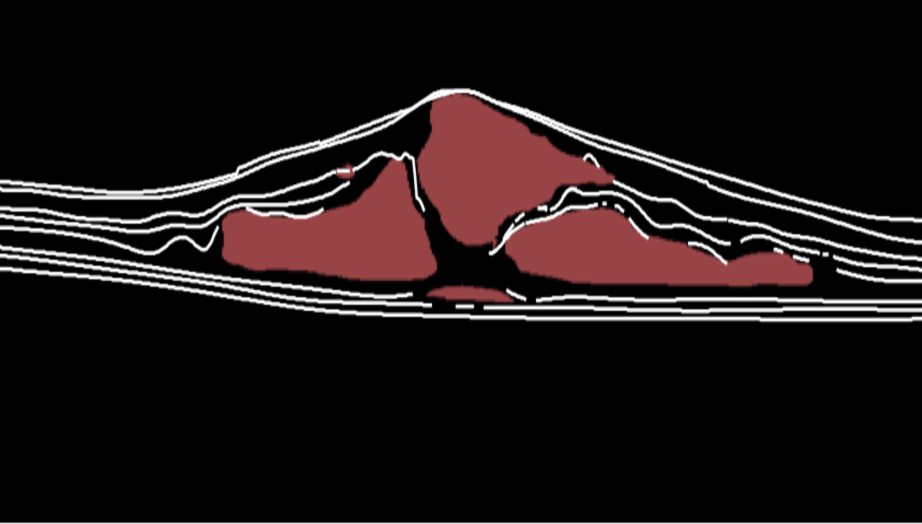} &
        \includegraphics[trim={0cm 2cm 0 0},clip,width=0.2\textwidth]{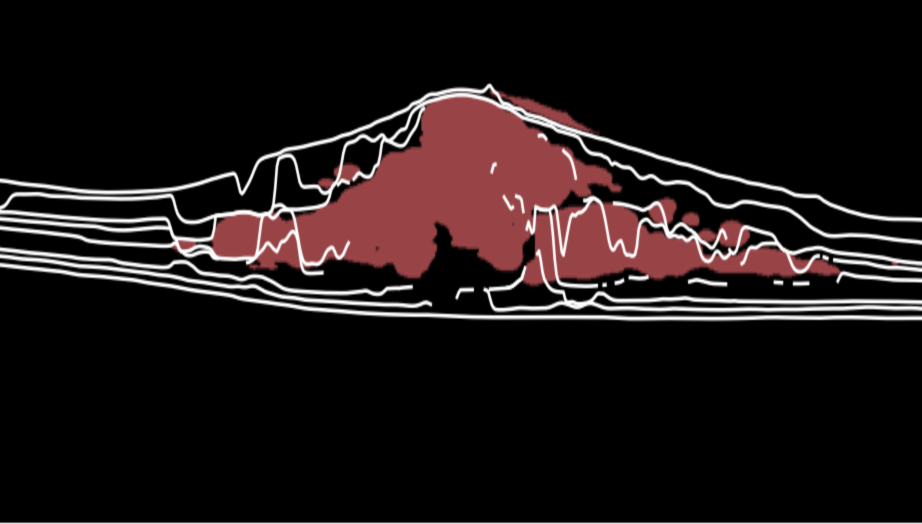} &
        \includegraphics[trim={0cm 2cm 0 0},clip,width=0.2\textwidth]{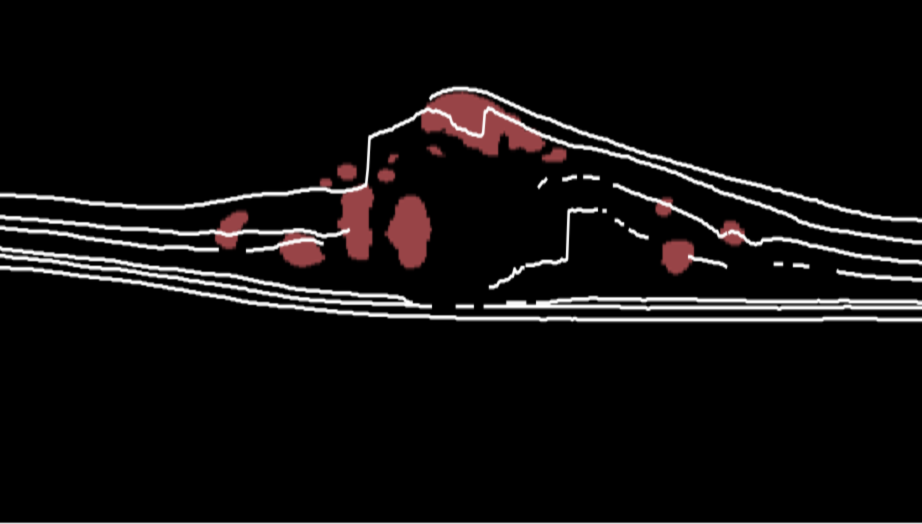} &
        \includegraphics[trim={0cm 2cm 0 0},clip,width=0.2\textwidth]{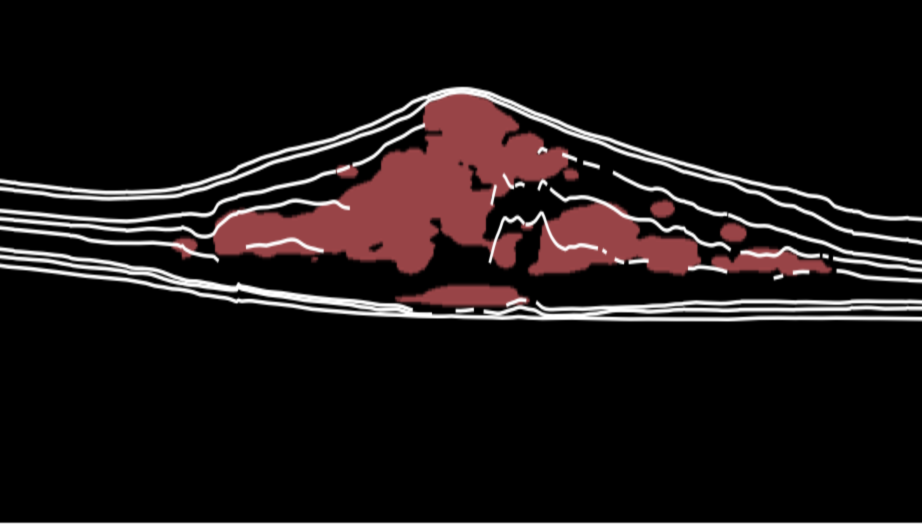} &
        \includegraphics[trim={0cm 2cm 0 0},clip,width=0.2\textwidth]{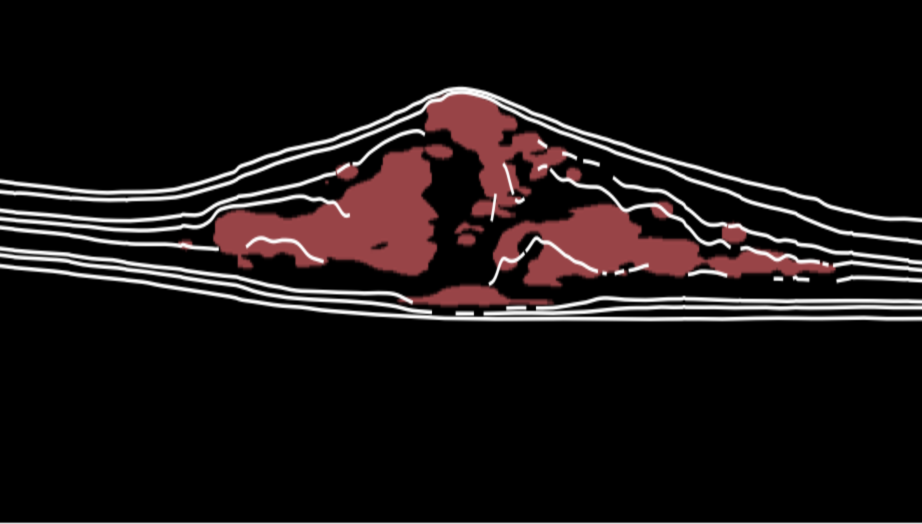} \\
        \includegraphics[trim={0cm 5cm 0 1.0cm},clip,width=0.2\textwidth]{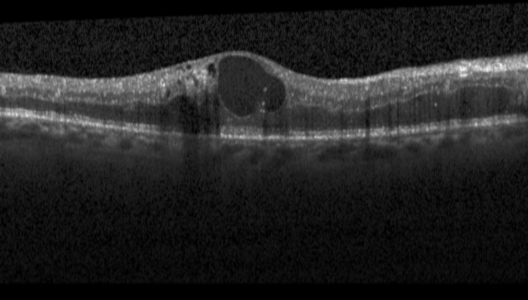} &
        \includegraphics[trim={0cm 3.5cm 0 0.5cm},clip,width=0.2\textwidth]{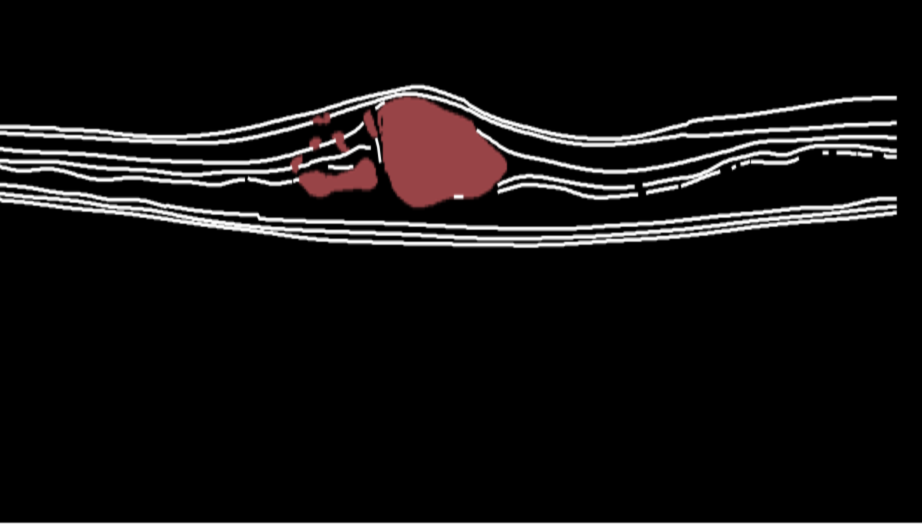} &
        \includegraphics[trim={0cm 3.5cm 0 0.5cm},clip,width=0.2\textwidth]{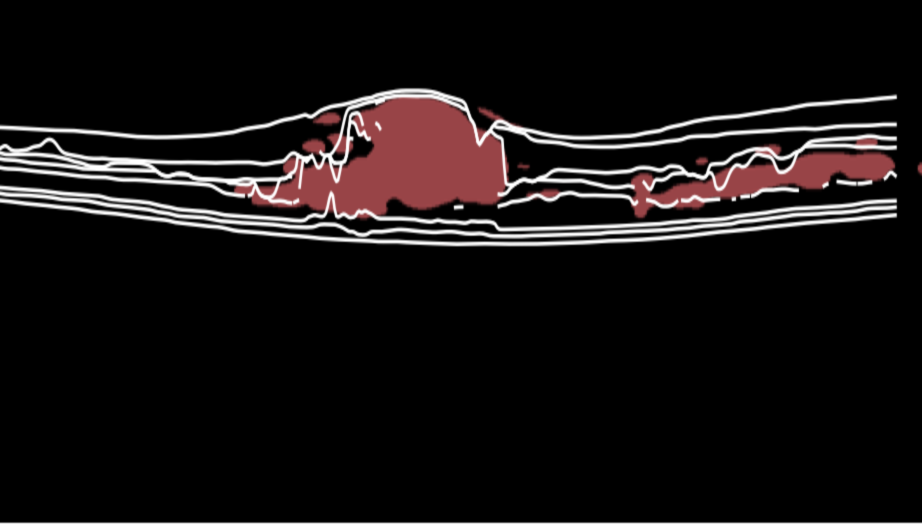} &
        \includegraphics[trim={0cm 3.5cm 0 0.5cm},clip,width=0.2\textwidth]{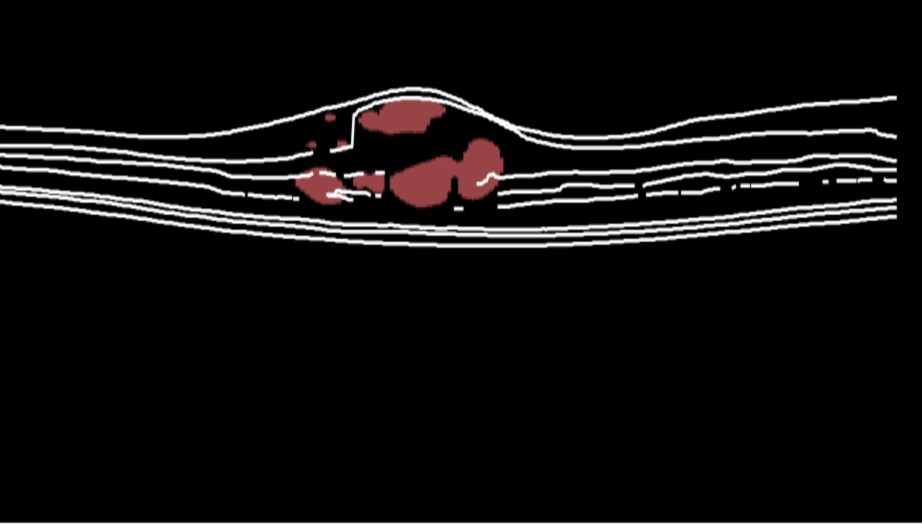} &
        \includegraphics[trim={0cm 3.5cm 0 0.5cm},clip,width=0.2\textwidth]{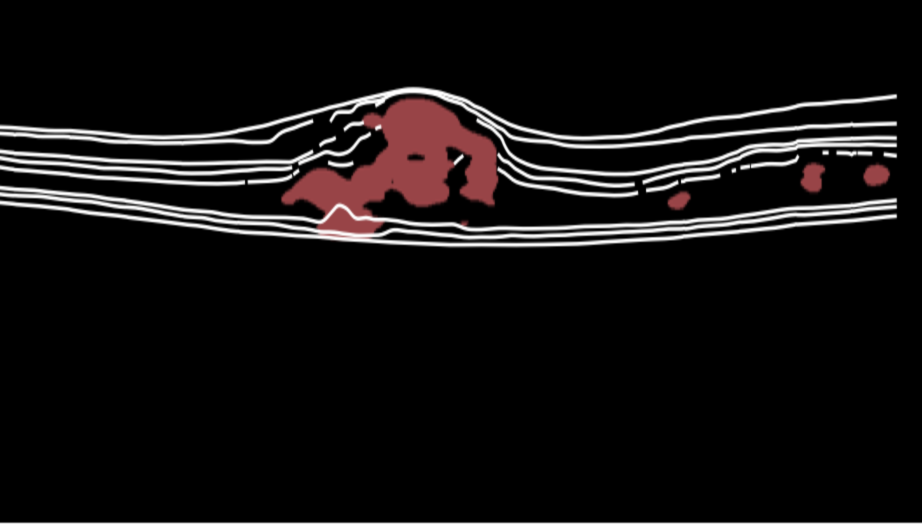} &
        \includegraphics[trim={0cm 3.5cm 0 0.5cm},clip,width=0.2\textwidth]{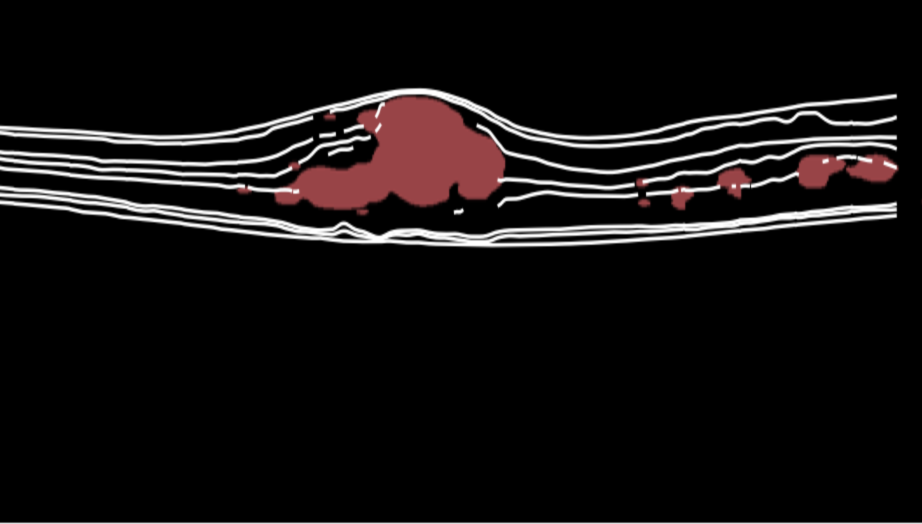} \\
        \includegraphics[width=0.2\textwidth]{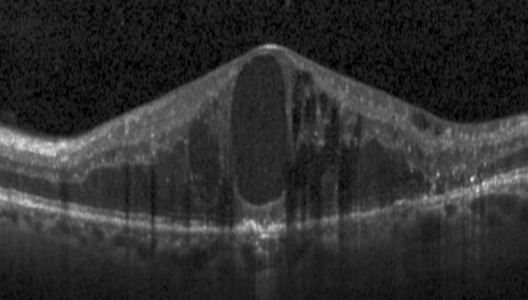} &
        \includegraphics[width=0.2\textwidth]{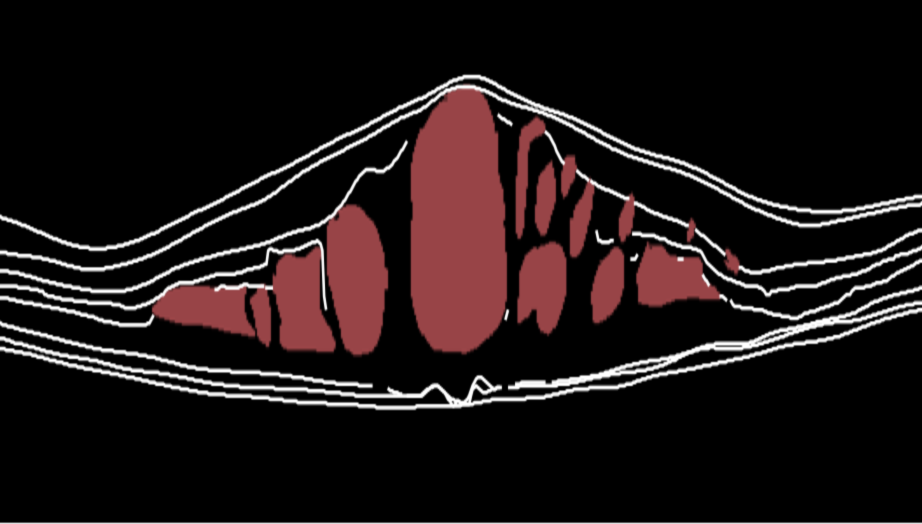} &
        \includegraphics[width=0.2\textwidth]{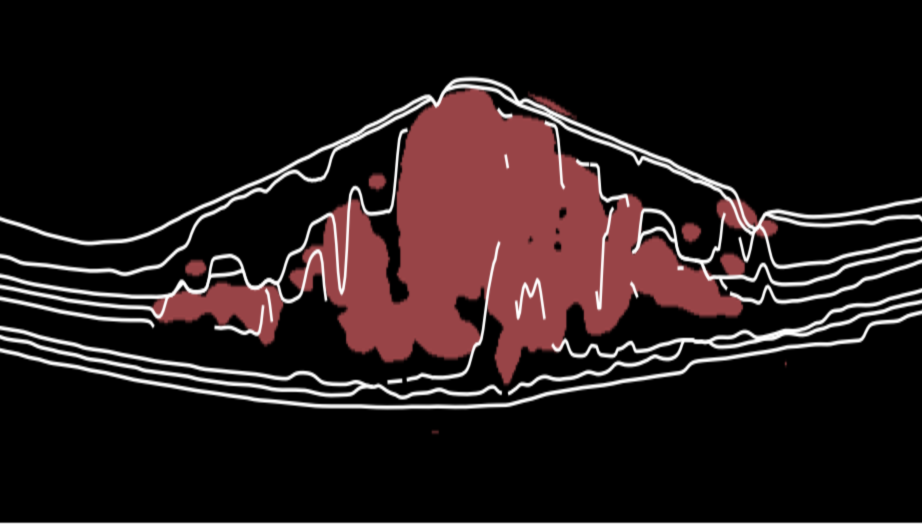} &
        \includegraphics[width=0.2\textwidth]{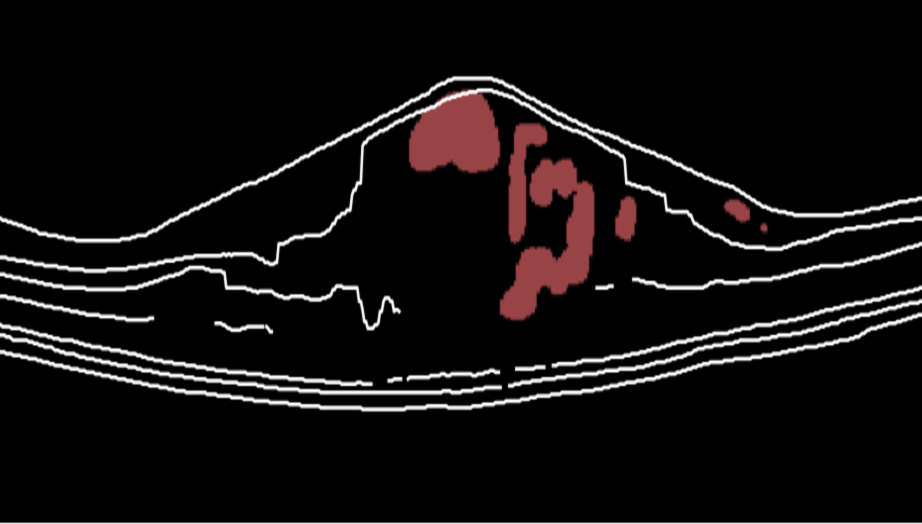} &
        \includegraphics[width=0.2\textwidth]{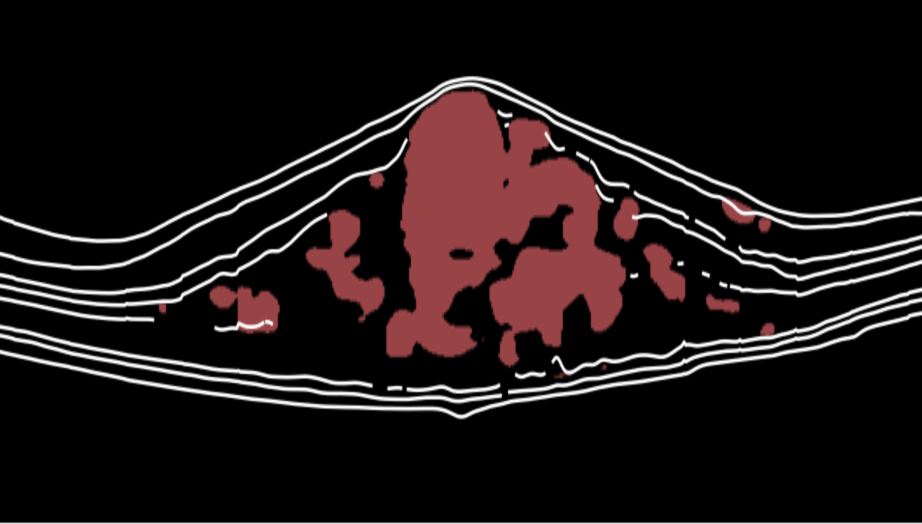} &
        \includegraphics[width=0.2\textwidth]{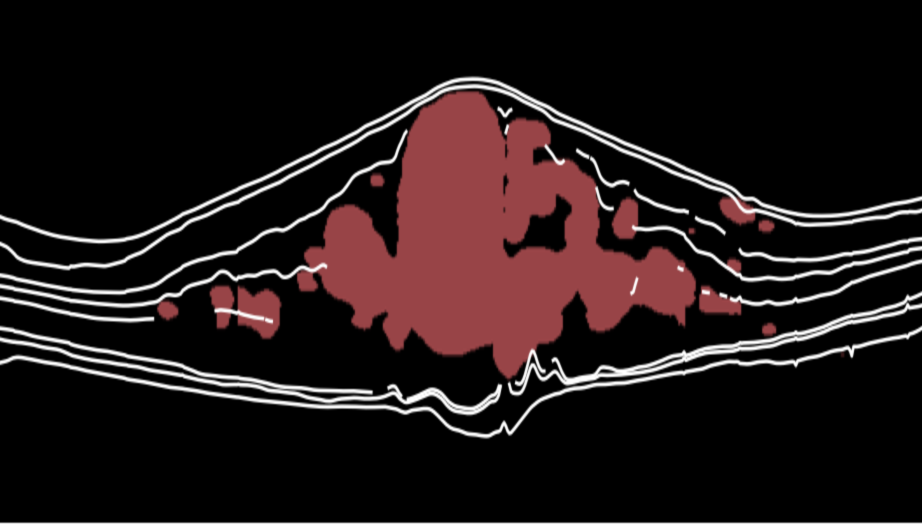} \\
    \end{tabular}
    }
    \caption{Qualitative results on the DukeDME test set, showing the original B-scan, expert annotations, the result of our method and of two baseline methods. TCCT-BP could not be trained with lesions only, so that example is not displayed. White: layer boundaries; Red: fluid.}
    \label{fig:duke_qualitative}
\end{figure*}

The partial annotation experiments on the DukeDME dataset (Table~\ref{tab:duke_layer_results}) 
demonstrate that semi-supervised methods trained on HCMS + RETOUCH Spectralis, have a distinct advantage over baseline methods that do not utilize multiple sources of partial labels simultaneously. All baseline methods faced significant challenges with the inner layers, which were heavily distorted by the fluids present in DME cases. Ablation studies indicate that both the reconstruction and lesion position constraints significantly enhance our model's performance. Incorporating the forced disentanglement constraint $\mathcal{L}_{\text{triplet}}$ significantly improved layer segmentation performance in the partial annotation experiments. This suggests that the contrastive separation of style and spatial factors aids the method in image domain shift scenarios and when using multiple datasets from different centers. 

The qualitative examples (Fig.~\ref{fig:duke_qualitative}) on three hard cases show that SD-RetinaNet + $\mathcal{L}_{\text{triplet}}$ was able to create plausible segmentations even in this challenging setting. For U-Mamba, several layers are completely missing, while also large areas with fluid were incorrectly segmented. \NVdold{The He\etal{} segmentation method}\NVold{LBRM} was able to identify more layer boundaries, however they are unstable and often crossing each other. \NV{It is important to note that the lower absolute Dice scores for all methods in this experiment (\Cref{tab:duke_layer_results}) reflect the extreme difficulty of the task. This scenario involves a significant domain shift from the training sources (HCMS, RETOUCH) to the test set (DukeDME) while simultaneously learning from disparate, partially-labeled data. The key finding is therefore the superior \emph{relative} performance of our framework under these challenging conditions.}

\subsection{\NV{Robustness to limited labeled data}\NVd{Label efficiency evaluation}}\label{section:limited_data_resusults}

\begin{figure}[tbh]
\centering
\includegraphics[width=0.45\textwidth,page=1]{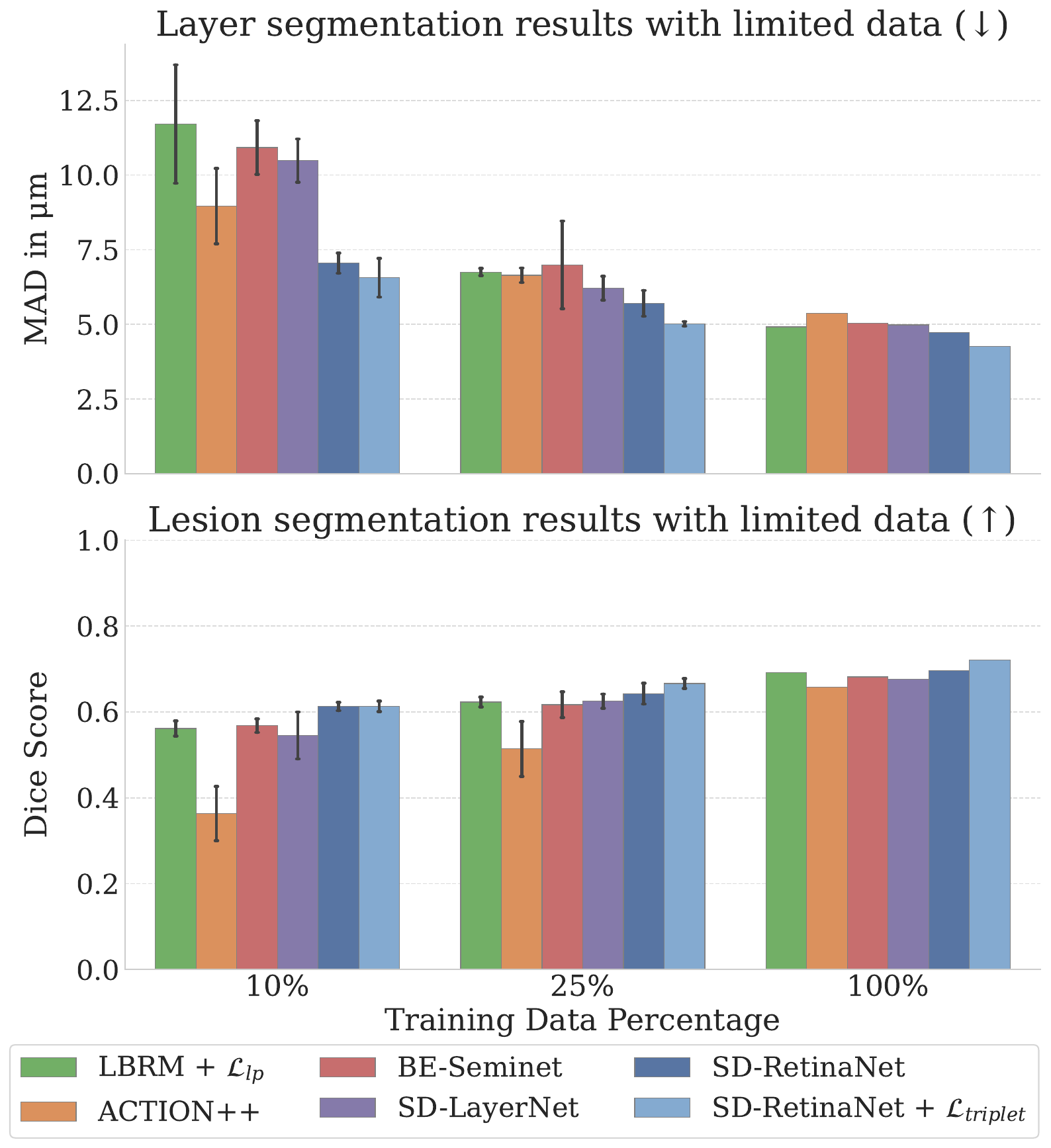}
\caption{\NVold{Segmentation performance robustness to reduced (10\%, 25\%) and full (100\%) training data  for proposed methods versus semi-supervised baselines. \textbf{Top}: Layer segmentation performance measured by Total Mean Absolute Distance (MAD in µm, lower is better). \textbf{Bottom}: Lesion segmentation performance measured by Total Dice score (higher is better). Error bars indicate standard deviation over multiple runs (10 runs for 10\%, 4 runs for 25\%). Results demonstrate superior robustness of the proposed methods across both layer and lesion segmentation tasks compared to baselines.}}
\label{fig:limited_data}
\end{figure}

\NVold{Figure~\ref{fig:limited_data} demonstrates the robustness of our methods compared to semi-supervised baselines when trained with reduced data (10\%, 25\%, 100\%). While all methods degraded with less data, our SD-RetinaNet variants consistently maintained a larger performance advantage in data-scarce settings. This was particularly evident for layer segmentation (Figure~\ref{fig:limited_data}, Top), where the MAD performance gap between SD-RetinaNet+$\mathcal{L}_{\text{triplet}}$ and the respective best baseline widened from 0.75 µm (100\% data) to 2.40 µm (10\% data). A similar, though less pronounced, trend favouring the robustness of our methods was observed for lesion segmentation Dice scores (Figure~\ref{fig:limited_data}, Bottom). These results highlight the effectiveness of our framework in leveraging anatomical priors and semi-supervised signals when direct annotation is limited.}

\subsection{\NV{Cross-Device Generalization}}
\label{section:cross_device}
\begin{figure}[tbh]
\centering
\includegraphics[width=0.45\textwidth,page=1]{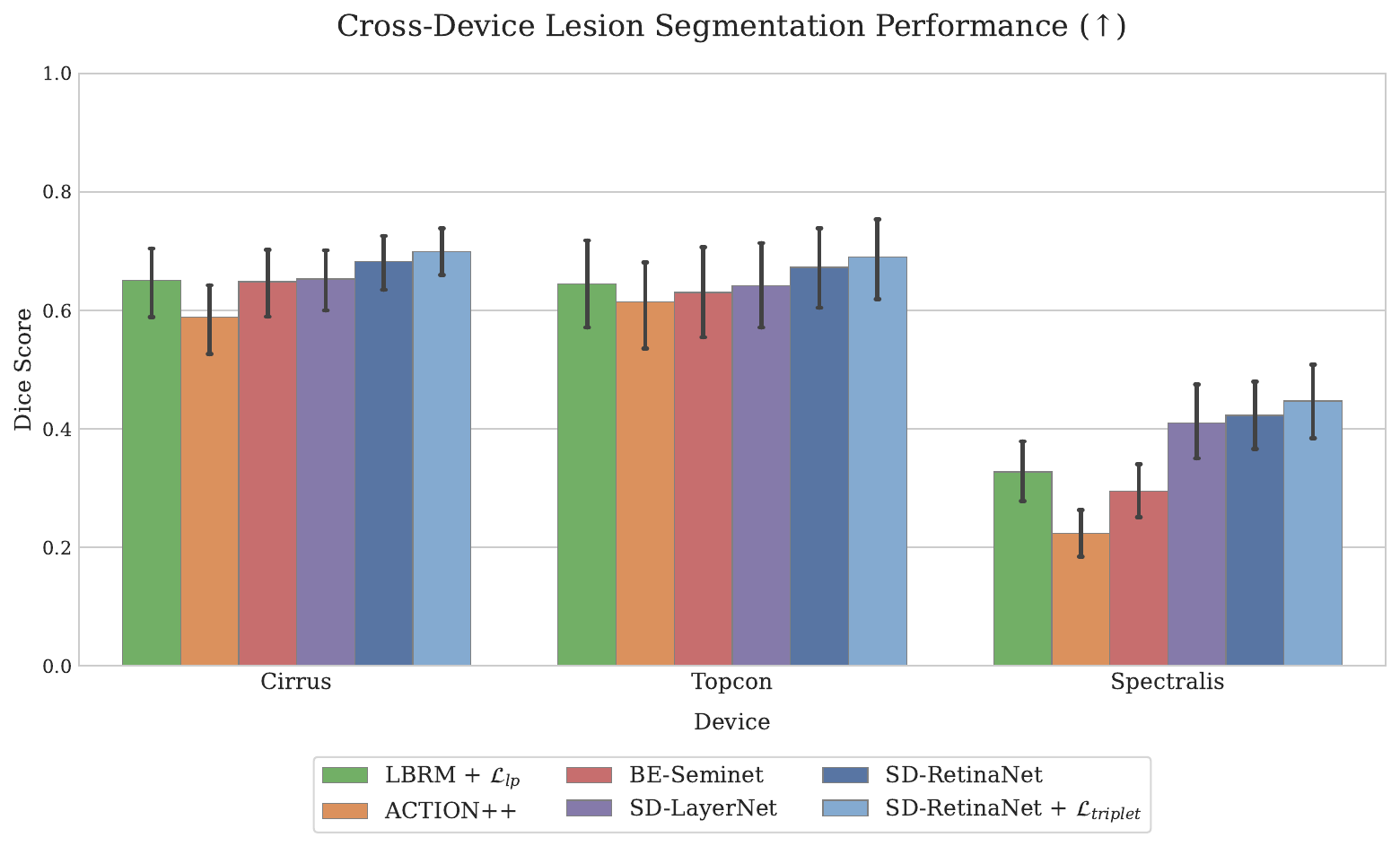}
\caption{\NV{Cross-device generalization performance for lesion segmentation. Each group on the x-axis represents the held-out test vendor. Models were trained on a combination of the other two vendors from the RETOUCH dataset (for lesions) and the HCMS dataset (for layers). Performance is measured by Dice score (higher is better). The proposed SD-RetinaNet variants consistently outperform baseline methods when generalizing to an unseen device.}}
\label{fig:cross_device_results}
\end{figure}

\NV{The results of the cross-device generalization experiments are presented in \Cref{fig:cross_device_results}. Across all three scenarios, our proposed SD-RetinaNet variants demonstrated superior generalization to unseen device vendors compared to all semi-supervised baselines, consistently achieving the highest Dice scores when testing on the Topcon and Cirrus devices.}

\NV{This advantage was most pronounced in the most challenging task where models were trained only on Spectralis data. While all methods' performance dropped, our framework proved most robust, with SD-RetinaNet + $\mathcal{L}_{\text{triplet}}$ achieving a Dice score of 0.45, maintaining a clear margin over the more significantly degraded baseline methods. This provides strong evidence that our disentanglement framework effectively separates device-specific style from anatomical content, leading to enhanced cross-device robustness. Nevertheless, the overall performance drop highlights that domain adaptation in this context remains an unsolved problem.}

\subsection{\NV{Analysis of disentangled latent space}\NVd{Effect of forced disentanglement on the latent space}}

\begin{figure}[tbh]
\centering
\includegraphics[width=0.45\textwidth,page=1]{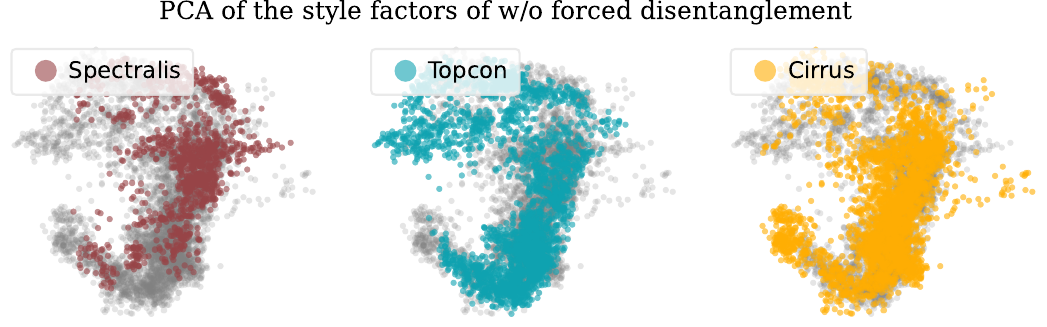}
\includegraphics[width=0.45\textwidth,page=1]{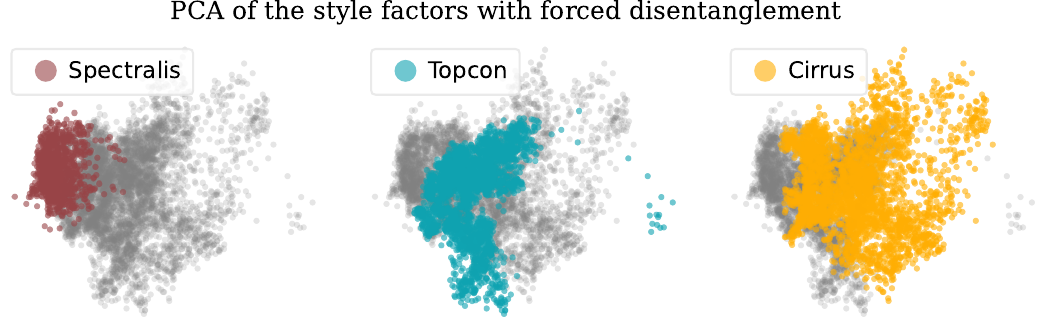}
\caption{\NVdold{Effect of forced disentanglement. The dots correspond to the style factors of B-scans. In each column, a specific device is highlighted, while the dots belonging to other devices are greyed out.}\NVold{Effect of forced disentanglement ($\mathcal{L}_{\text{triplet}}$) on the style factor ($\Omega$) latent space, visualized using 2D PCA projection. Each dot represents the style factor of a single B-scan. \textbf{Top row:} Style factors from a model trained without $\mathcal{L}_{\text{triplet}}$. \textbf{Bottom row:} Style factors from a model trained with $\mathcal{L}_{\text{triplet}}$. Each column highlights scans from a specific vendor, with other vendors greyed out. The bottom row demonstrates that adding $\mathcal{L}_{\text{triplet}}$ leads to a clearer separation of style factors based on the OCT device vendor, particularly separating Spectralis scans. While Topcon and Cirrus factors also show better separation, some overlap remains due to their visual similarities. This indicates that forced disentanglement enhances the style factors' ability to capture device-specific appearance characteristics.}}
\label{fig:forced_disentanglement_result}
\end{figure}

\NV{To qualitatively support the cross-device generalization results, we analyzed the style factor latent space (\Cref{fig:forced_disentanglement_result}). The experiment demonstrates that training with the forced disentanglement loss ($\mathcal{L}_{\text{triplet}}$) leads to a more organized representation, where the model more effectively clusters style factors based on their device vendor. This provides a visual confirmation of the mechanism enabling the robust cross-device performance shown in \Cref{fig:cross_device_results}.}
\NVd{The multi-device experiment (Figure~\ref{fig:forced_disentanglement_result}) demonstrates that with $\mathcal{L}_{\text{triplet}}$ the model developed a more organized representation of the style factors, effectively separating the style factors of scans acquired with different device vendors.}

\subsection{\NV{Analysis of multi-component loss optimization}}
\begin{figure}[tbh]
    \centering
        \includegraphics[width=0.48\textwidth]{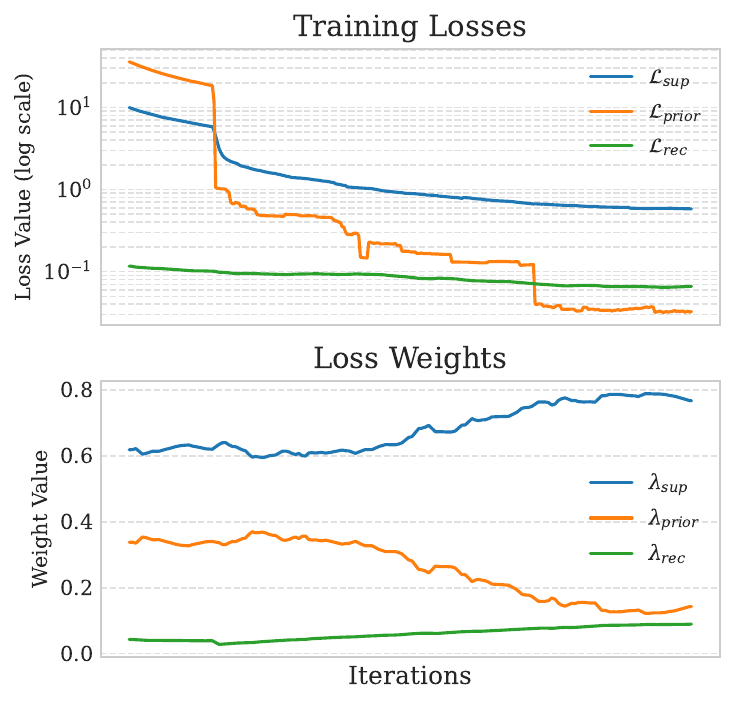}
    \caption{Analysis of training dynamics showing the interplay between key loss component groups and their adaptive weights assigned by SoftAdapt over training iterations. \emph{Top:} Training loss values (log scale) for the supervised ($\mathcal{L}_{\text{sup}}$), anatomical prior ($\mathcal{L}_{\text{prior}}$), and reconstruction ($\mathcal{L}_{\text{rec}}$) groups. \emph{Bottom:} Dynamic weights for each group as balanced by SoftAdapt. The supervised group ($\mathcal{L}_{\text{sup}}$) is the sum of individual supervised losses ($\mathcal{L}_{\text{KL}}, \mathcal{L}_{\text{L1}}, \mathcal{L}_{\text{dice}}$), and the plotted weight $\lambda_{\text{sup}}$ is the sum of their respective dynamic weights ($\lambda_{1-3}$). Likewise, the prior group ($\mathcal{L}_{\text{prior}}$) sums the anatomical prior losses ($\mathcal{L}_{\text{to}}, \mathcal{L}_{\text{bc}}, \mathcal{L}_{\text{lp}}$), and $\lambda_{\text{prior}}$ sums their weights ($\lambda_{4-6}$).}

    \label{fig:loss_dynamics}
\end{figure}

\NV{To address the complexity of our multi-term loss function and provide quantitative evidence of its stable optimization, we analyzed the training dynamics of key loss components and their corresponding weights assigned by the SoftAdapt technique (Figure~\ref{fig:loss_dynamics}).}

\NV{\Cref{fig:loss_dynamics} (Top) shows the values of the supervised ($\mathcal{L}_{sup}$), anatomical prior ($\mathcal{L}_{prior}$), and reconstruction ($\mathcal{L}_{rec}$) loss groups over 100 training epochs. All components consistently decrease, indicating that they all actively contribute to the learning process and none become stagnant.}

\NV{\Cref{fig:loss_dynamics} (Bottom) visualizes the dynamic weights automatically assigned to these loss components by SoftAdapt. The plot demonstrates the adaptive balancing in action: for example, the weight for the anatomical prior loss may be initially high while the network learns basic structural rules, and then relatively decrease as the supervised loss becomes more dominant in later epochs for fine-tuning. This dynamic balancing ensures that all loss terms contribute meaningfully throughout training, preventing any single component from overwhelming the optimization process due to differences in scale or learning speed. This analysis, combined with our ablation studies (\Cref{section:baseline_results}), provides strong evidence that our multi-component loss framework is optimized effectively and that each component plays a valuable role.}

\section{Discussion}
\NVold{The primary contribution of SD-RetinaNet is a novel framework tackling the significant challenge of joint, anatomically-plausible segmentation of interacting retinal layers and lesions in OCT.} \NVold{Modeling the inherent bidirectional influence – where layers constrain lesions and lesions simultaneously deform layers – especially while leveraging partially annotated datasets, represents a substantial advancement beyond prior layer-only methods \cite{2022_Fazekas_CONF, 2024_Tan, 2023_Lu} or approaches treating these tasks independently or sequentially \cite{2021_He}. This crucial capability is enabled by our core innovations: a differentiable biomarker topology engine combined with specific anatomical priors encoded as loss terms (particularly $\mathcal{L}_{\text{lp}}$) that facilitate this interaction and enforce topological consistency.} \NVold{While this framework is implemented using established network architectures, its novelty lies in these specialized components for joint, constrained learning.}
 The introduction of \NVdold{topological} \NVold{these} constraints \NVdold{in the proposed semi-supervised model} represents a significant advancement in the field of retinal lesion and layer segmentation within OCT images. By enforcing these constraints, the model ensures that the segmentation of lesions adheres to the anatomical structure of the retina, thereby reducing the occurrence of anatomically implausible segmentations that have been a limitation of previous methods. \NVold{This is facilitated by the proposed framework where initial segmentations from the Anatomy Module provide anatomical context, which is then actively refined. Semi-supervised anatomical prior losses guide the initial segmentation towards correctness by penalizing topological violations, while supervised losses calculated on the topologically corrected outputs further refine accuracy and enable the crucial bidirectional influence between layer and lesion predictions. The described refinement process maintains overall anatomical plausibility by accounting for all pixels throughout refinement. Notably, the proposed joint learning approach, driven by the anatomically-informed losses, allows the model to accurately segment layer boundaries distorted by pathology, even when layer training data primarily originates from healthy scans, by leveraging information from lesion segmentations. } \NVdold{This is particularly important for lesions, where accurate representation is crucial for effective diagnosis and treatment planning.}

The impact of these topological constraints is evident in the enhanced precision of lesion segmentation. Unlike traditional models that may produce topologically implausible segmentations, our approach guarantees that the lesions are segmented within their anatomically correct locations. This not only improves the overall accuracy of the segmentation but also provides a higher level of trust in the results, making it a valuable tool for clinicians who rely on these segmentations for critical decision-making. \NV{This advantage is particularly evident when compared against older, foundational architectures like ReLayNet, whose lower performance in our evaluation (\Cref{tab:lesions}) is largely attributable to a high number of false positive predictions in anatomically implausible locations, which our topological framework is explicitly designed to prevent.}

\NVold{The use of a multi-component loss function, incorporating over ten distinct terms, raises questions regarding optimization stability and the effective contribution of each component. The employed SoftAdapt technique \cite{2019_Heydari} automatically balances the influence of these diverse terms based on their relative performance during training, mitigating issues arising from different scales or convergence rates. While concerns about specific loss terms becoming inactive due to extreme weights are valid for complex optimization landscapes, our empirical results across multiple datasets and metrics, supported by ablation studies showing the negative impact of removing key components (Table~\ref{tab:mb_layer_results}, Table~\ref{tab:lesions}), indicate that the various losses contributed effectively to the final anatomically plausible and accurate segmentations. The effectiveness of leveraging these diverse loss terms within our semi-supervised framework is further highlighted by the label efficiency evaluation experiments (Section~\ref{section:limited_data_resusults}, Figure~\ref{fig:limited_data}). These results demonstrated superior robustness of our proposed methods compared to baseline semi-supervised approaches when labeled training data was significantly reduced, indicating that the anatomical priors and reconstruction task effectively compensate for scarce supervision.}

\NV{Furthermore, the value of the disentangled representation was directly tested in the challenging cross-device generalization experiments (Section~\ref{section:cross_device}, \Cref{fig:cross_device_results}). While all methods experienced a notable performance decrease when evaluated on a previously unseen device vendor, our model consistently outperformed the baseline methods, demonstrating greater robustness to this domain shift. This supports the conclusion that our framework's ability to separate device-specific style from anatomical content provides a tangible advantage for generalization, a critical capability for real-world clinical applications where data from multiple device types is common.}

Despite these advancements, there are still areas that require further investigation. The model's performance in more complex scenarios, such as cases involving multiple overlapping lesions or unusual anatomical variations, remains an open question. \NV{Analysis of the method's failure cases (e.g., \Cref{fig:mb_qualitative}, last row) reveals that limitations are not systemic but typically occur in the most challenging scenarios present in this real-world clinical dataset. Limitations were observed in cases with extreme signal loss due to shadowing artifacts or with subtle, rare pathological morphologies not well-represented in the training data, which can lead to false positives (e.g., for SHRM) or missed detections.} Additionally, the reliance on topological constraints, while beneficial, may need to be adapted for different types of retinal diseases where the anatomical structure might differ significantly. \NVold{Furthermore, the current lesion position constraint ($\mathcal{L}_{\text{lp}}$) relies on predefined anatomical associations ($\mathcal{S}_k$) for known lesion types. Consequently, for novel or rare lesion types not included in $\mathcal{S}_k$, this specific constraint would not apply, potentially leading to such lesions being segmented as part of the retinal layers or misclassified as known lesion types, although other loss terms continue to guide the overall segmentation.} Future work should focus on generalizing these topological constraints to other retinal conditions \NVold{or developing methods to handle unknown lesion types,} and exploring the potential of integrating this model with other imaging modalities to improve its robustness and applicability.

\section{Conclusion}
Our novel approach incorporates joint segmentation of layers and lesions, topological constraints, anatomical priors and contrastive disentanglement to train more robust retinal OCT biomarker representations, particularly in challenging regions around retinal lesions. Our proposed method outperforms the state of the art by a statistically significant margin in both layer and lesion segmentation tasks, while ensuring topological consistency between retinal layers and lesions and hence anatomical plausibility of the output. This provides a promising solution tackling the lack of topological coherence and spatial synergies between retinal layers and lesions in \NVb{previous methods}, resulting in superior performance.

\bibliographystyle{IEEETran}
\bibliography{sd_retinanet}

\begin{thebibliography}{10}
\providecommand{\url}[1]{#1}
\csname url@samestyle\endcsname
\providecommand{\newblock}{\relax}
\providecommand{\bibinfo}[2]{#2}
\providecommand{\BIBentrySTDinterwordspacing}{\spaceskip=0pt\relax}
\providecommand{\BIBentryALTinterwordstretchfactor}{4}
\providecommand{\BIBentryALTinterwordspacing}{\spaceskip=\fontdimen2\font plus
\BIBentryALTinterwordstretchfactor\fontdimen3\font minus \fontdimen4\font\relax}
\providecommand{\BIBforeignlanguage}[2]{{%
\expandafter\ifx\csname l@#1\endcsname\relax
\typeout{** WARNING: IEEEtran.bst: No hyphenation pattern has been}%
\typeout{** loaded for the language `#1'. Using the pattern for}%
\typeout{** the default language instead.}%
\else
\language=\csname l@#1\endcsname
\fi
#2}}
\providecommand{\BIBdecl}{\relax}
\BIBdecl

\bibitem{2004_Bressler}
N.~M. Bressler, ``Age-{Related} {Macular} {Degeneration} {Is} the {Leading} {Cause} of {Blindness}...'' \emph{Jama}, vol. 291, no.~15, pp. 1900--1901, 2004.

\bibitem{2017_SchmidtErfurth}
U.~Schmidt-Erfurth, S.~Klimscha, S.~M. Waldstein \emph{et~al.}, ``\BIBforeignlanguage{en}{A view of the current and future role of optical coherence tomography in the management of age-related macular degeneration},'' \emph{\BIBforeignlanguage{en}{Eye}}, vol.~31, no.~1, pp. 26--44, Jan. 2017.

\bibitem{2021_Liefers}
B.~Liefers, P.~Taylor, A.~Alsaedi \emph{et~al.}, ``\BIBforeignlanguage{en}{Quantification of {Key} {Retinal} {Features} in {Early} and {Late} {Age}-{Related} {Macular} {Degeneration} {Using} {Deep} {Learning}},'' \emph{\BIBforeignlanguage{en}{American Journal of Ophthalmology}}, vol. 226, pp. 1--12, Jun. 2021.

\bibitem{2010_Abramoff}
\BIBentryALTinterwordspacing
M.~D. Abràmoff, M.~K. Garvin, and M.~Sonka, ``Retinal {Imaging} and {Image} {Analysis},'' \emph{IEEE Reviews in Biomedical Engineering}, vol.~3, pp. 169--208, 2010.
\BIBentrySTDinterwordspacing

\bibitem{2018_DeFauw}
J.~De~Fauw, J.~R. Ledsam, B.~Romera-Paredes \emph{et~al.}, ``\BIBforeignlanguage{en}{Clinically applicable deep learning for diagnosis and referral in retinal disease},'' \emph{\BIBforeignlanguage{en}{Nature Medicine}}, vol.~24, no.~9, pp. 1342--1350, Sep. 2018.

\bibitem{2017_Montuoro}
\BIBentryALTinterwordspacing
A.~Montuoro, S.~M. Waldstein, B.~S. Gerendas \emph{et~al.}, ``\BIBforeignlanguage{EN}{Joint retinal layer and fluid segmentation in {OCT} scans of eyes with severe macular edema using unsupervised representation and auto-context},'' \emph{\BIBforeignlanguage{EN}{Biomedical Optics Express}}, vol.~8, no.~3, pp. 1874--1888, Mar. 2017.
\BIBentrySTDinterwordspacing

\bibitem{2019_Hassan}
\BIBentryALTinterwordspacing
T.~Hassan, M.~U. Akram, M.~F. Masood \emph{et~al.}, ``Deep structure tensor graph search framework for automated extraction and characterization of retinal layers and fluid pathology in retinal {SD}-{OCT} scans,'' \emph{Computers in Biology and Medicine}, vol. 105, pp. 112--124, Feb. 2019.
\BIBentrySTDinterwordspacing

\bibitem{2021_He}
Y.~He, A.~Carass, Y.~Liu \emph{et~al.}, ``\BIBforeignlanguage{en}{Structured layer surface segmentation for retina {OCT} using fully convolutional regression networks},'' \emph{\BIBforeignlanguage{en}{Medical Image Analysis}}, vol.~68, p. 101856, Feb. 2021.

\bibitem{2021_Wang}
B.~Wang, W.~Wei, S.~Qiu \emph{et~al.}, ``\BIBforeignlanguage{eng}{Boundary {Aware} {U}-{Net} for {Retinal} {Layers} {Segmentation} in {Optical} {Coherence} {Tomography} {Images}},'' \emph{\BIBforeignlanguage{eng}{IEEE journal of biomedical and health informatics}}, vol.~25, no.~8, pp. 3029--3040, Aug. 2021.

\bibitem{2023_Melo}
\BIBentryALTinterwordspacing
T.~Melo, {\^A}.~Carneiro, A.~Campilho \emph{et~al.}, ``{Retinal layer and fluid segmentation in optical coherence tomography images using a hierarchical framework},'' \emph{Journal of Medical Imaging}, vol.~10, no.~1, p. 014006, 2023.
\BIBentrySTDinterwordspacing

\bibitem{2019_Chartsias}
A.~Chartsias, T.~Joyce, G.~Papanastasiou \emph{et~al.}, ``\BIBforeignlanguage{en}{Disentangled representation learning in cardiac image analysis},'' \emph{\BIBforeignlanguage{en}{Medical Image Analysis}}, vol.~58, p. 101535, Dec. 2019.

\bibitem{2022_Fazekas_CONF}
B.~Fazekas, G.~Aresta, D.~Lachinov \emph{et~al.}, ``\BIBforeignlanguage{en}{{SD}-{LayerNet}: {Semi}-supervised {Retinal} {Layer} {Segmentation} in {OCT} {Using} {Disentangled} {Representation} with {Anatomical} {Priors}},'' in \emph{\BIBforeignlanguage{en}{Medical {Image} {Computing} and {Computer} {Assisted} {Intervention} – {MICCAI} 2022}}, ser. Lecture {Notes} in {Computer} {Science}, L.~Wang, Q.~Dou, P.~T. Fletcher \emph{et~al.}, Eds.\hskip 1em plus 0.5em minus 0.4em\relax Cham: Springer Nature Switzerland, 2022, pp. 320--329.

\bibitem{2006_Li}
K.~Li, X.~Wu, D.~Chen \emph{et~al.}, ``Optimal surface segmentation in volumetric images-{A} graph-theoretic approach,'' \emph{IEEE Transactions on Pattern Analysis and Machine Intelligence}, vol.~28, no.~1, pp. 119--134, Jan. 2006.

\bibitem{2014_Zhang}
L.~Zhang, M.~Sonka, J.~C. Folk \emph{et~al.}, ``\BIBforeignlanguage{eng}{Quantifying disrupted outer retinal-subretinal layer in {SD}-{OCT} images in choroidal neovascularization},'' \emph{\BIBforeignlanguage{eng}{Investigative Ophthalmology \& Visual Science}}, vol.~55, no.~4, pp. 2329--2335, Apr. 2014.

\bibitem{2010_Chiu}
\BIBentryALTinterwordspacing
S.~J. Chiu, X.~T. Li, P.~Nicholas \emph{et~al.}, ``\BIBforeignlanguage{EN}{Automatic segmentation of seven retinal layers in {SDOCT} images congruent with expert manual segmentation},'' \emph{\BIBforeignlanguage{EN}{Optics Express}}, vol.~18, no.~18, pp. 19\,413--19\,428, Aug. 2010.
\BIBentrySTDinterwordspacing

\bibitem{2013_Dufour}
P.~A. Dufour, L.~Ceklic, H.~Abdillahi \emph{et~al.}, ``Graph-based multi-surface segmentation of {OCT} data using trained hard and soft constraints,'' \emph{IEEE Transactions on Medical Imaging}, vol.~32, no.~3, pp. 531--543, Mar. 2013.

\bibitem{2014_Srinivasan}
\BIBentryALTinterwordspacing
P.~P. Srinivasan, S.~J. Heflin, J.~A. Izatt \emph{et~al.}, ``\BIBforeignlanguage{EN}{Automatic segmentation of up to ten layer boundaries in {SD}-{OCT} images of the mouse retina with and without missing layers due to pathology},'' \emph{\BIBforeignlanguage{EN}{Biomedical Optics Express}}, vol.~5, no.~2, pp. 348--365, Feb. 2014.
\BIBentrySTDinterwordspacing

\bibitem{2024_Chen}
\BIBentryALTinterwordspacing
Z.~Chen, H.~Zhang, E.~F. Linton \emph{et~al.}, ``Hybrid deep learning and optimal graph search method for optical coherence tomography layer segmentation in diseases affecting the optic nerve,'' \emph{Biomed. Opt. Express}, vol.~15, no.~6, pp. 3681--3698, Jun 2024.
\BIBentrySTDinterwordspacing

\bibitem{2010_Quellec}
\BIBentryALTinterwordspacing
G.~Quellec, K.~Lee, M.~Dolejsi \emph{et~al.}, ``Three-{Dimensional} {Analysis} of {Retinal} {Layer} {Texture}: {Identification} of {Fluid}-{Filled} {Regions} in {SD}-{OCT} of the {Macula},'' \emph{IEEE Transactions on Medical Imaging}, vol.~29, no.~6, pp. 1321--1330, Jun. 2010.
\BIBentrySTDinterwordspacing

\bibitem{2012_Chen}
\BIBentryALTinterwordspacing
X.~Chen, M.~Niemeijer, L.~Zhang \emph{et~al.}, ``Three-{Dimensional} {Segmentation} of {Fluid}-{Associated} {Abnormalities} in {Retinal} {OCT}: {Probability} {Constrained} {Graph}-{Search}-{Graph}-{Cut},'' \emph{IEEE Transactions on Medical Imaging}, vol.~31, no.~8, pp. 1521--1531, Aug. 2012.
\BIBentrySTDinterwordspacing

\bibitem{2015_Chiu}
\BIBentryALTinterwordspacing
S.~J. Chiu, M.~J. Allingham, P.~S. Mettu \emph{et~al.}, ``\BIBforeignlanguage{EN}{Kernel regression based segmentation of optical coherence tomography images with diabetic macular edema},'' \emph{\BIBforeignlanguage{EN}{Biomedical Optics Express}}, vol.~6, no.~4, pp. 1172--1194, Apr. 2015.
\BIBentrySTDinterwordspacing

\bibitem{2015_Ronneberger_CONF}
O.~Ronneberger, P.~Fischer, and T.~Brox, ``\BIBforeignlanguage{en}{U-{Net}: Convolutional networks for biomedical image segmentation},'' in \emph{\BIBforeignlanguage{en}{{MICCAI} 2015}}, ser. Lecture {Notes} in {Computer} {Science}.\hskip 1em plus 0.5em minus 0.4em\relax Cham: Springer International Publishing, 2015, pp. 234--241.

\bibitem{2017_Roy}
A.~G. Roy, S.~Conjeti, S.~P.~K. Karri \emph{et~al.}, ``\BIBforeignlanguage{EN}{{ReLayNet}: retinal layer and fluid segmentation of macular optical coherence tomography using fully convolutional networks},'' \emph{\BIBforeignlanguage{EN}{Biomedical Optics Express}}, vol.~8, no.~8, pp. 3627--3642, Aug. 2017.

\bibitem{2019_Bogunovic}
H.~Bogunovic, F.~Venhuizen, S.~Klimscha \emph{et~al.}, ``{RETOUCH}: {The} {Retinal} {OCT} {Fluid} {Detection} and {Segmentation} {Benchmark} and {Challenge},'' \emph{IEEE Transactions on Medical Imaging}, vol.~38, no.~8, pp. 1858--1874, Aug. 2019.

\bibitem{2021_Isensee}
\BIBentryALTinterwordspacing
F.~Isensee, P.~F. Jaeger, S.~A.~A. Kohl \emph{et~al.}, ``\BIBforeignlanguage{en}{{nnU}-{Net}: a self-configuring method for deep learning-based biomedical image segmentation},'' \emph{\BIBforeignlanguage{en}{Nature Methods}}, vol.~18, no.~2, pp. 203--211, Feb. 2021.
\BIBentrySTDinterwordspacing

\bibitem{2021_Dosovitskiy}
A.~Dosovitskiy, L.~Beyer, A.~Kolesnikov \emph{et~al.}, ``An {Image} is {Worth} 16x16 {Words}: {Transformers} for {Image} {Recognition} at {Scale},'' in \emph{International Conference on Learning Representations (2020)}, Jun. 2020.

\bibitem{2022_Hatamizadeh}
A.~Hatamizadeh, V.~Nath, Y.~Tang \emph{et~al.}, ``\BIBforeignlanguage{en}{Swin {UNETR}: {Swin} {Transformers} for {Semantic} {Segmentation} of {Brain} {Tumors} in {MRI} {Images}},'' in \emph{\BIBforeignlanguage{en}{Brainlesion: {Glioma}, {Multiple} {Sclerosis}, {Stroke} and {Traumatic} {Brain} {Injuries}}}, A.~Crimi and S.~Bakas, Eds.\hskip 1em plus 0.5em minus 0.4em\relax Cham: Springer International Publishing, 2022, pp. 272--284.

\bibitem{2024_Jun}
\BIBentryALTinterwordspacing
J.~Ma, F.~Li, and B.~Wang, ``U-{Mamba}: {Enhancing} {Long}-range {Dependency} for {Biomedical} {Image} {Segmentation},'' Tech. Rep., Jan. 2024, arXiv:2401.04722 [cs, eess] type: article.
\BIBentrySTDinterwordspacing

\bibitem{2024_Tan}
\BIBentryALTinterwordspacing
Y.~Tan, W.-D. Shen, M.-Y. Wu \emph{et~al.}, ``Retinal {Layer} {Segmentation} in {OCT} {Images} {With} {Boundary} {Regression} and {Feature} {Polarization},'' \emph{IEEE Transactions on Medical Imaging}, vol.~43, no.~2, pp. 686--700, Feb. 2024.
\BIBentrySTDinterwordspacing

\bibitem{2019_Liu}
X.~Liu, J.~Cao, T.~Fu \emph{et~al.}, ``Semi-{Supervised} {Automatic} {Segmentation} of {Layer} and {Fluid} {Region} in {Retinal} {Optical} {Coherence} {Tomography} {Images} {Using} {Adversarial} {Learning},'' \emph{IEEE Access}, vol.~7, pp. 3046--3061, 2019.

\bibitem{2018_Sedai_CONF}
S.~Sedai, B.~Antony, D.~Mahapatra \emph{et~al.}, ``Joint segmentation and uncertainty visualization of retinal layers in optical coherence tomography images using bayesian deep learning,'' in \emph{Computational Pathology and Ophthalmic Medical Image Analysis}.\hskip 1em plus 0.5em minus 0.4em\relax Springer International Publishing, 2018, pp. 219--227.

\bibitem{2019_Sedai_CONF}
\BIBentryALTinterwordspacing
S.~Sedai, B.~Antony, R.~Rai \emph{et~al.}, ``Uncertainty guided semi-supervised segmentation of retinal layers in oct images,'' in \emph{Medical Image Computing and Computer Assisted Intervention – MICCAI 2019}.\hskip 1em plus 0.5em minus 0.4em\relax Springer-Verlag, 2019, p. 282–290.
\BIBentrySTDinterwordspacing

\bibitem{2023_Lu}
\BIBentryALTinterwordspacing
Y.~Lu, Y.~Shen, X.~Xing \emph{et~al.}, ``Boundary-enhanced semi-supervised retinal layer segmentation in optical coherence tomography images using fewer labels,'' \emph{Computerized Medical Imaging and Graphics}, vol. 105, p. 102199, Apr. 2023.
\BIBentrySTDinterwordspacing

\bibitem{2017_Tarvainen}
\BIBentryALTinterwordspacing
A.~Tarvainen and H.~Valpola, ``Mean teachers are better role models: {Weight}-averaged consistency targets improve semi-supervised deep learning results,'' in \emph{Advances in {Neural} {Information} {Processing} {Systems}}, vol.~30.\hskip 1em plus 0.5em minus 0.4em\relax Curran Associates, Inc., 2017.
\BIBentrySTDinterwordspacing

\bibitem{2022_Wang}
\BIBentryALTinterwordspacing
K.~Wang, B.~Zhan, C.~Zu \emph{et~al.}, ``Semi-supervised medical image segmentation via a tripled-uncertainty guided mean teacher model with contrastive learning,'' \emph{Medical Image Analysis}, vol.~79, p. 102447, Jul. 2022.
\BIBentrySTDinterwordspacing

\bibitem{2022_Wu}
\BIBentryALTinterwordspacing
Y.~Wu, Z.~Ge, D.~Zhang \emph{et~al.}, ``Mutual consistency learning for semi-supervised medical image segmentation,'' \emph{Medical Image Analysis}, vol.~81, p. 102530, Oct. 2022.
\BIBentrySTDinterwordspacing

\bibitem{2023_Shi}
\BIBentryALTinterwordspacing
J.~Shi, T.~Gong, C.~Wang \emph{et~al.}, ``Semi-{Supervised} {Pixel} {Contrastive} {Learning} {Framework} for {Tissue} {Segmentation} in {Histopathological} {Image},'' \emph{IEEE Journal of Biomedical and Health Informatics}, vol.~27, no.~1, pp. 97--108, Jan. 2023.
\BIBentrySTDinterwordspacing

\bibitem{2022_Wua}
H.~Wu, Z.~Wang, Y.~Song \emph{et~al.}, ``Cross-patch dense contrastive learning for semi-supervised segmentation of cellular nuclei in histopathologic images,'' in \emph{Proceedings of the IEEE/CVF Conference on Computer Vision and Pattern Recognition (CVPR)}, June 2022, pp. 11\,666--11\,675.

\bibitem{2023_You}
C.~You, W.~Dai, Y.~Min \emph{et~al.}, ``\BIBforeignlanguage{en}{{ACTION}++: {Improving} {Semi}-supervised {Medical} {Image} {Segmentation} with {Adaptive} {Anatomical} {Contrast}},'' in \emph{\BIBforeignlanguage{en}{Medical {Image} {Computing} and {Computer} {Assisted} {Intervention} – {MICCAI} 2023}}, H.~Greenspan, A.~Madabhushi, P.~Mousavi \emph{et~al.}, Eds.\hskip 1em plus 0.5em minus 0.4em\relax Cham: Springer Nature Switzerland, 2023, pp. 194--205.

\bibitem{2019_Tan_CONF}
M.~Tan and Q.~Le, ``\BIBforeignlanguage{en}{{EfficientNet}: {Rethinking} {Model} {Scaling} for {Convolutional} {Neural} {Networks}},'' in \emph{\BIBforeignlanguage{en}{Proceedings of the 36th International Conference on Machine Learning}}.\hskip 1em plus 0.5em minus 0.4em\relax PMLR, May 201te9, pp. 6105--6114.

\bibitem{2021_Cleland}
S.~C. Cleland, S.~M. Konda, R.~P. Danis \emph{et~al.}, ``\BIBforeignlanguage{English}{Quantification of {Geographic} {Atrophy} {Using} {Spectral} {Domain} {OCT} in {Age}-{Related} {Macular} {Degeneration}},'' \emph{\BIBforeignlanguage{English}{Ophthalmology Retina}}, vol.~5, no.~1, pp. 41--48, Jan. 2021.

\bibitem{2018_Perez}
E.~Perez, F.~Strub, H.~de~Vries \emph{et~al.}, ``Film: Visual reasoning with a general conditioning layer,'' \emph{Proceedings of the AAAI Conference on Artificial Intelligence}, vol.~32, no.~1, Apr. 2018.

\bibitem{2017_Rathke_CONF}
F.~Rathke, M.~Desana, and C.~Schnörr, ``\BIBforeignlanguage{en}{Locally {Adaptive} {Probabilistic} {Models} for {Global} {Segmentation} of {Pathological} {OCT} {Scans}},'' in \emph{\BIBforeignlanguage{en}{Medical {Image} {Computing} and {Computer} {Assisted} {Intervention} - {MICCAI} 2017}}, ser. Lecture {Notes} in {Computer} {Science}, M.~Descoteaux, L.~Maier-Hein, A.~Franz \emph{et~al.}, Eds.\hskip 1em plus 0.5em minus 0.4em\relax Cham: Springer International Publishing, 2017, pp. 177--184.

\bibitem{2016_Karri}
\BIBentryALTinterwordspacing
S.~P.~K. Karri, D.~Chakraborthi, and J.~Chatterjee, ``\BIBforeignlanguage{EN}{Learning layer-specific edges for segmenting retinal layers with large deformations},'' \emph{\BIBforeignlanguage{EN}{Biomedical Optics Express}}, vol.~7, no.~7, pp. 2888--2901, Jul. 2016.
\BIBentrySTDinterwordspacing

\bibitem{2014_Bates}
D.~Bates, M.~M{\"a}chler, B.~Bolker \emph{et~al.}, ``Fitting linear mixed-effects models using lme4,'' \emph{arXiv preprint arXiv:1406.5823}, 2014.

\bibitem{2024_Liu}
\BIBentryALTinterwordspacing
H.~Liu, D.~Wei, D.~Lu \emph{et~al.}, ``Simultaneous alignment and surface regression using hybrid {2D}–{3D} networks for {3D} coherent layer segmentation of retinal {OCT} images with full and sparse annotations,'' \emph{Medical Image Analysis}, vol.~91, p. 103019, Jan. 2024.
\BIBentrySTDinterwordspacing

\bibitem{2019_Heydari}
\BIBentryALTinterwordspacing
A.~A. Heydari, C.~A. Thompson, and A.~Mehmood, ``{SoftAdapt}: {Techniques} for {Adaptive} {Loss} {Weighting} of {Neural} {Networks} with {Multi}-{Part} {Loss} {Functions},'' Tech. Rep., Dec. 2019, arXiv:1912.12355 [cs, math, stat] type: article.
\BIBentrySTDinterwordspacing

\end{thebibliography}

\end{document}